%% file: main.tex
\documentclass{article}

\usepackage[final]{corl_2026}

\usepackage{amsmath,amssymb,amsfonts,amstext}
\usepackage{bm}
\usepackage{graphicx}
\usepackage{booktabs}
\usepackage{longtable}
\usepackage{multirow}
\usepackage{siunitx}
\PassOptionsToPackage{table}{xcolor}
\usepackage{tcolorbox}
\usepackage{tabularx}
\newcolumntype{Y}{>{\raggedright\arraybackslash}X}
\usepackage{algorithm}
\usepackage{algpseudocode}

\usepackage{microtype}

\definecolor{frankaBlue}{HTML}{2A4F7D}
\definecolor{ur5Red}{HTML}{6E1622}
\definecolor{scoreBlue}{HTML}{527299}
\definecolor{scoreRed}{HTML}{CA7981}
\definecolor{dagBlue}{HTML}{527299}
\definecolor{dagRed}{HTML}{CA7981}

\definecolor{corlblue}{HTML}{2A4F7D}
\definecolor{corlred}{HTML}{6e1622}

\definecolor{dagmod}{HTML}{1B6E8C}
\newcommand{\dgm}[1]{\textcolor{dagmod}{#1}}
\newcommand{\tclowrho}[1]{\cellcolor{blue!8}{#1}}
\newcommand{\tcabove}[1]{\cellcolor{green!15}{#1}}
\newcommand{\tcnear}[1]{\cellcolor{gray!15}{#1}}

\newcommand{\linedashed}[1]{%
  \textcolor{#1}{%
    \rule[.5ex]{3pt}{0.8pt}\kern2pt%
    \rule[.5ex]{3pt}{0.8pt}\kern2pt%
    \rule[.5ex]{3pt}{0.8pt}}}

\newcommand{\linethick}[1]{%
  \textcolor{#1}{\rule[.5ex]{1.5em}{1.5pt}}}

\newcommand{\linedotted}[1]{%
  \textcolor{#1}{%
    \rule[.5ex]{1pt}{1pt}\kern3pt%
    \rule[.5ex]{1pt}{1pt}\kern3pt%
    \rule[.5ex]{1pt}{1pt}\kern3pt%
    \rule[.5ex]{1pt}{1pt}}}

\hypersetup{
    citecolor=corlred,
    colorlinks=true,
    linkcolor=black,
    urlcolor=corlred
}

\input{notation-shortcuts}

\title{Exploring the Intrinsic Geometry of Diffusion\\ Models with Constrained Inverse Kinematics}

\author{
  Miguel Angel Rogel Garcia\thanks{Equal contribution.}\,\,\,,
  Phone Thiha Kyaw\footnotemark[1]\,\,\,, and
  Jonathan Kelly \\
  Institute for Aerospace Studies, University of Toronto, Toronto, Canada \\
  \texttt{\{miguel.rogel,\,phone.thiha,\,jonathan.kelly\}@robotics.utias.utoronto.ca} \\
  \texttt{\url{https://papers.starslab.ca/diffusion-ik-geometry/}}
}

\begin{document}
\maketitle

\begin{figure}[!h]
\centering
\vspace{-2.5\baselineskip}
\includegraphics[width=0.72\linewidth]{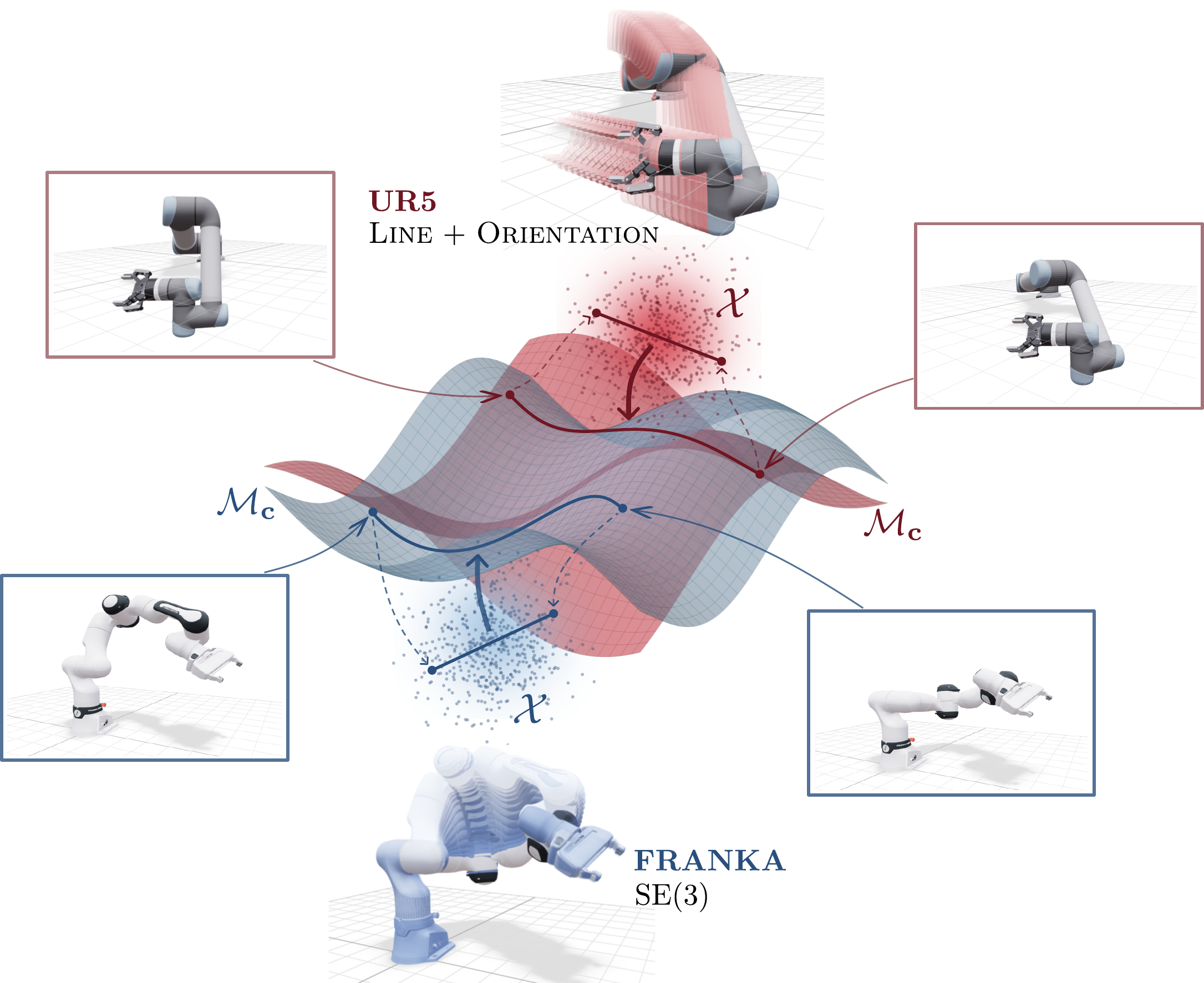}
\vspace{-2mm}
\caption{
\textbf {Diffusion models capture the geometry of constraint manifolds in inverse kinematics.}
Given a constraint family and a target pose, we analyze the learned diffusion model along two axes: the estimated intrinsic dimension of the model's generated samples and the behaviour of linear interpolations between samples in latent space.
}
\vspace{-3mm}
\label{fig:infographic}
\end{figure}

%===============================================================================

\begin{abstract}
Recent studies suggest that diffusion models can recover geometric structure in the data they are trained on, yet the supporting evidence has so far come mostly from natural-image data, where the underlying geometry itself is unknown.
We study this question in a setting where the geometry is analytically tractable: \emph{constrained inverse kinematics} (IK).
Each task-space constraint defines a configuration-space manifold with known intrinsic dimension, giving direct ground truth for evaluating the geometry learned by the model.
For each of the 6-DoF UR5 and 7-DoF Franka, we train a single conditional diffusion model across seven constraint families, spanning solution manifolds from discrete IK branches to self-motion manifolds.
Our empirical results reveal that the intrinsic dimension recovered from the model's score function matches the analytical degrees of freedom of the corresponding constraint manifold across both robots.
Moreover, linear interpolation in the latent space leads to generated solutions that remain close to the appropriate constraint manifold, indicating that the learned representation further captures geometric structure of the constraint family beyond intrinsic dimension alone.
Constrained IK therefore offers a controlled setting for studying the intrinsic geometry learned by diffusion models.
\end{abstract}


%===============================================================================

\input{sections/introduction}
\input{sections/background}
\input{sections/geometric_structure}
\input{sections/applications}
\input{sections/discussion}

%===============================================================================

\clearpage
\acknowledgments{This work was supported in part by the Canada Research Chairs Program, the Natural Sciences and Engineering Research Council of Canada (NSERC) through Discovery Grant no. RGPIN-2023-05036, and the Secretaría de Ciencia, Humanidades, Tecnología e Innovación (SECIHTI), Mexico, under CVU no. 1066015.}

%===============================================================================

\bibliography{main}

\clearpage
\appendix
\input{sections/appendix}

\end{document}

%% file: notation-shortcuts.tex
\usepackage{xparse}

\let\originalleft\left
\let\originalright\right
\renewcommand{\left}{\mathopen{}\mathclose\bgroup\originalleft}
\renewcommand{\right}{\aftergroup\egroup\originalright}

\NewDocumentCommand\Set{m}{ \left\{#1\right\} }
\NewDocumentCommand\Real{}{ \mathbb{R} }

\NewDocumentCommand\atantwo{}{\mathrm{atan2}}

\NewDocumentCommand\bbm{}{ \begin{bmatrix} } % Shortcut: start.
\NewDocumentCommand\ebm{}{ \end{bmatrix} }   % Shortcut: end.
\NewDocumentCommand\T{}{\mathsf{T}}          % Transpose symbol.

\NewDocumentCommand\Matrix{m}{ \bm{\mathbf{#1}} }

\NewDocumentCommand\Transpose{m}{ \left.{#1}\right.^\T }

\NewDocumentCommand\Inv{m}{{#1}^{-1}}
\NewDocumentCommand\Trace{m}{ \mathrm{tr}\left(#1\right) }

\NewDocumentCommand\Norm{m}{ \left\Vert#1\right\Vert }

\NewDocumentCommand\Zero{}{ \Matrix{0} }
\NewDocumentCommand\Identity{}{ \Matrix{I} }

\NewDocumentCommand\ArgMax{m}{ \operatorname*{argmax}_{#1} }

\NewDocumentCommand\LieGroupSO{m}{ \mathrm{SO}(#1) }

\NewDocumentCommand\LieGroupSE{m}{ \mathrm{SE}(#1) }

\NewDocumentCommand\NormalDistribution{mm}{ \mathcal{N}\left(#1, #2\right) }

%% file: sections/introduction.tex
\section{Introduction}
\label{sec:introduction}
Diffusion models have become the dominant generative framework for modelling high-dimensional data.
Although they are trained only to denoise, a growing line of work reveals that their internal representations recover surprisingly rich geometric structure.
The score function aligns with the normal bundle of the data manifold and recovers its intrinsic dimension~\citep{stanczuk2024diffusion}, which itself varies with the conditioning input~\citep{kvinge2023exploring}.
Pullback metrics on feature spaces yield semantically meaningful tangent directions~\citep{park2023understanding}.
Riemannian metrics built from the score admit geodesics that remain on the data manifold~\citep{diepeveen2024score,azeglio2025s}.
Taken together, these findings suggest that diffusion models can encode aspects of the geometry of the distributions on which they are trained.

\looseness=-1
Much of this evidence has been collected on natural-image manifolds, where the intrinsic dimension itself is an estimated quantity~\citep{pope2021intrinsic,brown2022verifying} and the topology is recovered only indirectly.
We complement this line of work by studying the \textit{constrained inverse kinematics} setting. 
Here, the data manifold is given analytically across a parametric family, as the IK solution set forms a constraint manifold of known dimension and topology.
This manifold varies smoothly with a small conditioning vector, and we ask how much of that variation is reflected in the geometry of the trained model's latent space.

While prior generative IK solvers have focused on sample quality~\cite{ames2022ikflow,park2022nodeik,bensadoun2022neural,limoyo2024generative,zhang2025ikdiffuser}, we instead treat constrained IK as a controlled setting for analyzing diffusion-model geometry.
Closest to our work, \citet{foland2025constraint} measure constraint-manifold adherence on a bimanual task whose constraint is defined implicitly by demonstrations; our analytic constraints let us evaluate manifold adherence directly, and we also ask whether the model captures intrinsic geometry.
For each of the 6-DoF UR5 and the 7-DoF Franka manipulators, we train a single diffusion model across all constraint types and analyze the latent space the model induces.
Our empirical findings show that the intrinsic dimension estimated from the learned score function matches the analytically known intrinsic dimension of the constraint manifold for every constraint type on both robots.
We further show that solutions generated from linear interpolation in the latent space remain close to the corresponding constraint manifold, indicating that the learned representation captures the geometric structure of the entire constraint family beyond dimensionality.
We demonstrate two practical applications enabled by this learned geometric structure.
First, we estimate the normal space to the data manifold and project the score onto this space, lowering the constraint error of generated configurations without an analytical Jacobian.
Second, we compare the estimated intrinsic dimensionality between training and held-out configurations, and utilize the measured gap as a diagnostic to verify training convergence and generalization to unseen samples.

\textbf{Contributions.}
(i) We present constrained IK as a controlled setting for diffusion-model interpretability in which the data manifold is given analytically across a parametric family of varying intrinsic dimension.
(ii) We train a single conditional diffusion model that encodes this entire family, and show that the dimension of its latent representations matches the analytical ground truth for every constraint.
(iii) We provide evidence that linear interpolation in the learned latent space remains close to the constraint manifold, with generated motion largely following the local geometry of the manifold, demonstrating that the latent representation encodes more than intrinsic dimension alone.

%% file: sections/background.tex
\vspace{-1mm}
\section{Background}
\label{sec:background}

This section reviews the relevant technical background on constraint manifolds, diffusion models and intrinsic dimensionality.
A notation glossary is provided in Appendix~\ref{app:notation}.

\paragraph{Constraint manifolds and their intrinsic geometry}
Constrained inverse kinematics provides a setting in which the geometry of the constraint manifold around which the data is distributed is available analytically.
This makes constrained IK well-suited for studying the internal structure of diffusion models: rather than inferring the data manifold from samples, we can compare the learned representation against a manifold whose geometry is known a priori.
Unlike in classical IK, a task-space constraint need not specify a single end-effector pose; it may fix only some of the task-space degrees of freedom, yielding a constraint manifold in configuration space.
Such constraint sets are well characterized in the robot kinematics literature, with their dimension, redundancy structure, and connectivity available in closed form for a wide range of manipulators~\cite{burdick1989inverse,siciliano2009robotics,lynch2017modern,murray2017mathematical,kingston2019exploring}.
We use this body of analytical results to define the ground truth for our model.

Let $\mathbf{q} \in \Real{}^n$ denote the joint configuration of a manipulator with $n$ actuated joints, and let $f : \Real{}^n \to \LieGroupSE{3}$ denote its forward kinematics.
For a task-space specification $\mathbf{c}$ that restricts $f(\mathbf{q})$ to a subset $\mathcal{S}_\mathbf{c} \subseteq \LieGroupSE{3}$, the set of configurations satisfying the constraint is
\begin{equation}
\mathcal{M}_\mathbf{c} \;=\; \{\mathbf{q} \in \Real{}^n : f(\mathbf{q}) \in \mathcal{S}_\mathbf{c}\} \;\subset\; \Real{}^n.
\end{equation}
When $\mathcal{S}_\mathbf{c}$ is a smooth submanifold of $\LieGroupSE{3}$ of codimension $k$ and $f$ has full-rank Jacobian on $\mathcal{M}_\mathbf{c}$, the implicit function theorem gives $\mathcal{M}_\mathbf{c}$ the structure of a smooth embedded submanifold of configuration space with \emph{intrinsic dimension} $n - k$~\cite{lee2012smooth}.
The codimension $k$ equals the number of task-space coordinates fixed by the constraint; as $k$ increases, the feasible set loses degrees of freedom and $\mathcal{M}_\mathbf{c}$ shrinks from higher-dimensional continuous families through two- and one-dimensional ones down to a finite point set.
The constraints we consider range from $k=6$ for a full $\LieGroupSE{3}$ point-target down to $k=1$ for a single-coordinate restriction.
These include both purely positional restrictions (point, line, or plane in $\Real{}^3$) and combinations involving end-effector orientation in $\LieGroupSO{3}$.

The topology of $\mathcal{M}_\mathbf{c}$ also depends on the kinematic redundancy of the manipulator relative to the constraint.
For the non-redundant 6-DoF UR5 under a full $\LieGroupSE{3}$ pose constraint, $\mathcal{M}_\mathbf{c}$ collapses to a finite set of configurations: the classical IK branches for 6R manipulators~\cite{pieper1969kinematics,raghavan1993inverse}.
For the redundant 7-DoF Franka under the same constraint, the extra joint promotes $\mathcal{M}_\mathbf{c}$ to a one-dimensional self-motion manifold along which the arm can reconfigure while the end-effector pose stays fixed~\cite{burdick1989inverse}.

The quantities we use to study the learned latent space (the dimension of $\mathcal{M}_\mathbf{c}$, its tangent space $\mathcal{T}_{\mathbf{q}}\mathcal{M}_\mathbf{c}$, and the constraint error from a generated configuration to $\mathcal{S}_\mathbf{c}$) are available in closed form from $f$ and $\mathbf{c}$, and do not need to be estimated from samples.
The same cannot be said of natural-image manifolds, where the dimension being recovered is itself an estimate.
The analytical $\mathcal{M}_\mathbf{c}$ serves as the reference against which we compare both the score-based intrinsic dimension estimate in Section~\ref{sec:id} and the on-manifold behaviour of latent-space interpolations in Section~\ref{sec:interpolation}.

\paragraph{Conditional diffusion models and DDIM inversion}
In constrained inverse kinematics, the support of the data distribution varies with the conditioning input.
Rather than fitting a separate model per constraint, we train a single model whose conditioning input $\mathbf{c}$ selects both the constraint type and the task-space target (Section~\ref{sec:results}), so varying $\mathbf{c}$ changes the dimension and topology of the manifold on which the conditional distribution is supported.

\looseness-1
A denoising network $\hat{\boldsymbol{\epsilon}}_\theta(\mathbf{x}_t, t, \mathbf{c})$ is trained to predict the Gaussian noise added to a clean sample $\mathbf{x}_0$ at noise level $t$, with variance schedule $\bar\alpha_t$, from the noised latent $\mathbf{x}_t \in \mathcal{X} \subseteq \mathbb{R}^n$ and the condition $\mathbf{c}$~\cite{ho2020denoising}.
Up to a time-dependent scaling, this predicted noise is an estimator of the Stein score of the conditional noised marginal~\citep{song2021score,song2021denoising},
\begin{equation}
s_\theta(\mathbf{x}_t, t, \mathbf{c})
=
\nabla_{\mathbf{x}_t}\log p_t(\mathbf{x}_t \mid \mathbf{c})
\approx
-\frac{\hat{\boldsymbol{\epsilon}}_\theta(\mathbf{x}_t, t, \mathbf{c})}
{\sqrt{1-\bar\alpha_t}}.
\end{equation}
The deterministic DDIM sampler~\citep{song2021denoising} defines a map $g_{\mathbf{c}}: \mathcal{X} \to \Real^{n}$ from a latent $\mathbf{x}_T$ to a sample $\mathbf{x}_0$, and running its update forward in time (\emph{DDIM inversion}) recovers a latent that maps back to $\mathbf{x}_0$.
Stopping inversion at an intermediate noise level $\bar{t} < T$ instead returns a \emph{partial} latent $\mathbf{x}_{\bar{t}}$ that retains most of $\mathbf{x}_0$'s structure; we use this latent variable for the interpolation experiments in Section~\ref{sec:interpolation}.
The forward process, the DDIM update and its inversion are given in Appendix~\ref{app:diffusion-background}.

\paragraph{Intrinsic dimension estimation}
The manifold hypothesis posits that real-world high-dimensional data often concentrate near a much lower-dimensional submanifold, whose intrinsic dimension (ID) reflects the number of local factors of variation~\citep{fukunaga1971algorithm, fefferman2013manifoldhypothesis}.
Estimating this dimension matters in practice: it lower-bounds the sample complexity of learning~\citep{pope2021intrinsic}, and linear directions in latent space often carry semantic meaning when the dimension is small~\citep{park2023understanding}.
A first family of ID estimators infers the dimension from data-space neighbourhood statistics: local PCA on the nearest neighbours~\citep{fukunaga1971algorithm}, maximum-likelihood on Poisson-process neighbour counts~\citep{levina2004maximum}, or the closed-form 1NN/2NN ratio of TwoNN~\cite{facco2017estimating}.
A second, more recent family uses the score function of a trained diffusion model directly.
\citet{stanczuk2024diffusion} show that, as the noise level $t \to 0$, the score $s_\theta(\mathbf{x}_{t}, t)$ at a noised sample $\mathbf{x}_{t}$ points back toward the manifold $\mathcal{M}_\mathbf{c}$, in a direction normal to it.
Stacking the score at $K$ independent perturbations of $\mathbf{x}_0$ therefore yields a matrix whose top singular vectors span the normal space $\mathcal{N}_{\mathbf{x}_0}\mathcal{M}_\mathbf{c}$.

We adopt the method proposed in \citep{stanczuk2024diffusion} to estimate the ID of the learned latent space. Concretely, for a sample $\mathbf{x}_0$, one draws $K$ independent Gaussian perturbations at a single time $t$, evaluates the network on each, and stacks the resulting score vectors as the columns of
\begin{equation*}
S \;=\; \left[s_\theta(\mathbf{x}^{(1)}_{t},t) \, \ldots \, s_\theta(\mathbf{x}^{(K)}_{t},t)\right] \in \Real{}^{n \times K}. 
\end{equation*}
Recall that $\mathcal{M}_\mathbf{c}$ has codimension $k$ in $\mathbb{R}^n$, so its normal space at any point is $k$-dimensional.
The left singular vectors of $S$ corresponding to the $k$ largest singular values then span $\mathcal{N}_{\mathbf{x}_0}\mathcal M_{\mathbf{c}}$, so $\hat{d} = n - \hat{k}$ estimates the intrinsic dimension, with $\smash{\hat{k}}$ selected as the index of the largest consecutive drop in the singular value spectrum
$\sigma_1 \geq \cdots \geq \sigma_n$,
\begin{equation}
\label{eq:score_svd_rule}
\hat{k} \;=\; \ArgMax{i=1,\ldots,n-1}\,(\sigma_i - \sigma_{i+1}).
\end{equation}
\looseness=-1
Two hyperparameters control the estimator: the time $t$, which must be small enough for the normal-bundle alignment to hold yet large enough that $\mathbf{x}_{t}$ stays in the network's training distribution, and the perturbation count $K$, which must exceed the codimension $k$ for the normal space to be fully spanned.
In our experiments, within a useful intermediate range of $t$, $\hat{d}$ recovers the analytical ID, while values of $t$ too small or too large degrade the estimate.
The construction above, which we call Score-SVD, assumes $\mathcal{M}_\mathbf{c}$ is positive-dimensional. It therefore does not directly apply to 0-dimensional data manifolds such as finite, isolated point sets, which is the case for the UR5 under a full $\LieGroupSE{3}$ pose constraint.
In Section~\ref{sec:id}, we modify the Score-SVD approach to estimate the ID in this 0-dimensional case.

%% file: sections/geometric_structure.tex
\vspace{-1mm}
\section{Geometric Structures in the Latent Space}
\label{sec:results}
 
\begin{table}[t]
\centering
\caption{\textbf{The seven task-space constraints considered in this work.}
``Fixed'' denotes the number $k$ of task-space DoFs imposed by each constraint, and
``Free'' denotes the intrinsic dimension $n - k$ of the joint-space solution manifold, given for the $6$-DoF UR5 and the $7$-DoF Franka.
}
\label{tab:constraints}
\small
\begin{tabular}{llccc}
\toprule
 & & & \multicolumn{2}{c}{Free (intrinsic dim.)} \\
\cmidrule(lr){4-5}
Constraint & End-effector quantity fixed & Fixed & UR5 & Franka \\
\midrule
\textsc{Plane}              & one position coordinate                 & 1 & 5 & 6 \\
\textsc{Line}               & two position coordinates                & 2 & 4 & 5 \\
\textsc{Position}           & three position coordinates              & 3 & 3 & 4 \\
\textsc{Orientation}        & three rotation coordinates              & 3 & 3 & 4 \\
\textsc{Plane} $+$ \textsc{Orientation}\ & one position coord.\ $+$ full rotation  & 4 & 2 & 3 \\
\textsc{Line} $+$ \textsc{Orientation}\  & two position coords.\ $+$ two rotation coords. & 4 & 2 & 3 \\
$\LieGroupSE{3}$   & full pose                               & 6 & 0 & 1 \\
\bottomrule
\end{tabular}
\vspace{-4mm}
\end{table}

\looseness=-1
Throughout this work, we study the geometry of the learned latent space under a family of progressively tighter task-space constraints, evaluated on two manipulators with different kinematic redundancy.
The seven constraints (Table~\ref{tab:constraints}) range from fixing one position coordinate for a plane constraint, to fixing the complete $\LieGroupSE{3}$ pose.
Each constraint is evaluated on both the non-redundant $6$-DoF UR5 and the redundant $7$-DoF Franka, yielding fourteen constraint--manipulator conditions in total.
The two manipulators are chosen because they produce qualitatively different solution-set geometries under identical constraints: for the UR5 under a full $\LieGroupSE{3}$ pose constraint, IK solutions form a finite set of disconnected configuration branches, whereas the additional redundant joint of the Franka induces a one-dimensional self-motion manifold.
Precise constraint definitions are given in Appendix~\ref{app:constraints}.

\begin{figure}[t]
\centering
\includegraphics[width=0.85\linewidth]{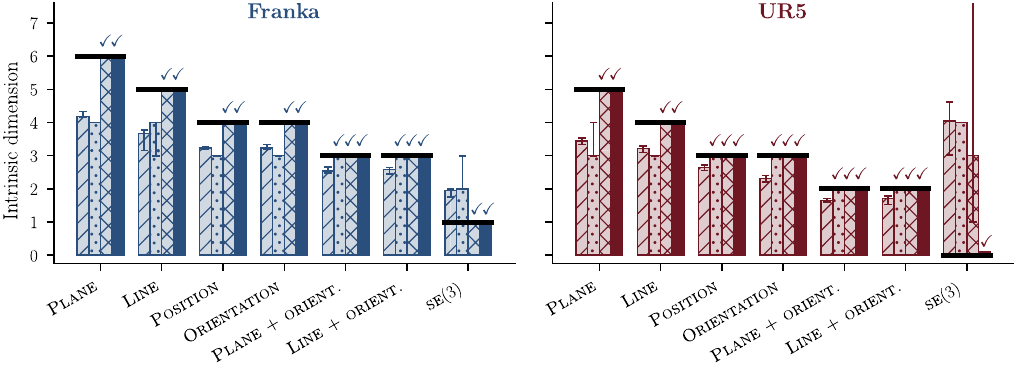}\\[-1mm]
\includegraphics[width=0.8\linewidth]{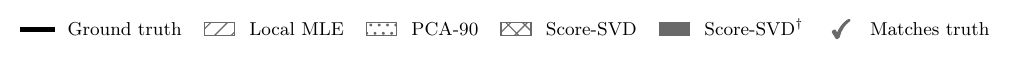}
\vspace{-4mm}
\caption{
\textbf{Intrinsic dimension estimation.}
We compare the ID computed by four estimators on seven task-space constraints per robot.
Bars show the per-target median with 95\% bootstrap CIs.
}
\vspace{-4mm}
\label{fig:id_estimation}
\end{figure}

\subsection{Recovering the intrinsic dimension across a family of constraints}
\label{sec:id}

If the diffusion model has learned the geometry of the constraint family, we would expect the ID recovered from its score function at a configuration $\mathbf{q} \in \mathcal{M}_{\mathbf{c}}$ to match the analytical intrinsic dimension $n-k$ for every conditioning input.
We test this hypothesis across the fourteen constraint--manipulator conditions; for each robot, we train a single conditional diffusion model on all seven constraints of Table~\ref{tab:constraints}.
For each family, we sample task-space targets and compute their analytical IK solutions as training data; the conditioning vector $\mathbf{c}$ encodes both the constraint type and the target.
Full training details are given in Appendix~\ref{app:training}.

We estimate the intrinsic dimension of $\mathcal{M}_{\mathbf{c}}$ using the Score-SVD procedure of Section~\ref{sec:background}~\cite{stanczuk2024diffusion}.
Figure~\ref{fig:id_estimation} shows the resulting estimates for every constraint--manipulator condition, together with two neighbourhood-based baselines (local MLE~\cite{levina2004maximum} and PCA with a 90\% variance threshold~\cite{fukunaga1971algorithm}) and the analytical ID as ground truth.
On both robots, the Score-SVD estimate matches the analytical ID for every condition with $\textrm{ID} > 0$.
Both baselines underestimate the analytical dimension by a gap that widens as the ID grows, with PCA only matching the true value at the smallest dimensions. 
On the two $\LieGroupSE{3}$ conditions, both baselines overestimate the ID.

The SE(3) condition for the UR5 is a special case: $\mathcal{M}_{\mathbf{c}}$ collapses to a finite set of isolated IK branches with $\textrm{ID} = 0$.
Equivalently, the tangent space is zero-dimensional, so every nonzero ambient direction is normal.
In this case, the standard Score-SVD rule cannot detect the relevant spectral gap, as Score-SVD assumes $\mathcal{M}_{\mathbf{c}}$ is positive dimensional.
We therefore modify the estimator to recover the analytical $\textrm{ID}=0$ of this discrete case.
Our variant, Score-SVD$^{\dagger}$, appends a virtual zero singular value $\sigma_{n+1}=0$ to the descending spectrum: when the smallest singular value $\sigma_n$ is well separated from this zero value, the argmax in~\eqref{eq:score_svd_rule} selects position $n$, returning $\hat{k}=n$ and hence $\hat d=0$ (implementation details of Score-SVD$^{\dagger}$ can be found in Appendix~\ref{app:discrete}).

\begin{figure}[t]
\centering
\includegraphics[width=0.9\linewidth]{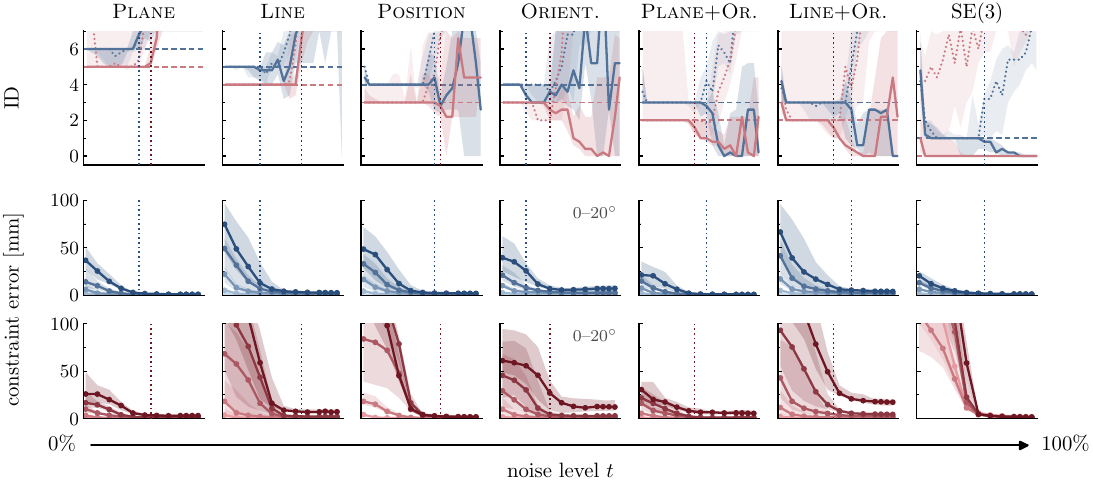}
\vspace{-2mm}
\caption{\textbf{Linear interpolation in latent space for \textcolor{frankaBlue}{Franka} and \textcolor{ur5Red}{UR5}.}
\emph{Top}: Estimated intrinsic dimension versus noise $t$; 
\@ \linedashed{scoreBlue} / \linedashed{scoreRed}~Ground truth;
\@ \linedotted{scoreBlue} / \linedotted{scoreRed}~Score-SVD;
\@ \linethick{dagBlue} / \linethick{dagRed}~Score-SVD$^{\dagger}$.
\emph{Bottom}: Constraint error of interpolated samples (position error, except orientation error on a $0$--$20^\circ$ scale for \textsc{Orientation}); darker colours indicate IK interpolation pairs that are further away from each other. For each robot, dotted vertical lines indicate the lowest noise level $t \geq 15$ at which the median Score-SVD$^{\dagger}$ estimate departs from the analytical ID, which coincides with the interpolation error converging towards zero on every constraint except \textsc{Orientation}.
}
\label{fig:error-svd-tmid-lineplots-main}
\end{figure}

\begin{figure}[t]
\centering
\includegraphics[width=0.95\linewidth]{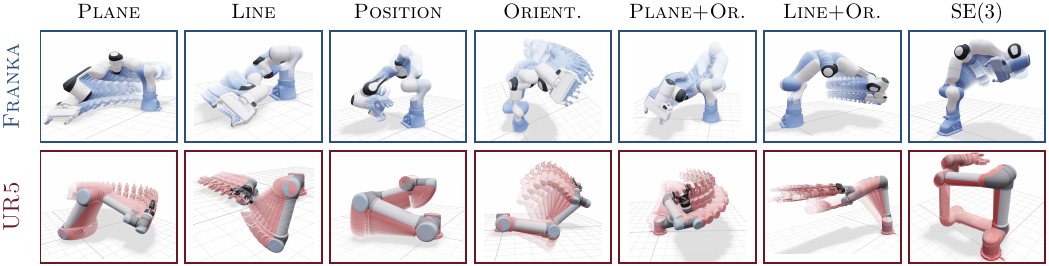}
\vspace{-1.5mm}
\caption{
\textbf{Visualization of linear interpolation between IK solutions in latent space.}  End-effector poses of the generated configurations are plotted in task space; opacity encodes the interpolation parameter, fading smoothly from one endpoint (fully opaque) to the other (fully transparent). The generated trajectory remains close to $\mathcal{M}_{\mathbf{c}}$ across constraint types and both robots.
}
\vspace{-3mm}
\label{fig:interpolations-compact}
\end{figure}

\subsection{Linear latent interpolation stays close to the constraint manifold}
\label{sec:interpolation}

In Section~\ref{sec:id}, we show that the intrinsic dimensionality of the latent space matches the analytical degrees of freedom of the constraint manifold.
However, matching dimensionality alone does not fully explain the underlying geometry of the learned representation.
A latent space could be dimensionally accurate yet geometrically curved, such that straight-line paths in latent coordinates decode into trajectories that leave $\mathcal{M}_{\mathbf{c}}$.
We therefore investigate whether the diffusion model learns a nonlinear mapping that locally linearizes the constraint manifold, translating curves on~$\mathcal{M}_{\mathbf{c}}$ into straight lines in latent space.
Specifically, we evaluate the extent to which linear interpolations in the DDIM-inverted latent space remain `on-manifold' when decoded.
Unlike image-based evaluations, our setting admits an exact, analytical test: given a generated configuration~$\mathbf{q}$, we measure the forward kinematics residual, i.e., the distance from the end-effector pose to the task-space constraint set~$\mathcal{S}_\mathbf{c}$.
Constraint satisfaction, or adherence to the data manifold, can thus be evaluated quantitatively using constrained IK, as opposed to the image-based setting, where qualitative perception plays a larger role.

\looseness=-1
For each of the fourteen constraint--manipulator conditions, we sample 30 task-space targets.
At each target, we select five pairs of IK solutions of increasing joint-space separation (Appendix~\ref{app:latent-interpolation}); the widest pair represents an extreme boundary case.
For the UR5 under the full $\LieGroupSE{3}$ constraint, the widest pairs may connect IK branches that are disconnected in joint space, with no continuous pose-preserving path between them.
In contrast, for the Franka, the widest pairs span the full extent of a continuous self-motion manifold.

\paragraph{Latent interpolation between IK solutions.}

For each IK pair, we invert both configurations to an intermediate noise level of the partial DDIM process.
We then linearly interpolate their latent codes and decode the interpolants to joint configurations.
Across all constraints, the decoded joint-space paths remain close to the corresponding constraint manifolds.
The average position constraint error, measured by forward kinematics, ranges from sub-millimetre to a few millimetres; rotational errors for orientation-constrained tasks are reported in Appendix~\ref{app:interp-quantitative}.

\looseness-1
Figures~\ref{fig:error-svd-tmid-lineplots-main} and~\ref{fig:interpolations-compact} show linear interpolation results for all fourteen constraint--manipulator conditions, ordered from high to low ID.
We observe that interpolation errors are larger at lower noise levels and decrease toward zero at higher noise levels.
The errors also decrease as the joint-space separation between IK solutions becomes smaller.
The dotted line in Figure~\ref{fig:error-svd-tmid-lineplots-main} indicates the noise level at which the estimated ID begins to diverge from the analytical ID, coinciding with the level at which the interpolation error drops to near zero across all pairs for every constraint except \textsc{Orientation}.
\emph{Linear paths in latent space thus appear to remain close to the constraint manifold only once the
intermediate-noise representation expands beyond the analytical ID. This suggests that the DDIM
representation locally flattens the constraint geometry along interpolation directions, allowing simple
linear paths in latent space to decode into motions that remain close to $\mathcal M_{\mathbf c}$.}
This behaviour is not what a Gaussian approximation of the solution set would produce.
With a Gaussian score, DDIM inversion and decoding are affine maps, so a straight latent path decodes to a straight line in the sine--cosine encoding whichever Gaussian is fitted; the median constraint error of that path is larger than the learned model's by a median factor of $12$ in position and $7$ in orientation across conditions (Appendix~\ref{app:interp-extended}).

For the UR5 under the full $\LieGroupSE{3}$ constraint, the feasible configurations form a finite set of disconnected IK branches.
Any continuous joint-space path between two branches must therefore leave the constraint manifold along its interior.
We observe that the decoded latent interpolation is consistent with this behaviour: constraint error stays small along the decoded samples, while the joint-space trajectory makes a sharp transition when switching between branches.
This suggests that the latent interpolation is consistent with the disconnected topology of the constraint set rather than artificially constructing an infeasible continuous bridge in joint space.

\begin{table}[tbh]
\centering
\footnotesize
\setlength{\tabcolsep}{4pt}
\caption{\textbf{Median joint-space sensitivity and tangent overlap along latent interpolation paths.}
$\tau^{\mathbf{q}}_{\rm rand,med}$ and $\tau^{\mathbf{q}}_{\rm rand,p5}$ are the median and fifth percentile of the overlap of a uniformly random direction, and $\smash{\Delta^{\mathbf{q}}_{\rm rel}}$ compares the two medians.}
  \label{tab:sensitivity}
  \begin{tabular}{@{}llrrrrrrr@{}}
    \toprule
    & & \multicolumn{2}{c}{Sensitivity} & \multicolumn{5}{c}{Tangent overlap} \\
    \cmidrule(lr){3-4}
    \cmidrule(l){5-9}
    & Constraint
        & $\rho_{\rm med}$
        & $\nu_{\rm med}$
        & $\tau^{\mathbf{q}}_{\rm med}$
        & $\tau^{\mathbf{q}}_{\rm p5}$
        & $\tau^{\mathbf{q}}_{\rm rand,med}$
        & $\tau^{\mathbf{q}}_{\rm rand,p5}$
        & $\Delta^{\mathbf{q}}_{\rm rel}$ \\
    \midrule
\multirow{7}{*}{\rotatebox[origin=c]{90}{\textsc{Franka}}} & \textsc{Plane} & \tclowrho{0.551} & 1.24 & 1.000 & \tcabove{1.000} & 0.960 & 0.707 & \tcabove{+4\%} \\
 & \textsc{Line} & \tclowrho{0.463} & 1.13 & 1.000 & \tcabove{0.999} & 0.871 & 0.549 & \tcabove{+15\%} \\
 & \textsc{Position} & \tclowrho{0.565} & 0.95 & 1.000 & \tcabove{1.000} & 0.766 & 0.410 & \tcabove{+31\%} \\
 & \textsc{Orientation} & \tclowrho{0.451} & 1.06 & 1.000 & \tcabove{0.996} & 0.766 & 0.410 & \tcabove{+31\%} \\
 & \textsc{Plane} $+$ \textsc{Orientation} & \tclowrho{0.515} & 1.16 & 1.000 & \tcabove{0.999} & 0.643 & 0.276 & \tcabove{+55\%} \\
 & \textsc{Line} $+$ \textsc{Orientation} & \tclowrho{0.406} & 0.88 & 1.000 & \tcabove{0.983} & 0.643 & 0.276 & \tcabove{+55\%} \\
 & $\LieGroupSE{3}$ & \tclowrho{0.658} & 1.10 & 1.000 & \tcabove{0.998} & 0.281 & 0.027 & \tcabove{+256\%} \\
\midrule
\multirow{7}{*}{\rotatebox[origin=c]{90}{\textsc{UR5}}} & \textsc{Plane} & \tclowrho{0.645} & 1.72 & 1.000 & \tcabove{0.999} & 0.951 & 0.656 & \tcabove{+5\%} \\
 & \textsc{Line} & \tclowrho{0.592} & 1.63 & 1.000 & \tcabove{0.998} & 0.841 & 0.473 & \tcabove{+19\%} \\
 & \textsc{Position} & \tclowrho{0.701} & 1.55 & 1.000 & \tcabove{0.999} & 0.707 & 0.312 & \tcabove{+41\%} \\
 & \textsc{Orientation} & \tclowrho{0.590} & 1.66 & 1.000 & \tcabove{0.990} & 0.707 & 0.312 & \tcabove{+41\%} \\
 & \textsc{Plane} $+$ \textsc{Orientation} & \tclowrho{0.607} & 1.57 & 1.000 & \tcabove{0.986} & 0.541 & 0.159 & \tcabove{+85\%} \\
 & \textsc{Line} $+$ \textsc{Orientation} & \tclowrho{0.348} & 1.16 & 0.998 & \tcabove{0.754} & 0.541 & 0.159 & \tcabove{+84\%} \\
 & $\LieGroupSE{3}$ & \tclowrho{0.010} & 0.03 & 0.000 & \tcnear{0.000} & 0.000 & 0.000 & -- \\
    \bottomrule
  \end{tabular}
\vspace{-2mm}
\end{table}

\paragraph{Latent interpolation is consistent with local flattening of constraint geometry.}

\looseness=-1
The interpolation results show that the model does more than recover the ID of each constraint manifold;
it generates joint-space motions that are largely consistent with the local constraint geometry.
To test this, we examine the map $g_{\mathbf c}$ along the latent interpolation path $\mathbf{x}(\alpha)=(1-\alpha)\,\mathbf{x}_a + \alpha \, \mathbf{x}_b, \alpha \in \lbrack 0,1 \rbrack$, where subscripts $a$ and $b$ denote the two interpolation endpoints at a fixed noise level $t$, with unit direction $\widehat{\mathbf{v}}=(\mathbf{x}_b-\mathbf{x}_a)/\|\mathbf{x}_b-\mathbf{x}_a\|$.
At each point on this path, we measure how much the generated configuration $\mathbf{q}(\alpha)=g_{\mathbf c}(\mathbf{x}(\alpha))$ changes under an infinitesimal step along $\widehat{\mathbf{v}}$, giving the directional sensitivity
\begin{equation}
\rho(\alpha)
=
\left\| Dg_{\mathbf c}(\mathbf x(\alpha))\,\widehat{\mathbf v} \right\|_2.
\end{equation}
Because $\mathbf{q}$ is expressed in raw joint angles, $\rho$ has units of radians per latent unit; small values mean a unit latent step produces only a small change in joint space.
We also report the corresponding joint-space speed with respect to the interpolation parameter,
\begin{equation}
\nu(\alpha)
=
\left\| Dg_{\mathbf c}(\mathbf{x}(\alpha))(\mathbf{x}_b-\mathbf{x}_a) \right\|_2,
\end{equation}
which measures how quickly $\mathbf{q}$ changes as $\alpha$ moves from one endpoint to the other.
Across all conditions, the median directional sensitivity $\rho_{\rm med}$ stays below $1$ radian per latent unit, with an overall median of $0.558$.
The per-condition median of $\nu$ ranges from $0.03$ to $1.72$ radians per unit interpolation parameter, with an overall median of $1.16$ across both manipulators (Table~\ref{tab:sensitivity}).
Notably, both the sensitivity and the speed for the UR5 under $\LieGroupSE{3}$ remain close to zero, indicating that small movements in the latent space stay within a single discrete branch.

Small joint-space motion alone does not imply that the interpolation respects the constraint.
We therefore examine whether the motion of the decoded configuration preserves the constraint error to first order.
Let $h_{\mathbf c}$ denote the constraint error map, with $\mathcal M_{\mathbf c}=\{\mathbf{q}:h_{\mathbf c}(\mathbf{q})=0\}$,
and let $\mathbf{u}(\alpha)=Dg_{\mathbf c}(\mathbf{x}(\alpha))\,\widehat{\mathbf{v}}$ be the decoded joint-space direction at $\mathbf{q}(\alpha)=g_{\mathbf c}(\mathbf{x}(\alpha))$.
We measure the fraction of $\mathbf u(\alpha)$ lying in the null space of the constraint Jacobian,
\begin{equation}
\tau^{\mathbf q}(\alpha)
=
\frac{
\|\Pi_{\ker Dh_{\mathbf c}(\mathbf{q}(\alpha))}\mathbf{u}(\alpha)\|_2
}{
\|\mathbf{u}(\alpha)\|_2
},
\end{equation}
where $\Pi_{\ker Dh_{\mathbf c}(\mathbf{q}(\alpha))}$ denotes orthogonal projection onto that null space.
Intuitively, $\tau^{\mathbf{q}}$ measures how much of the generated configuration moves \emph{along} the constraint manifold rather than \emph{away} from it.
A value near one means the direction lies in the constraint tangent space, leaving any existing constraint error unchanged; when $\mathbf{q}(\alpha)$ lies on $\mathcal{M}_{\mathbf{c}}$, this null space coincides with $\mathcal{T}_{\mathbf{q}(\alpha)}\mathcal{M}_{\mathbf{c}}$.
For comparison, the squared projection of a uniformly random unit direction in $\mathbb{R}^n$ onto an $(n-k)$-dimensional tangent space follows a $\mathrm{Beta}\big(\tfrac{n-k}{2},\tfrac{k}{2}\big)$ distribution, and we compare $\tau^{\mathbf{q}}_{\rm med}$ and $\tau^{\mathbf{q}}_{\rm p5}$ with the median $\tau^{\mathbf{q}}_{\rm rand,med}$ and fifth percentile $\tau^{\mathbf{q}}_{\rm rand,p5}$ of its square root.
We defer full definitions and derivations of $\rho$, $\nu$, and $\tau^{\mathbf{q}}$ to Appendix~\ref{app:interp-definitions}.

\looseness-1
Empirically, $\mathbf{u}(\cdot)$ is small and aligns closely with the analytical tangent space of $\mathcal{M}_{\mathbf{c}}$,
as measured respectively by $\rho = \Norm{\mathbf{u}(\cdot)}$ and $\tau^{\mathbf{q}}$ in Table~\ref{tab:sensitivity}.
The median tangent overlap is $\tau^{\mathbf{q}}_{\rm med}\approx1.00$ across both manipulators, and in all thirteen conditions with $\textrm{ID} > 0$ both $\tau^{\mathbf{q}}_{\rm med}$ and $\tau^{\mathbf{q}}_{\rm p5}$ exceed the corresponding quantile of the random-direction baseline.
The UR5 under SE(3) is excluded since its tangent space is zero-dimensional, making the tangent overlap zero by construction, consistent with the disconnected-branch behaviour above.
The lower tail is weakest for the UR5 under \textsc{Line}~$+$~\textsc{Orientation}, where $\tau^{\mathbf{q}}_{\rm p5}=0.754$ remains well above the baseline value of $0.159$.
Since the tangent space is evaluated at the sample $\mathbf{q}(\alpha)$, which need not lie on $\mathcal{M}_{\mathbf{c}}$, $\tau^{\mathbf{q}}\approx1$ indicates that the generated motion stays tangent to $\mathcal{M}_{\mathbf{c}}$ at a fixed offset, preserving whatever constraint error is already present.
Together, the small $\rho$ and the large $\tau^{\mathbf{q}}$ suggest that linear interpolation in the latent space succeeds because it produces joint-space motion that changes slowly and largely follows the local linearized constraint geometry of $\mathcal{M}_{\mathbf{c}}$, even when the generated configurations are not exactly feasible.

%% file: sections/applications.tex
\vspace{-1mm}
\section{Applications of the Recovered Geometry}
\label{sec:applications}

\looseness=-1
The score encodes the dimension and local geometry of $\mathcal{M}_{\mathbf{c}}$; we now leverage that geometry in two ways.
We use the normal basis implied by Score-SVD to pull generated configurations onto the learned manifold without an analytical Jacobian in Section~\ref{sec:projection}, and compare the dimension estimate at seen and unseen configurations to verify whether a model has been trained on enough data in Section~\ref{sec:convergence}.

\vspace{-1mm}
\subsection{Score projection onto the estimated normal bundle}
\label{sec:projection}
The singular vectors that Score-SVD uses to estimate the dimension of $\mathcal{M}_{\mathbf{c}}$ in Section~\ref{sec:id} also provide a basis of its normal space, which lets us project a configuration onto the learned manifold without an analytical Jacobian.
Starting from a configuration $\mathbf{x}_0$ generated by the model, we iterate
\begin{align}
  \mathbf{x}_0 &\;\leftarrow\; \mathbf{x}_0 \;+\; \gamma\,\frac{1-\bar\alpha_t}{\sqrt{\bar\alpha_t}}\;
      U_{\hat{k}} U_{\hat{k}}^{\T}\,
      s_\theta\!\left(\sqrt{\bar\alpha_t}\,\mathbf{x}_0,\, t,\, \mathbf{c}\right),
      \label{eq:score-svd-update}
\end{align}
at a fixed noise level $t$ and step size $\gamma$, where $U_{\hat{k}}$ collects the
left singular vectors of the stacked score matrix $S$ paired with its $\smash{\hat{k}}$ largest singular values.
The scaled score in~\eqref{eq:score-svd-update} equals $\hat{\mathbf{x}}_0 - \mathbf{x}_0$, the displacement to the denoised estimate of~\eqref{eq:ddim_step}, and $U_{\hat{k}} U_{\hat{k}}^{\T}$ retains
only its component normal to $\mathcal{M}_{\mathbf{c}}$.
The update thus mirrors a Newton--Raphson IK step $\smash{\Delta\mathbf{q} = -Dh_{\mathbf{c}}^{+}\,h_{\mathbf{c}}(\mathbf{q})}$, where the learned score approximates the error, the pseudo-inverse, and the projector onto the constrained directions, without the need for an analytical Jacobian.

We apply $20$ iterations to $120$ generated configurations for each of the fourteen conditions and measure the constraint error $\Norm{h_{\mathbf{c}}(\mathbf{q})}$, which enters the evaluation only.
The median error decreases on every condition, by factors of $2.6$--$6.8$ on the UR5 and $1.5$--$5.9$ on the Franka, and settles within about two iterations.
The unprojected score also lowers the error by a similar amount, but it keeps moving the configuration along $\mathcal{M}_{\mathbf{c}}$ to other solutions. With the projection, the configuration stays close to where it started.
The algorithm and per-condition results are given in Appendix~\ref{app:applications}.

\vspace{-1mm}
\subsection{Intrinsic dimensionality as a training convergence diagnostic}
\label{sec:convergence}

\begin{figure}[t]
\centering
\includegraphics[width=0.88\linewidth]{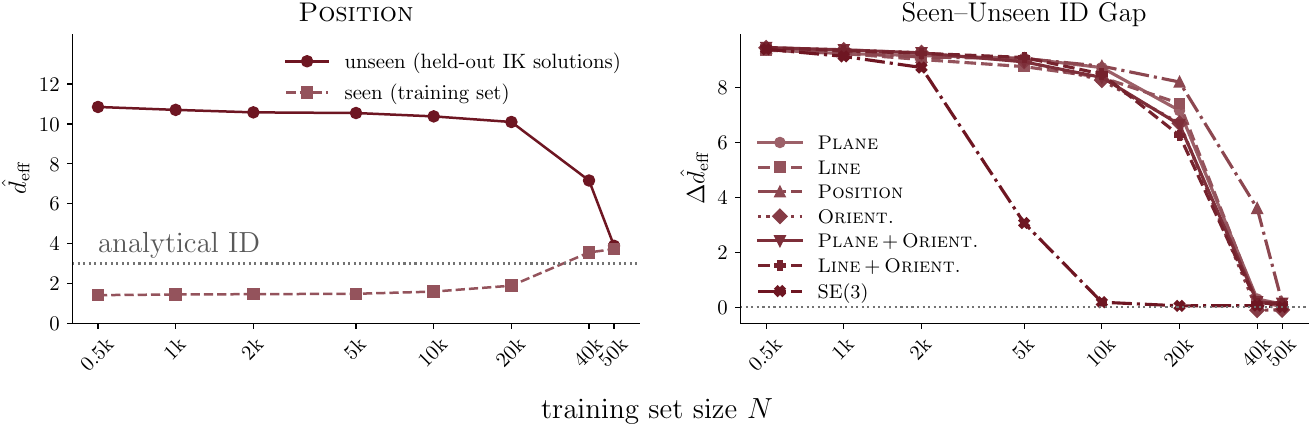}
\vspace{-2mm}
\caption{
\textbf{The ID gap between seen and unseen configurations closes as data grows (UR5).}
The ID is $\smash{\hat{d}_{\mathrm{eff}}}$, obtained from the effective rank of $S$ instead of the spectral-gap rule of Algorithm~\ref{alg:score-svd-dagger}.
\emph{Left}: \textsc{Position}, with both estimates meeting near the analytical ID (dotted).
\emph{Right}: difference between the two estimates on all seven constraints.
}
\vspace{-4mm}
\label{fig:idgap}
\end{figure}

\looseness=-1
The ID estimate can be evaluated at any configuration, so the geometry a model has learned at configurations in the training set can be compared with the geometry it assigns to held-out configurations.
We train one UR5 model per constraint on $N = 500$ to $50{,}000$ configurations and compute $S$ at \emph{seen} configurations from the training set and at \emph{unseen}, held-out analytical IK solutions.
To follow convergence continuously, we replace the integer rule~\eqref{eq:score_svd_rule} with the effective rank of $S$~\cite{roy2007effective}, giving $\smash{\hat{d}_{\mathrm{eff}} = n - \exp\!\big(-\textstyle\sum_i p_i \log p_i\big)}$ with $\smash{p_i = \sigma_i^2 / \sum_j \sigma_j^2}$. 
With low amounts of data the two estimates disagree: at $N = 500$, $\hat{d}_{\mathrm{eff}}$ lies between $10.8$ and $10.9$ at unseen configurations, close to the ambient dimension $n = 12$, and equals $1.4$ at seen ones on every constraint, as expected of a model that has memorized its training points.
As $N$ grows, the two approach the analytical ID from opposite sides (Figure~\ref{fig:idgap}), and at $N = 50{,}000$, their difference is below $0.3$ on all seven constraints.
The gap therefore indicates generalization to unseen samples and training convergence, and computing it does not require knowledge of the true intrinsic dimension. The gap can therefore be checked before a model is deployed.
On the Franka, the gap closes on four constraints and is still decreasing on the other three (Appendix~\ref{app:applications}).

%% file: sections/discussion.tex
\vspace{-1.5mm}
\section{Discussion and Limitations}

We studied the latent-space geometry of conditional diffusion models trained across seven task-space constraints on two manipulators, using the analytically known constraint manifold as ground truth.
The intrinsic dimension recovered from the model's score function matches the analytical degrees of freedom for every positive-dimensional condition, and a small modification of the estimator extends this to the degenerate case in which the solution set collapses to a finite collection of disconnected IK branches.
Beyond dimension, linear interpolation in the partial-DDIM latent space decodes into joint-space paths that stay close to the constraint manifold and move along directions aligned with its tangent space. Together these results indicate that the model captures aspects of each manifold's local geometry, not only its dimension, in a setting where that geometry is known rather than estimated.

\paragraph{Limitations.}
\looseness=-1
Our analysis treats each constraint condition as having a single, fixed dimension and tangent direction, whereas these can change near joint limits and singular configurations; handling that variation is left to future work.
Because we use straight-line paths in latent space, accuracy drops as the two interpolated solutions grow farther apart, where the manifold curves too much for a linear path to follow; geodesic alternatives are a natural next step.
Downstream applications for learned policies based on diffusion, such as measuring intrinsic dimensionality for dataset aggregation or latent interpolation for behaviour editing, are also left as extensions.
Finally, our empirical scope covers only two manipulators, seven constraint families, and one model per robot, leaving more complex robots and other generative models open for analysis.
%

%% file: sections/appendix.tex
\input{sections/appendix/notation}
\input{sections/appendix/mathematical-background}
\input{sections/appendix/dataset-training}
\input{sections/appendix/id-estimation}
\input{sections/appendix/latent-interpolation}
\input{sections/appendix/applications}

%% file: sections/appendix/notation.tex
\section{Notation}
\label{app:notation}

Table~\ref{tab:notation} collects the notation used throughout the paper.

\begin{longtable}{@{}l p{3.5in}@{}}
\caption{Summary of notation used throughout the paper.}\label{tab:notation}\\
\toprule
\textbf{Symbol} & \textbf{Meaning} \\
\midrule
\endfirsthead
\multicolumn{2}{c}{\tablename\ \thetable{} -- continued from previous page}\\
\toprule
\textbf{Symbol} & \textbf{Meaning} \\
\midrule
\endhead
\midrule
\multicolumn{2}{r}{\emph{continued on next page}}\\
\endfoot
\bottomrule
\endlastfoot

% ---- Group 1: Manipulator and kinematics ----
\multicolumn{2}{@{}l}{\textbf{Manipulator and kinematics}}\\
\addlinespace[2pt]
$\mathbf{q}$ & joint configuration of the manipulator \\
$n$ & number of actuated joints ($6$ for the UR5, $7$ for the Franka); the same symbol also denotes the dimension of the sine--cosine score space ($12$ and $14$) on which Score-SVD acts \\
$f : \Real{}^n \to \LieGroupSE{3}$ & forward kinematics map of the manipulator \\
$\LieGroupSE{3},\ \LieGroupSO{3}$ & special Euclidean and special orthogonal groups \\
$\mathbf{p},\ \Matrix{R}$ & end-effector position and orientation, $f(\mathbf{q}) = (\mathbf{p}(\mathbf{q}), \Matrix{R}(\mathbf{q}))$ \\
$p_x,\ p_y,\ p_z$ & end-effector position coordinates \\
$\mathbf{e}_x,\ \mathbf{e}_y,\ \mathbf{e}_z$ & end-effector body axes \\
$d(\Matrix{R}, \Matrix{R}^\star)$ & geodesic angle between two orientations on $\LieGroupSO{3}$ \\
$\mathbf{c}$ & conditioning input: constraint family and target \\
$\mathbf{p}^\star,\ \Matrix{R}^\star$ & target end-effector position and orientation \\
$\mathbf{m}$ & selection vector $(m_1,\dots,m_6) \in \Set{0,1}^{6}$ marking the constrained coordinates ($k = \Norm{\mathbf{m}}_1$) \\
$\mathcal{S}_\mathbf{c}$ & task-space constraint set in $\LieGroupSE{3}$ \\
$\mathcal{M}_\mathbf{c}$ & constraint manifold in configuration space \\
$k$ & codimension; number of fixed task-space coordinates \\
$n-k$ & intrinsic dimension of the constraint manifold \\
$\mathrm{ID}$ & intrinsic dimension $n-k$; abbreviation used in tables and figures \\
$h_\mathbf{c}$ & constraint error map; zero set is $\mathcal{M}_\mathbf{c}$ \\
$Dh_\mathbf{c}$ & Jacobian of the constraint map \\ % ADDITION (not in requested list)
$\mathcal{T}_{\mathbf{q}}\mathcal{M}_\mathbf{c}$ & tangent space of $\mathcal{M}_\mathbf{c}$ at a configuration \\
$\mathcal{N}_{\mathbf{x}_0}\mathcal{M}_\mathbf{c}$ & normal space of $\mathcal{M}_\mathbf{c}$ at a sample \\
$\Pi_{\ker Dh_\mathbf{c}}$ & orthogonal projector onto the tangent space \\

\addlinespace
% ---- Group 2: Diffusion model ----
\multicolumn{2}{@{}l}{\textbf{Diffusion model}}\\
\addlinespace[2pt]
$\mathbf{x}_0,\ \mathbf{x}_t$ & clean and noised samples at time $t$ \\
$\mathcal{X}$ & latent (noise) space of the diffusion model \\ % ADDITION (not in requested list)
$t,\ T$ & diffusion timestep and maximum timestep \\
$\bar\alpha_t$ & variance schedule at time $t$ \\
$\boldsymbol{\epsilon}$ & injected Gaussian noise \\
$\hat{\boldsymbol{\epsilon}}_\theta$ & denoising network with parameters $\theta$ \\
$s_\theta$ & score of the conditional noised marginal \\
$p_t(\mathbf{x}_t \mid \mathbf{c})$ & conditional noised marginal density \\
$\hat{\mathbf{x}}_0$ & predicted clean sample from a noised one \\
$g_{\mathbf{c}},\ g_{\mathbf{c}}^{-1}$ & DDIM decoder and its inverse (inversion) \\
$\bar{t}$ & intermediate noise level for partial inversion \\
$N$ & number of training configurations (dataset size) \\

\addlinespace
% ---- Group 3: Dataset and training ----
\multicolumn{2}{@{}l}{\textbf{Dataset and training}}\\
\addlinespace[2pt]
$\mathbf{q}_{\mathrm{rand}}$ & random joint configuration used to sample reachable target poses \\
$r,\ r_{\min},\ r_{\max}$ & cylindrical workspace radius $r=\sqrt{x^2+y^2}$ and its bounds \\
$z_{\min},\ z_{\max}$ & workspace height bounds \\
$\phi(\mathbf{q})$ & sine--cosine joint encoding $(\sin\mathbf{q},\,\cos\mathbf{q})$, of dimension $2n$ \\
$(\hat{\mathbf{a}},\,\hat{\mathbf{b}})$ & model estimates of $(\sin\mathbf{q},\,\cos\mathbf{q})$ at inference \\
$\beta,\ \beta_{\min},\ \beta_{\max}$ & per-step forward-process noise variance (sigmoid schedule) and its range \\
$\eta$ & DDIM stochasticity parameter ($\eta = 0$, deterministic) \\

\addlinespace
% ---- Group 4: Intrinsic dimension estimation ----
\multicolumn{2}{@{}l}{\textbf{Intrinsic dimension estimation}}\\
\addlinespace[2pt]
$K$ & number of score perturbations \\
$K_s$ & number of analytical solutions per target at which Score-SVD is evaluated ($K_s = 5$) \\
$S$ & stacked score matrix, $S \in \Real{}^{n \times K}$ \\
$\sigma_i$ & singular values of $S$, in descending order \\
$\hat{d},\ \hat{k}$ & estimated intrinsic dimension and codimension \\
$\sigma_{n+1}$ & virtual zero singular value in Score-SVD$^{\dagger}$ \\
$U_{\hat{k}}$ & left singular vectors of $S$ paired with its $\hat{k}$ largest singular values; basis of the estimated normal space \\
$\hat{d}_{\mathrm{eff}}$ & effective-rank intrinsic dimension, $n - \exp(-\sum_i p_i \log p_i)$ \\
$p_i$ & normalized squared singular values, $\sigma_i^2 / \sum_j \sigma_j^2$ \\
$\mathrm{NN}$ & neighbourhood size of the local MLE and PCA baselines \\

\addlinespace
% ---- Group 4b: Score projection ----
\multicolumn{2}{@{}l}{\textbf{Score projection (Section~\ref{sec:projection})}}\\
\addlinespace[2pt]
$\gamma$ & step size of the projection update~\eqref{eq:score-svd-update} ($\gamma = 1$) \\
$M$ & number of projection iterations ($M = 20$; $200$ in the long run) \\
$R$ & refresh period of the normal basis $U_{\hat{k}}$ ($R = 5$) \\
$\delta$ & task-space displacement of exact IK solutions in the re-projection study \\
$Dh_{\mathbf{c}}^{+}$ & pseudo-inverse of the constraint Jacobian \\

\addlinespace
% ---- Group 5: Latent interpolation and local flattening ----
\multicolumn{2}{@{}l}{\textbf{Latent interpolation and local flattening}}\\
\addlinespace[2pt]
$\mathbf{x}_a,\ \mathbf{x}_b$ & latent-space interpolation endpoints \\
$\alpha$ & interpolation parameter in $[0,1]$ \\
$\mathbf{x}(\alpha)$ & linear interpolation path in latent space \\
$\mathbf{q}(\alpha)$ & decoded joint configuration along the path, $g_{\mathbf{c}}(\mathbf{x}(\alpha))$ \\
$\widehat{\mathbf{v}}$ & unit direction between latent endpoints \\
$Dg_{\mathbf{c}}$ & Jacobian of the DDIM decoder, $Dg_{\mathbf{c}} \in \Real{}^{n \times 2n}$ \\
$\rho(\alpha)$ & decoder directional sensitivity \\
$\nu(\alpha)$ & joint-space speed along the path \\
$\mathbf{u}(\alpha)$ & decoded joint-space direction \\
$\tau^{\mathbf{q}}(\alpha)$ & tangent overlap of the decoded direction \\
$\tau^{\mathbf{q}}_{\rm rand,med},\ \tau^{\mathbf{q}}_{\rm rand,p5}$ & median and fifth percentile of the tangent overlap $\Norm{\Pi_{\ker Dh_\mathbf{c}}\,\mathbf{r}}_2$ of a uniformly random unit direction $\mathbf{r}$; random-direction baseline \\
$\mathbf{r}$ & random unit vector on the sphere $S^{n-1}$ \\
$S^{n-1}$ & unit sphere in joint space \\
$\rho_{\rm med},\ \nu_{\rm med},\ \tau^{\mathbf{q}}_{\rm med}$ & median sensitivity, speed, and tangent overlap \\
$\tau^{\mathbf{q}}_{\rm p5}$ & 5th-percentile tangent overlap (lower tail) \\
$\Delta^{\mathbf{q}}_{\rm rel}$ & relative tangent-overlap improvement over baseline \\
lin,\ sph & straight-line and spherical (slerp) latent interpolation \\

\addlinespace
% ---- Group 6: Constraint families (Table~\ref{tab:constraints}) ----
\multicolumn{2}{@{}l}{\textbf{Constraint families (Table~\ref{tab:constraints})}}\\
\addlinespace[2pt]
\textsc{Plane} & one position coordinate fixed ($k=1$) \\
\textsc{Line} & two position coordinates fixed ($k=2$) \\
\textsc{Position} & position fixed, orientation free ($k=3$) \\
\textsc{Orientation} & orientation fixed, position free ($k=3$) \\
\textsc{Plane} $+$ \textsc{Orientation} & one position coordinate and full rotation ($k=4$) \\
\textsc{Line} $+$ \textsc{Orientation} & two position and two rotation coordinates ($k=4$) \\
$\LieGroupSE{3}$ & full end-effector pose fixed ($k=6$) \\

\end{longtable}

%% file: sections/appendix/mathematical-background.tex
\section{Mathematical Background}
\label{app:mathematical-background}

This section provides the geometric and kinematic preliminaries underlying the analysis in our work.
We first review the smooth-manifold machinery used to equip the constraint manifold $\mathcal{M}_\mathbf{c}$ with a dimension, a tangent space, and a normal space (Appendix~\ref{app:smooth-manifolds}).
We then define the seven task-space constraints of Table~\ref{tab:constraints} precisely, specifying the task-space set $\mathcal{S}_\mathbf{c}$ and the constraint error map $h_\mathbf{c}$ for each (Appendix~\ref{app:constraints}).
Finally, we give the forward process, the DDIM update and its inversion that underlie the diffusion model of Section~\ref{sec:background} (Appendix~\ref{app:diffusion-background}).

\subsection{Smooth manifolds}
\label{app:smooth-manifolds}

Every constraint manifold in this paper is the solution set of a system of smooth equations, so we briefly recall the geometry of such sets~\cite{lee2012smooth}.
A smooth manifold is a topological space that is locally diffeomorphic to a Euclidean space.
A subset $\mathcal{M} \subseteq \Real{}^n$ is an \emph{embedded submanifold} of codimension $k$ if, near each of its points, it is the common zero set of $k$ \emph{local defining functions}, smooth scalar functions whose differentials are linearly independent, so that $\mathcal{M}$ inherits the smooth structure of the ambient space and has intrinsic dimension $n-k$.
Each constraint manifold we study takes exactly this form, so its dimension and tangent geometry follow directly from the defining functions.

\looseness=-1
Let $h_{\mathbf{c}} : \Real{}^n \to \Real{}^k$ be smooth with $k \le n$.
A value is \emph{regular} if the Jacobian $Dh_{\mathbf{c}}(\mathbf{q}) \in \Real{}^{k \times n}$ has full row rank $k$ at every $\mathbf{q}$ in its preimage.
When $\Zero$ is a regular value, the regular value theorem, a consequence of the implicit function theorem, makes the preimage
\begin{equation}
\mathcal{M} = \Set{\mathbf{q} \in \Real{}^n : h_{\mathbf{c}}(\mathbf{q}) = \Zero} = \Inv{h_{\mathbf{c}}}(\Zero)
\label{eq:app-levelset}
\end{equation}
an embedded submanifold of $\Real{}^n$ of dimension $n-k$~\cite{lee2012smooth}.
The codimension $k$ equals the number of independent scalar constraints imposed by $h_{\mathbf{c}}$.

At a point $\mathbf{q} \in \mathcal{M}$, the tangent space is the kernel of the Jacobian,
\begin{equation}
\mathcal{T}_{\mathbf{q}}\mathcal{M} = \ker Dh_{\mathbf{c}}(\mathbf{q}) \subseteq \Real{}^n,
\label{eq:app-tangent}
\end{equation}
a linear subspace of dimension $n-k$ collecting the velocities along which $h_{\mathbf{c}}$ is unchanged to first order.
Its orthogonal complement is the normal space $\mathcal{N}_{\mathbf{q}}\mathcal{M} = (\mathcal{T}_{\mathbf{q}}\mathcal{M})^{\perp}$, the $k$-dimensional row space of $Dh_{\mathbf{c}}(\mathbf{q})$, equivalently the image of $\Transpose{Dh_{\mathbf{c}}(\mathbf{q})}$.
The orthogonal projector onto the tangent space admits the closed form
\begin{equation}
\Pi_{\ker Dh_{\mathbf{c}}(\mathbf{q})} = \Identity - \Transpose{Dh_{\mathbf{c}}(\mathbf{q})}\Inv{\left(Dh_{\mathbf{c}}(\mathbf{q})\,\Transpose{Dh_{\mathbf{c}}(\mathbf{q})}\right)}Dh_{\mathbf{c}}(\mathbf{q}),
\label{eq:app-projector}
\end{equation}
which is well defined at every regular point, since $Dh_{\mathbf{c}}(\mathbf{q})\,\Transpose{Dh_{\mathbf{c}}(\mathbf{q})}$ is invertible.

We compare the dimension, the normal space, and the projector against the trained model.
The dimension $n-k$ is the analytical intrinsic dimension recovered by the score-based estimator of Section~\ref{sec:id}, the normal space $\mathcal{N}_{\mathbf{q}}\mathcal{M}$ is the subspace that the score spans in Score-SVD, and the projector~\eqref{eq:app-projector} defines the tangent overlap $\tau^{\mathbf{q}}$ of Section~\ref{sec:interpolation}.

\subsection{Precise constraint definitions}
\label{app:constraints}

We write the forward kinematics in components as $f(\mathbf{q}) = (\mathbf{p}(\mathbf{q}), \Matrix{R}(\mathbf{q}))$, with end-effector position $\mathbf{p}(\mathbf{q}) \in \Real{}^3$ and orientation $\Matrix{R}(\mathbf{q}) \in \LieGroupSO{3}$.
A constraint $\mathbf{c} = (\mathbf{p}^\star, \Matrix{R}^\star, \mathbf{m})$ pairs a target pose $(\mathbf{p}^\star, \Matrix{R}^\star) \in \LieGroupSE{3}$ with a \emph{selection vector}
\begin{equation*}
\mathbf{m} = (m_1, \dots, m_6) \in \Set{0,1}^{6},
\end{equation*}
whose entries indicate which task-space coordinates are constrained: $m_i = 1$ holds the $i$-th coordinate at its target value, while $m_i = 0$ leaves it free.
The first three entries correspond to the position coordinates $p_x, p_y, p_z$ and the last three to the end-effector's body axes $\mathbf{e}_x, \mathbf{e}_y, \mathbf{e}_z$.
Varying $\mathbf{m}$ thus yields the seven constraints in our family; how $\mathbf{m}$ enters the conditioning vector is described in Appendix~\ref{app:training}.

\paragraph{Position.} Each position entry acts independently on its own axis. A fixed entry ($m_i = 1$) imposes $p_i(\mathbf{q}) = p^{\star}_i$, contributing the scalar residual $p_i(\mathbf{q}) - p^{\star}_i$; a free entry leaves that axis unconstrained.

\paragraph{Orientation.} The three orientation entries are read jointly rather than per axis, and their effect depends only on how many are set.

If all three are set, the orientation is fully fixed, $\Matrix{R}(\mathbf{q}) = \Matrix{R}^\star$, removing all three rotational degrees of freedom. The error is measured by the angle between the current and target orientations,
\begin{equation}
d\left(\Matrix{R}(\mathbf{q}), \Matrix{R}^\star\right) = \arccos\left(\tfrac{1}{2}\left(\Trace{\Transpose{\Matrix{R}^\star}\,\Matrix{R}(\mathbf{q})} - 1\right)\right),
\label{eq:app-so3-geodesic}
\end{equation}
which is zero exactly when the two orientations coincide.

If two are set, one entry is free; let $\mathbf{e}_j$ be the body axis it indexes. The constraint fixes the spatial direction of that axis, $\Matrix{R}(\mathbf{q})\,\mathbf{e}_j = \Matrix{R}^\star \mathbf{e}_j$, removing two degrees of freedom while leaving rotation about $\mathbf{e}_j$ free. The error is simply the angle between the current and target directions of that axis.
There is no one-entry case, since a single rotational coordinate has no axis-independent meaning.

Stacking the constrained residuals gives the constraint-error map $h_\mathbf{c} : \Real{}^n \to \Real{}^k$, where $k = \Norm{\mathbf{m}}_1$ counts the constrained coordinates: one per constrained position axis, plus two or three for orientation depending on how many orientation entries are set.
The configurations that satisfy the constraint form the constraint manifold $\mathcal{M}_\mathbf{c} = \Inv{h_\mathbf{c}}(\Zero)$, i.e., those $\mathbf{q}$ whose end-effector pose agrees with the target $(\mathbf{p}^\star, \Matrix{R}^\star)$ on the constrained coordinates.
Wherever $\Zero$ is a regular value of $h_\mathbf{c}$, $\mathcal{M}_\mathbf{c}$ is a smooth manifold of dimension $n-k$ (Appendix~\ref{app:smooth-manifolds}).

We report the constraint residual through the constraint error $h_\mathbf{c}(\mathbf{q})$.
Because its position and orientation parts carry different units, we report them separately:
the position part as the Euclidean error over the constrained axes in millimetres, and the orientation part as the full geodesic angle~\eqref{eq:app-so3-geodesic} when the orientation is fully fixed, or the single axis angle when only an axis direction is constrained.

\begin{table}[t]
\centering
\small
\caption{\textbf{Precise definitions of the seven task-space constraints.}
For each constraint we list the selection vector $\mathbf{m} = (m_1\, m_2\, m_3 \mid m_4\, m_5\, m_6)$, the constrained task-space set $\mathcal{S}_\mathbf{c}$, and the codimension $k = \Norm{\mathbf{m}}_1$.
This table complements Table~\ref{tab:constraints}, which reports the resulting intrinsic dimension $n-k$ for each manipulator.}
\label{tab:constraint-defs}
\begin{tabular}{lllc}
\toprule
Constraint & Selection $\mathbf{m}$ & Constrained quantities ($\mathcal{S}_\mathbf{c}$) & $k$ \\
\midrule
\textsc{Plane}                            & $001 \mid 000$ & $p_z = p_z^\star$                                                          & 1 \\
\textsc{Line}                             & $011 \mid 000$ & $p_y = p_y^\star,\ p_z = p_z^\star$                                          & 2 \\
\textsc{Position}                         & $111 \mid 000$ & $\mathbf{p} = \mathbf{p}^\star$                                          & 3 \\
\textsc{Orientation}                      & $000 \mid 111$ & $\Matrix{R} = \Matrix{R}^\star$                                          & 3 \\
\textsc{Plane} $+$ \textsc{Orientation}   & $001 \mid 111$ & $p_z = p_z^\star,\ \Matrix{R} = \Matrix{R}^\star$                          & 4 \\
\textsc{Line} $+$ \textsc{Orientation}    & $011 \mid 110$ & $p_y = p_y^\star,\ p_z = p_z^\star,\ \Matrix{R}\mathbf{e}_z = \Matrix{R}^\star\mathbf{e}_z$ & 4 \\
$\LieGroupSE{3}$                          & $111 \mid 111$ & $\mathbf{p} = \mathbf{p}^\star,\ \Matrix{R} = \Matrix{R}^\star$          & 6 \\
\bottomrule
\end{tabular}
\end{table}

Table~\ref{tab:constraint-defs} lists the seven constraints together with their selection vectors and constrained sets.
Every constraint except \textsc{Line} $+$ \textsc{Orientation} leaves the orientation either fully fixed or fully free.
\textsc{Line} $+$ \textsc{Orientation} is the only case that constrains just an axis direction: it fixes the two position coordinates of the line and the direction of the end-effector $z$-axis, leaving translation along the line and rotation about that axis free.
The codimension $k$ is a property of the constraint alone; the intrinsic dimension $n-k$ also depends on the number of joints $n$, so the redundant 7-DoF Franka keeps one extra self-motion DoF relative to the 6-DoF UR5 under the same constraint (Table~\ref{tab:constraints}).

\subsection{Diffusion models and DDIM inversion}
\label{app:diffusion-background}

Diffusion models learn the conditional data distribution by reversing a Gaussian noising process.
For a clean sample $\mathbf{x}_0$ and variance schedule $\bar\alpha_t$, the forward process, defined by
\begin{equation}
\mathbf{x}_t \;=\; \sqrt{\bar\alpha_t}\,\mathbf{x}_0 \;+\; \sqrt{1-\bar\alpha_t}\,\boldsymbol{\epsilon}, \qquad \boldsymbol{\epsilon}\sim\NormalDistribution{\Zero}{\Identity},
\label{eq:forward}
\end{equation}
produces the latent variable $\mathbf{x}_t$ from which the denoising network $\hat{\boldsymbol{\epsilon}}_\theta(\mathbf{x}_t, t, \mathbf{c})$ predicts $\boldsymbol{\epsilon}$~\cite{ho2020denoising}.
DDIM~\citep{song2021denoising}, a non-Markovian reformulation of the reverse process, generates samples by iterating the deterministic update
\begin{equation}
    \mathbf{x}_{t-1} \;=\; \sqrt{\bar\alpha_{t-1}}\,\hat{\mathbf{x}}_0(\mathbf{x}_t,t,\mathbf{c}) \;+\; \sqrt{1-\bar\alpha_{t-1}}\,\hat{\boldsymbol{\epsilon}}_\theta(\mathbf{x}_t,t,\mathbf{c}),
    \qquad
    \hat{\mathbf{x}}_0 \;=\; \frac{\mathbf{x}_t - \sqrt{1-\bar\alpha_t}\,\hat{\boldsymbol{\epsilon}}_\theta}{\sqrt{\bar\alpha_t}}.
    \label{eq:ddim_step}
\end{equation}
Iterating~\eqref{eq:ddim_step} from $\mathbf{x}_T$ to $\mathbf{x}_0$ defines the deterministic map $g_{\mathbf{c}}: \mathcal{X} \to \Real^{n}, \mathbf{x}_T \mapsto \mathbf{x}_0$, which is locally invertible under the small-step approximation
$\hat{\boldsymbol{\epsilon}}_\theta(\mathbf{x}_{t-1},t-1,\mathbf{c}) \approx
\hat{\boldsymbol{\epsilon}}_\theta(\mathbf{x}_t,t,\mathbf{c})$~\citep{song2021denoising},
with generated samples concentrated near $\mathcal{M}_\mathbf{c}$ when the learned score closely approximates the true score.
Solving \eqref{eq:ddim_step} for $\mathbf{x}_t$ yields the forward-time \emph{DDIM inversion} update
\begin{equation}
\mathbf{x}_t \;\approx\; \sqrt{\bar\alpha_t}\,\hat{\mathbf{x}}_0(\mathbf{x}_{t-1}, t-1, \mathbf{c}) \;+\; \sqrt{1-\bar\alpha_t}\,\hat{\boldsymbol{\epsilon}}_\theta(\mathbf{x}_{t-1}, t-1, \mathbf{c}),
\label{eq:ddim_inv}
\end{equation}
which iterated from $\mathbf{x}_0$ recovers a latent $\mathbf{x}_T = g_\mathbf{c}^{-1}(\mathbf{x}_0)$ that maps back to $\mathbf{x}_0$.

%% file: sections/appendix/dataset-training.tex
\section{Dataset and Training}
\label{app:dataset-training}

This section details how the training data is generated and how the diffusion model is trained.
We first describe the analytical inverse-kinematics solver and how the discrete solution branches and continuous self-motion manifold are recovered (Appendix~\ref{app:ik-solver}), then the sampling of task-space targets for each constraint family (Appendix~\ref{app:target-sampling}), and finally the network architecture and optimization (Appendix~\ref{app:training}) that underlie the main results and every model in Appendix~\ref{app:id-estimation}.

\subsection{IK solver and solution recovery}
\label{app:ik-solver}
Our training data consists of joint configurations that satisfy a given task-space constraint, obtained by solving inverse kinematics for sampled task-space targets (the sampling procedure is described in Appendix~\ref{app:target-sampling}).
We use the analytical solver \texttt{IKFast}~\citep{diankov2010ikfast}, which symbolically solves the inverse-kinematics equations of a fixed kinematic chain and returns all closed-form solution branches for a queried end-effector pose.
An analytical solver is preferable to a numerical one here because, for each queried pose, it enumerates the discrete IK branches exhaustively and without an initial guess, so the recovered configurations are not tied to the basin of a particular seed.
For a full $\LieGroupSE{3}$ constraint, this returns the solution set of $\mathcal{M}_\mathbf{c}$ exactly; for the partial constraints, where the free task-space DoFs are sampled or swept on a grid (Appendix~\ref{app:target-sampling}) and near-duplicates are removed from the pooled solutions, it yields a finite sampled approximation of $\mathcal{M}_\mathbf{c}$.

Recovering the configurations for a single target differs between the two manipulators by their kinematic redundancy.
For the UR5, a full $\LieGroupSE{3}$ pose is reached by at most eight discrete configurations; one solver call returns this finite branch set.
For the Franka, the redundant joint promotes the solution set of a fixed pose to a one-dimensional self-motion manifold, which a single analytical query cannot return in closed form.
We therefore choose the seventh joint to parameterize the redundancy and sweep it over $50$ values spaced uniformly across its joint-limit interval $[-2.9671, 2.9671]$~rad, solving analytically with this joint held fixed at each value and collecting the in-limit solutions; this samples the self-motion manifold while still enumerating the discrete branches at every sweep value.
This sweep provides a discretization of the self-motion manifold under the chosen free-joint parameterization; it is not assumed to be a globally uniform parameterization of the manifold, and may under-sample narrow components or regions near singular parameterizations.

The procedure above can return many near-identical configurations, both from the redundant-joint sweep and from the over-sampling of task-space targets (Appendix~\ref{app:target-sampling}).
We therefore remove near-duplicate configurations in joint space: a candidate is kept only if its $\ell_2$ distance to every already-kept configuration exceeds a tolerance of $0.05$~rad ($0.1$~rad for the $\LieGroupSE{3}$ family).
This distance is computed on the raw joint angles and therefore does not account for angle wrap-around, so a pair of physically close configurations straddling the $0/2\pi$ boundary of a UR5 joint would be treated as roughly $2\pi$ apart and both kept.
In practice, the analytical solver returns each physical solution once within a fixed joint interval, which spans less than $2\pi$ for the Franka and follows the native $[0, 2\pi)$ convention for the UR5, so such boundary-split near-duplicates are rare; in any case, before training every configuration is mapped to the sine--cosine representation of Appendix~\ref{app:training}, which is free of the $2\pi$ wrap-around, so any residual boundary pair introduces no discontinuity for the model.

\begin{table}[!t]
\centering
\caption{\textbf{Reachable-workspace bounds used for task-space target
sampling.} Targets are drawn within a cylindrical shell: radius
$r=\sqrt{x^2+y^2}\in[r_{\min},r_{\max}]$ and height $z\in[z_{\min},z_{\max}]$,
in metres. Sampled positions outside these bounds are rejected.}
\label{tab:workspace-bounds}
\small
\begin{tabular}{lcccc}
\toprule
Manipulator & $r_{\min}$ & $r_{\max}$ & $z_{\min}$ & $z_{\max}$ \\
\midrule
UR5    & 0.20 & 0.70 & $-0.10$ & 0.80 \\
Franka & 0.10 & 0.85 & $-0.10$ & 1.10 \\
\bottomrule
\end{tabular}
\end{table}

\begin{table}[!t]
\centering
\caption{
\looseness=-1
\textbf{Per-family data-generation procedures.}
Each row defines a target by fixing the variables in column~2, drawn across targets on a uniform grid, by rejection sampling in the reachable shell of Table~\ref{tab:workspace-bounds}, or from the forward kinematics of a random configuration, as the orientation $\Matrix{R}(\mathbf{q}_{\text{rand}})$ or the full pose $f(\mathbf{q}_{\text{rand}})$.
The remaining task-space DoFs (column~3) are sampled or swept only to generate full-pose IK queries whose solutions lie on $\mathcal{M}_\mathbf{c}$.
Column~4 is the nominal number of targets and column~5 the per-target cap on the number of configurations kept under the greedy duplicate removal; there is no per-target cap for the $\LieGroupSE{3}$ constraint and all enumerated branches are included.
The final column reports the number of configurations kept per family after joint-limit filtering and near-duplicate removal.}
\label{tab:target-sampling}
\small
\setlength{\tabcolsep}{4pt}
\begin{tabularx}{\linewidth}{@{}l l Y r r c@{}}
\toprule
Constraint & Fixed & Free DoFs (sampled\,/\,swept) & \# targ. & Cap/targ. & Total$^\ast$ ($\times10^3$) \\
\midrule
\textsc{Plane}        & $p_z^\star$            & $(p_x,p_y)$ in shell, $\Matrix{R}$ random & 400  & 600       & $240\,/\,240$ \\
\textsc{Line}         & $(p_y^\star,p_z^\star)$  & $p_x$ along line, $\Matrix{R}$ random   & 600  & 400       & $239\,/\,211$ \\
\textsc{Position}     & $\mathbf{p}^\star$         & $\Matrix{R}$ random orientations        & 3000 & $\le 200$ & $546\,/\,577$ \\
\textsc{Orientation}  & $\Matrix{R}^\star$           & $\mathbf{p}$ grid ball about seed       & 800  & 600       & $274\,/\,106$ \\
\textsc{Plane}$+$\textsc{Orient.} & $(p_z^\star, \Matrix{R}^\star)$ & $(p_x,p_y)$ in shell        & 1000 & 300       & $300\,/\,299$ \\
\textsc{Line}$+$\textsc{Orient.}  & $(p_y^\star,p_z^\star,\Matrix{R}^\star\mathbf{e}_z)$ & $p_x$, rotation about $\mathbf{e}_z$ ($30\times30$ grid) & 800 & 500 & $240\,/\,95$ \\
$\LieGroupSE{3}$      & $(\mathbf{p}^\star,\Matrix{R}^\star)$ & none (discrete branches)     & 8000 & all branches & $62\,/\,255$ \\
\bottomrule
\end{tabularx}

{\footnotesize $^\ast$Configurations kept after joint-limit filtering and near-duplicate removal, reported as UR5\,/\,Franka; each full-pose IK query may return several IK branches, so these totals fall below the product of columns~4 and~5.}
\end{table}

\subsection{Task-space target sampling}
\label{app:target-sampling}
Each target fixes its constrained task-space DoFs at the values of a sampled pose $(\mathbf{p}^\star, \Matrix{R}^\star)$, according to the selection vector $\mathbf{m}$ (Appendix~\ref{app:constraints}).
Generating training data for a family therefore consists of (i) sampling a target by choosing values for its constrained DoFs, and (ii) populating the corresponding manifold $\mathcal{M}_\mathbf{c}$ by sweeping the free DoFs and solving IK at each, with near-duplicates removed from the recovered configurations as in Appendix~\ref{app:ik-solver}.

Two conventions are shared across all families.
First, every sampled end-effector position is constrained to the reachable shell of Table~\ref{tab:workspace-bounds}, and positions falling outside it are rejected; this keeps targets within the manipulator's workspace and away from near-base and near-singular regions.
Second, sampling a target orientation $\Matrix{R}^\star$ directly in $\LieGroupSO{3}$ would frequently yield poses unreachable for the non-redundant UR5; we instead obtain orientations as the rotation component $\Matrix{R}(\mathbf{q}_{\text{rand}})$ of the forward kinematics of random joint configurations, which biases sampled orientations toward the robot's reachable orientation set.
This restricts each orientation to one the arm reaches in some configuration, but does not guarantee that the orientation is jointly reachable with a separately sampled position or partial-position constraint, so a small fraction of full poses remain infeasible and are removed by the joint-limit filtering of Appendix~\ref{app:ik-solver}.

The per-family data-generation procedures are summarized in Table~\ref{tab:target-sampling}.
A target fixes the constrained DoFs that define $\mathbf{c}$, drawn across targets on a uniform grid (e.g.\ the plane height $p_z^\star$), by rejection sampling in the shell, or as $\Matrix{R}(\mathbf{q}_{\text{rand}})$ for orientations, while the remaining free DoFs are swept on a grid (\textsc{Orientation}, \textsc{Line}$+$\textsc{Orientation}) or sampled at random (\textsc{Position}, \textsc{Plane}, \textsc{Line}, \textsc{Plane}$+$\textsc{Orientation}) only to generate the full-pose IK queries whose solutions populate $\mathcal{M}_\mathbf{c}$.
The per-family counts in the final column fall below the product of the target count and per-target cap, because the joint-space near-duplicate removal of Appendix~\ref{app:ik-solver} and the joint-limit filtering together discard near-identical and out-of-range solutions; the reduction is largest for \textsc{Orientation} and \textsc{Line}$+$\textsc{Orientation}, and is more pronounced on the Franka, whose tighter joint limits reject more candidate solutions.

\subsection{Network and training details}
\label{app:training}
The network is conditioned on the constraint $\mathbf{c} = (\mathbf{p}^\star, \Matrix{R}^\star, \mathbf{m})$ of Appendix~\ref{app:constraints}, encoded as a $15$-dimensional conditioning vector that captures both the target and the constraint family.
Its first three entries hold the target position $\mathbf{p}^\star$, normalized to the reachable shell of Table~\ref{tab:workspace-bounds}; the next six encode the target orientation $\Matrix{R}^\star$ in a continuous $6$-D rotation representation~\citep{zhou2019continuity}, which avoids the discontinuities of Euler-angle or quaternion parameterizations.
The final six entries are the selection vector $\mathbf{m}$ (Appendix~\ref{app:constraints}), which identifies the constraint family of Table~\ref{tab:constraints}.
For families that leave position partially free, the unconstrained entries of $\mathbf{p}^\star$ are set to zero, so the network is conditioned only on the coordinates the constraint actually fixes.

The model is an \emph{endpoint} diffusion model: it learns a distribution over the joint configuration $\mathbf{q}$ that solves the constraint, rather than
over a task-space trajectory.
To remove the $2\pi$ wrap-around discontinuity of raw joint angles, each configuration is mapped to a sine--cosine encoding $\phi(\mathbf{q}) =
(\sin\mathbf{q},\, \cos\mathbf{q}) \in \Real{}^{2n}$, giving a $12$-dimensional input for the $6$-DoF UR5 and $14$-dimensional for the $7$-DoF Franka.
The encoded endpoints are standardized to zero mean and unit variance per coordinate before diffusion.
The network is trained with the noise-prediction ($\epsilon$) objective of \citet{ho2020denoising}, minimizing the mean-squared error between the sampled
noise and the network output.
At inference we draw a sample and invert the per-coordinate standardization, giving a vector $(\hat{\mathbf{a}}, \hat{\mathbf{b}}) \in \Real{}^{2n}$ whose two halves are the model's estimates of $(\sin\mathbf{q}, \cos\mathbf{q})$.
These coordinates are unconstrained and need not lie on the unit circle, so we recover each joint angle as the polar angle of the corresponding generated $2$-vector, $q_j = \operatorname{atan2}(\hat{a}_j, \hat{b}_j) \in (-\pi, \pi]$, which depends only on its direction and discards the radial magnitude, equivalently projecting $(\hat{a}_j, \hat{b}_j)$ onto the unit circle before decoding.
The corresponding task-space pose follows from the forward kinematics $f(\mathbf{q})$.

\paragraph{Architecture.}
The denoiser is a conditional residual multilayer perceptron rather than a convolutional U-Net, since the endpoint carries no temporal structure to exploit.
The noisy input is projected to a hidden width of $512$ and passed through $10$ residual blocks, each applying a pre-LayerNorm, a feature-wise linear
modulation (FiLM) of the normalized activations, and a two-layer MLP with a $4\times$ inner expansion and Mish activations; a final LayerNorm and linear
layer map back to the input dimension.
The diffusion timestep $t$ is embedded with a sinusoidal positional encoding followed by a two-layer MLP, the conditioning vector $\mathbf{c}$ is embedded
by a separate three-layer MLP, and the two embeddings (each of width $384$) are summed to form the modulation signal that drives the FiLM scale and shift in
every block.
This conditioning scheme lets a single network serve all seven constraint families. The selection vector $\mathbf{m}$ tells the model which task-space
coordinates are active, so one set of weights covers the whole family of Table~\ref{tab:constraints}.
The model has $25.6$M parameters.

\paragraph{Diffusion and optimization.}
We use a discrete forward process of $T = 100$ steps whose per-step noise variances follow a sigmoid $\beta$ schedule, interpolating
from $\beta_{\min} = 10^{-4}$ to $\beta_{\max} = 2 \times 10^{-2}$; this schedule gave the lowest constraint error in our ablation.
Samples are drawn with a deterministic DDIM sampler \citep{song2021denoising} using the full $100$-step grid.
The network is trained for $500{,}000$ steps with AdamW (learning rate $2\times10^{-4}$, weight decay $10^{-5}$) at batch size $512$, with gradients clipped
to unit norm.
Each batch is drawn by a stratified sampler that balances the seven constraint families equally, so no family dominates the gradient.
We apply classifier-free conditioning dropout, setting the entire conditioning vector to zero with probability $0.10$, and maintain an exponential moving average of
the weights (decay $0.9995$, after a $1{,}000$-step warm-up) that is used for all evaluation.

\begin{table}[t]
\centering
\caption{\textbf{Network and training hyperparameters} (shared across both manipulators; only the input dimension differs).}
\label{tab:training-hparams}
\small
\begin{tabular}{ll}
\toprule
Hyperparameter & Value \\
\midrule
Hidden width                 & $512$ \\
Residual blocks (depth)      & $10$ \\
MLP inner expansion          & $4\times$ \\
Embedding width (time, cond) & $384$ \\
Parameters                   & $25.6$M \\
\midrule
Diffusion steps $T$          & $100$ \\
Noise schedule               & sigmoid ($\beta \in [10^{-4}, 2\times10^{-2}]$) \\
Sampler                      & DDIM, $100$ steps, $\eta = 0$ \\
\midrule
Optimizer                    & AdamW \\
Learning rate                & $2\times10^{-4}$ \\
Weight decay                 & $10^{-5}$ \\
Batch size                   & $512$ \\
Training steps               & $500{,}000$ \\
Gradient clipping            & $1.0$ \\
Conditioning dropout         & $0.10$ \\
EMA decay (warm-up)          & $0.9995$ ($1{,}000$ steps) \\
\bottomrule
\end{tabular}
\end{table}

\subsection{Training-set size ablation}
\label{app:samples}

In image classification, datasets of lower intrinsic dimension have been observed empirically to need fewer training samples to reach a given accuracy~\citep{pope2021intrinsic}.
Although higher-dimensional manifolds are often harder to estimate statistically, our data-generation setting has an additional effect: low-codimension constraints leave larger feasible sets per target and are therefore easier to populate with valid IK samples. 
A constraint family of codimension $k$ leaves an $(n-k)$-dimensional set of solutions for every task-space target, so a higher intrinsic dimension $n-k$ offers a larger family of valid configurations per target and could make the manifold $\mathcal{M}_\mathbf{c}$ easier to populate from limited data.
We examine whether this holds in our setting by training a separate model on each constraint family, using the architecture and optimization of Appendix~\ref{app:training} with $100{,}000$ training steps, while varying the number of training configurations $N$ from $500$ to $40{,}000$.
Each model is evaluated on $50$ targets with $512$ samples each on the UR5, and on $10$ targets with $32$ samples on the Franka ($20$ with $256$ for the two grid-generated families, and for $\LieGroupSE{3}$ at $N{=}500$, where no sample of the smaller budget is valid); we report the position and orientation error averaged over its valid samples, using the validity tolerances of Table~\ref{tab:tbar}, which admit per-axis errors of up to $100$~mm, as a function of $N$.

\begin{figure}[t]
\centering
\includegraphics[width=\linewidth]{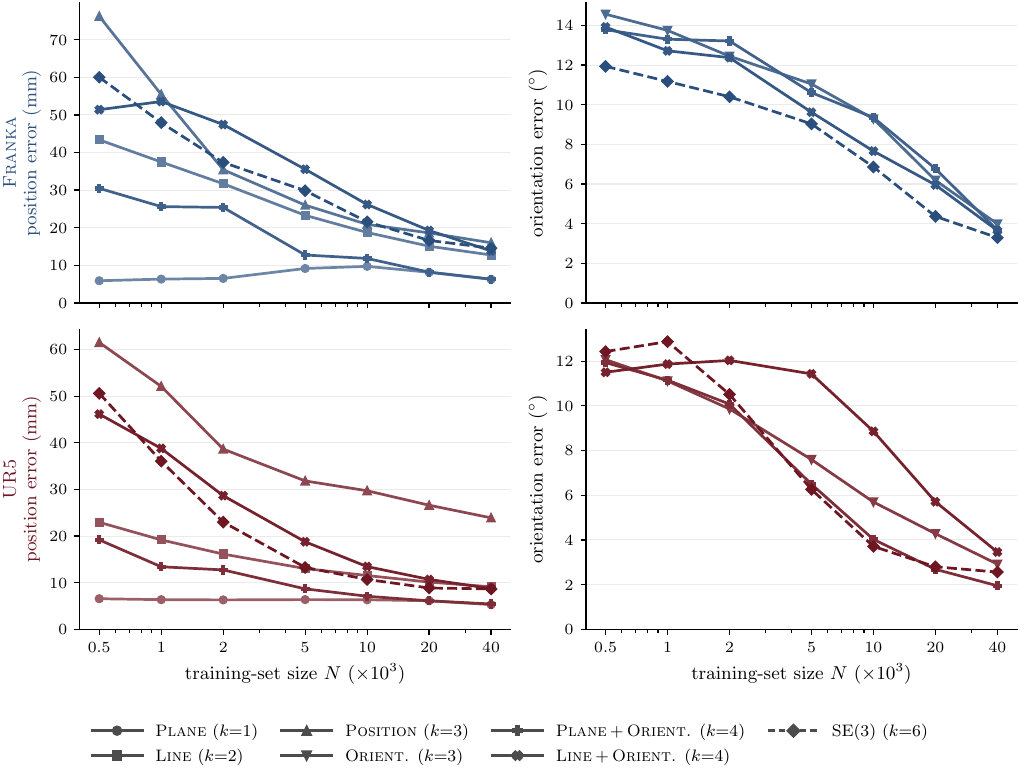}
\vspace{-4mm}
\caption{\textbf{Constraint error versus training-set size.}
Position error (left) and orientation error (right) of per-family models for \textcolor{frankaBlue}{Franka} (top) and \textcolor{ur5Red}{UR5} (bottom), against the number of training configurations $N$, with one curve per constraint family identified by its marker and drawn as a distinct shade of the manipulator's colour, from light to dark with increasing codimension $k$; the discrete $\LieGroupSE{3}$ condition is dashed.
The lowest-codimension family, \textsc{Plane}, reaches low error with little data, while higher-codimension families improve across the whole range. Errors are means over the samples inside the validity tolerances of Table~\ref{tab:tbar}.}
\label{fig:size-sweep}
\end{figure}

Figure~\ref{fig:size-sweep} shows the resulting error curves, which are broadly consistent with the expectation above on both manipulators, although not as a strict rule.
The low-codimension family \textsc{Plane} ($k{=}1$) reaches low error almost immediately and gains little from additional data on both manipulators, whereas \textsc{Line} ($k{=}2$) improves with data, from $23$ to $9$~mm on the UR5. The high-codimension families \textsc{Plane}$+$\textsc{Orientation} and \textsc{Line}$+$\textsc{Orientation} ($k{=}4$) start with large error at $N{=}500$ and improve across the whole range, the latter falling from $51$ to $14$~mm and from $13.9^\circ$ to $3.7^\circ$ on the Franka.
Ordering the families by codimension thus roughly tracks the amount of data each appears to need to reach a given error, in line with the view that larger solution families are easier to learn.

The two $\LieGroupSE{3}$ conditions are the hardest at small $N$.
On the Franka the redundant joint promotes the fixed pose to a one-dimensional self-motion manifold, but with $k{=}6$ the model generates valid configurations for only $0.1\%$ of its samples at $N{=}500$ ($5$ of $5{,}120$) and $8\%$ at $N{=}1{,}000$, where its position error is $48$~mm; it improves across the whole range, to $15$~mm and $3.3^\circ$ at $N{=}40{,}000$.
On the UR5 the same condition is as hard, since a full pose admits only the eight discrete configurations of a non-redundant arm, so its error is among the largest at small $N$ and approaches the continuous families by $N{=}10{,}000$.
The UR5 \textsc{Position} family is a further exception, its error remaining well above the others at large $N$, suggesting that training efficiency is not explained by codimension alone, since \textsc{Position} shares its codimension with \textsc{Orientation}, which trains without difficulty.

We finally compare the per-family models against the single jointly trained model of Appendix~\ref{app:training}, which is conditioned on all seven families at once and underlies our main results.
Table~\ref{tab:size-sweep-compare} reports this comparison at the largest training-set size $N{=}40{,}000$.
The jointly trained model achieves lower or equal position and orientation error on every condition except the position error of the UR5 under $\LieGroupSE{3}$ ($8.7$ against $9.1$~mm), most notably on the UR5 \textsc{Position} family, where the position error falls from $23.9$ to $11.7$~mm.
Training across families thus appears to help most on the UR5 \textsc{Position} family, the condition the per-family models found hardest.
The two models also differ in training length and dataset size, so this is a descriptive comparison of the trained models rather than a controlled study of joint versus per-family training.

\begin{table}[t]
\centering
\small
\setlength{\tabcolsep}{6pt}
\renewcommand{\arraystretch}{1.15}
\caption{\textbf{Per-family models against the jointly trained model.}
Position error (mm) and orientation error ($^\circ$) for the largest per-family models ($N{=}40{,}000$) and for the single jointly trained model of Appendix~\ref{app:training}, which conditions on all families.
The ID column is the analytical intrinsic dimension $n-k$ of the joint-space solution manifold; a dash marks a metric the constraint does not fix.}
\label{tab:size-sweep-compare}
\begin{tabular}{@{}llccccc@{}}
\toprule
& & & \multicolumn{2}{c}{Per-family ($N{=}40$k)} & \multicolumn{2}{c}{Jointly-trained} \\
\cmidrule(lr){4-5}\cmidrule(l){6-7}
& Constraint & ID & Pos.\ [mm] & Orn.\ [$^\circ$] & Pos.\ [mm] & Orn.\ [$^\circ$] \\
\midrule
\multirow{7}{*}{\rotatebox[origin=c]{90}{\textsc{Franka}}}
 & \textsc{Plane}                          & 6 & $6.3$  & --    & $4.1$  & --    \\
 & \textsc{Line}                           & 5 & $12.7$ & --    & $8.2$  & --    \\
 & \textsc{Position}                       & 4 & $16.0$ & --    & $12.5$ & --    \\
 & \textsc{Orientation}                    & 4 & --     & $4.0$ & --     & $2.3$ \\
 & \textsc{Plane} $+$ \textsc{Orientation} & 3 & $6.3$  & $3.7$ & $4.7$  & $2.2$ \\
 & \textsc{Line} $+$ \textsc{Orientation}  & 3 & $13.8$ & $3.7$ & $10.7$ & $2.0$ \\
 & $\LieGroupSE{3}$                        & 1 & $14.6$ & $3.3$ & $12.7$ & $2.5$ \\
\midrule
\multirow{7}{*}{\rotatebox[origin=c]{90}{\textsc{UR5}}}
 & \textsc{Plane}                          & 5 & $5.3$  & --    & $4.7$  & --    \\
 & \textsc{Line}                           & 4 & $9.1$  & --    & $7.2$  & --    \\
 & \textsc{Position}                       & 3 & $23.9$ & --    & $11.7$ & --    \\
 & \textsc{Orientation}                    & 3 & --     & $2.9$ & --     & $2.0$ \\
 & \textsc{Plane} $+$ \textsc{Orientation} & 2 & $5.4$  & $1.9$ & $4.9$  & $1.7$ \\
 & \textsc{Line} $+$ \textsc{Orientation}  & 2 & $8.7$  & $3.5$ & $7.8$  & $3.5$ \\
 & $\LieGroupSE{3}$                        & 0 & $8.7$  & $2.6$ & $9.1$ & $2.3$ \\
\bottomrule
\end{tabular}
\end{table}

%% file: sections/appendix/id-estimation.tex
\section{Intrinsic Dimension Estimation}
\label{app:id-estimation}

This section expands the intrinsic-dimension study of Section~\ref{sec:id}.
We first state the full Score-SVD$^{\dagger}$ algorithm with its sample-selection procedure and a worked example of the discrete UR5 $\LieGroupSE{3}$ case (Appendix~\ref{app:discrete}), then study the sensitivity to its two hyperparameters (Appendix~\ref{app:id-hyperparameters}), and describe the neighbourhood-based baselines we compare against (Appendix~\ref{app:id-baselines}).
We then report the full numerical results and score spectra across all conditions (Appendix~\ref{app:id-full-results}), and examine how the constraint error depends on the size of the training set (Appendix~\ref{app:samples}).

\subsection{Score-SVD\texorpdfstring{$^{\dagger}$}{} algorithm, sample selection, and example}
\label{app:discrete}

\looseness=-1
Section~\ref{sec:id} introduced Score-SVD$^{\dagger}$, our variant of the score-based estimator of \citet{stanczuk2024diffusion} that resolves the discrete, $\textrm{ID}=0$ case.
Algorithm~\ref{alg:score-svd-dagger} states the full procedure at a single sample $\mathbf{x}_0$, written as a modification of Algorithm~1 of \citet{stanczuk2024diffusion}, with the changes highlighted in \dgm{teal}.
The score is read from the \emph{conditional}, \emph{variance-preserving}, $\epsilon$-parameterized network of Appendix~\ref{app:training} rather than from an unconditional variance-exploding score model, so the $K$ perturbations are drawn with the forward kernel $\mathbf{x}_{t} = \sqrt{\bar\alpha_{t}}\,\mathbf{x}_0 + \sqrt{1-\bar\alpha_{t}}\,\boldsymbol{\epsilon}$ of \citet{ho2020denoising} and the score recovered as $s_\theta = -\hat{\boldsymbol{\epsilon}}_\theta / \sqrt{1-\bar\alpha_{t}}$, evaluated at an intermediate $t$ rather than in the $t \to 0$ limit (the sensitivity to $t$ is studied in Appendix~\ref{app:id-hyperparameters}).
We also modify the estimator itself, appending a virtual zero singular value $\sigma_{n+1} = 0$ to the descending spectrum and extending the gap search of~\eqref{eq:score_svd_rule} from $i = 1,\dots,n-1$ to $i = 1,\dots,n$, so that the drop from the smallest singular value $\sigma_n$ to zero becomes an admissible gap.

\begin{algorithm}[t]
\caption{Score-SVD\texorpdfstring{$^{\dagger}$}{} intrinsic-dimension estimate at a sample $\mathbf{x}_0$. Changes to Algorithm~1 of \citet{stanczuk2024diffusion} are shown in \dgm{teal}.}
\label{alg:score-svd-dagger}
\begin{algorithmic}[1]
\Require \dgm{conditional} score model $s_\theta(\cdot,\cdot\,\dgm{,\,\mathbf{c}})$; sample $\mathbf{x}_0 \in \mathcal{M}_\mathbf{c}$; \dgm{intermediate} noise level $t$; perturbation count $K$
\State $n \gets \dim(\mathbf{x}_0)$;\quad $S \gets [\;]$
\For{$i = 1, \dots, K$}
  \State $\boldsymbol{\epsilon}^{(i)} \sim \NormalDistribution{\Zero}{\Identity}$
  \State \dgm{$\mathbf{x}_{t}^{(i)} \gets \sqrt{\bar\alpha_{t}}\,\mathbf{x}_0 + \sqrt{1-\bar\alpha_{t}}\,\boldsymbol{\epsilon}^{(i)}$} \Comment{\dgm{variance-preserving kernel}}
  \State \dgm{$\mathbf{s}^{(i)} \gets -\,\hat{\boldsymbol{\epsilon}}_\theta(\mathbf{x}_{t}^{(i)},\, t,\, \mathbf{c}) \,/\, \sqrt{1-\bar\alpha_{t}}$} \Comment{\dgm{score from $\epsilon$-prediction}}
  \State append $\mathbf{s}^{(i)}$ as a column of $S$
\EndFor
\State $\sigma_1 \ge \cdots \ge \sigma_n \gets \mathrm{SVD}(S)$ \Comment{singular values of $S \in \Real{}^{n \times K}$}
\State \dgm{$\sigma_{n+1} \gets 0$} \Comment{\dgm{virtual zero singular value}}
\State $\hat{k} \gets \ArgMax{i=1,\dots,\dgm{n}}\,(\sigma_i - \sigma_{i+1})$ \Comment{\dgm{search extended from $n-1$ to $n$}}
\State \Return $\hat{d} \gets n - \hat{k}$
\end{algorithmic}
\end{algorithm}

\paragraph{Sample selection.}
\citet{stanczuk2024diffusion} draw the sample $\mathbf{x}_0 \sim p_0$ from the dataset; we instead draw it from the constraint manifold itself.
For each constraint family we use held-out targets, drawn by a generator separate from the one that produced the training data and with its own random stream. Every target is a constraint $\mathbf{c} = (\mathbf{p}^\star, \Matrix{R}^\star, \mathbf{m})$ together with the set of analytical IK solutions that satisfy it, with near-duplicates removed, i.e.\ points lying exactly on $\mathcal{M}_\mathbf{c}$.
For each target, we draw $512$ samples from the model and select the $K_s = 5$ analytical solutions nearest, in the sine--cosine representation of Appendix~\ref{app:training}, to the valid ones, falling back to a uniform draw from the solution set when no sample is valid; each selected solution is encoded in that representation and standardized with the model's normalizer, with the conditioning vector $\mathbf{c}$ held fixed at the target value.
Evaluating at analytical solutions keeps every $\mathbf{x}_0$ exactly on the true manifold, and selecting those nearest to the model's samples places them in the part of $\mathcal{M}_\mathbf{c}$ that the model generates.
We then run Algorithm~\ref{alg:score-svd-dagger} at each sample with $K = 50$ perturbations at $t = 15$, and report the median of $\hat{d}$ over the $K_s$ samples and over up to $15$ targets per family.
Here, the score is evaluated in the sine--cosine input space, so the ambient dimension $n$ is twice the joint count, namely $12$ for the UR5 and $14$ for the Franka; because this encoding is a smooth embedding of joint space, $\hat{d}$ recovers the dimension of $\mathcal{M}_\mathbf{c}$ unchanged.

\paragraph{Worked example: the discrete UR5 $\LieGroupSE{3}$ case.}
Under a full $\LieGroupSE{3}$ constraint, the UR5 reaches a target with at most eight isolated IK branches, so $\mathcal{M}_\mathbf{c}$ is $0$-dimensional. Its tangent space is zero-dimensional and every ambient direction is normal.
The score matrix therefore has no near-zero tangent block, and all $n = 12$ singular values stay comparably large.
A representative sample ($t = 15$, $K = 50$) gives the descending spectrum
\begin{align*}
\LieGroupSE{3}:\quad & (153.9,\ \mathbf{144.2},\ \mathbf{129.3},\ 125.3,\ 114.9,\ 105.7,\ 101.3,\ 88.5,\ 82.5,\ 74.6,\ 66.2,\ 55.6), \\
\textsc{Line}:\quad  & (146.8,\ 138.5,\ 127.0,\ 112.4,\ 107.3,\ 95.2,\ 83.8,\ \mathbf{70.2},\ \ \mathbf{9.6},\ \ 4.3,\ \ 3.4,\ \ 3.1),
\end{align*}
shown alongside a \textsc{Line} sample (true $\textrm{ID} = 4$) for contrast.
For $\LieGroupSE{3}$ the largest interior drop, $\sigma_2 - \sigma_3 = 14.9$, does not mark a tangent--normal boundary, so the standard rule~\eqref{eq:score_svd_rule} returns $\hat{k} = 2$ and a meaningless $\hat{d} = 10$.
Score-SVD$^{\dagger}$ appends $\sigma_{13} = 0$; the drop $\sigma_{12} - \sigma_{13} = 55.6$ now dominates every interior gap, giving $\hat{k} = 12$ and the correct $\hat{d} = 0$.
The \textsc{Line} sample shows that the modification has no effect whenever the manifold is positive-dimensional. Its four tangent directions collapse to a near-zero tail, the interior gap $\sigma_8 - \sigma_9 = 60.6$ is far larger than the appended $\sigma_{12} - \sigma_{13} = 3.1$, and both rules return $\hat{d} = 4$.

\subsection{Hyperparameter sensitivity}
\label{app:id-hyperparameters}

Algorithm~\ref{alg:score-svd-dagger} has two free hyperparameters, the noise level $t$ at which the score is evaluated and the number of perturbations $K$.
Every estimate in Section~\ref{sec:id} uses $t = 15$ and $K = 50$; here we sweep each in turn, holding the other at its canonical value, and show that both lie inside a wide range over which the recovered dimension is exactly the analytical one.
Throughout we report the median of $\hat{d}$ over targets together with its interquartile band, and overlay standard Score-SVD with Score-SVD$^{\dagger}$; unlike Appendix~\ref{app:discrete}, the sweeps draw the $K_s = 5$ samples uniformly without replacement from each target's solution set.
The two estimators coincide on every positive-dimensional condition and separate only on the discrete UR5 $\LieGroupSE{3}$ case, where standard Score-SVD has no spectral gap to detect and only Score-SVD$^{\dagger}$ recovers the analytical $\hat{d} = 0$.

\begin{figure}[t]
\centering
\includegraphics[width=\linewidth]{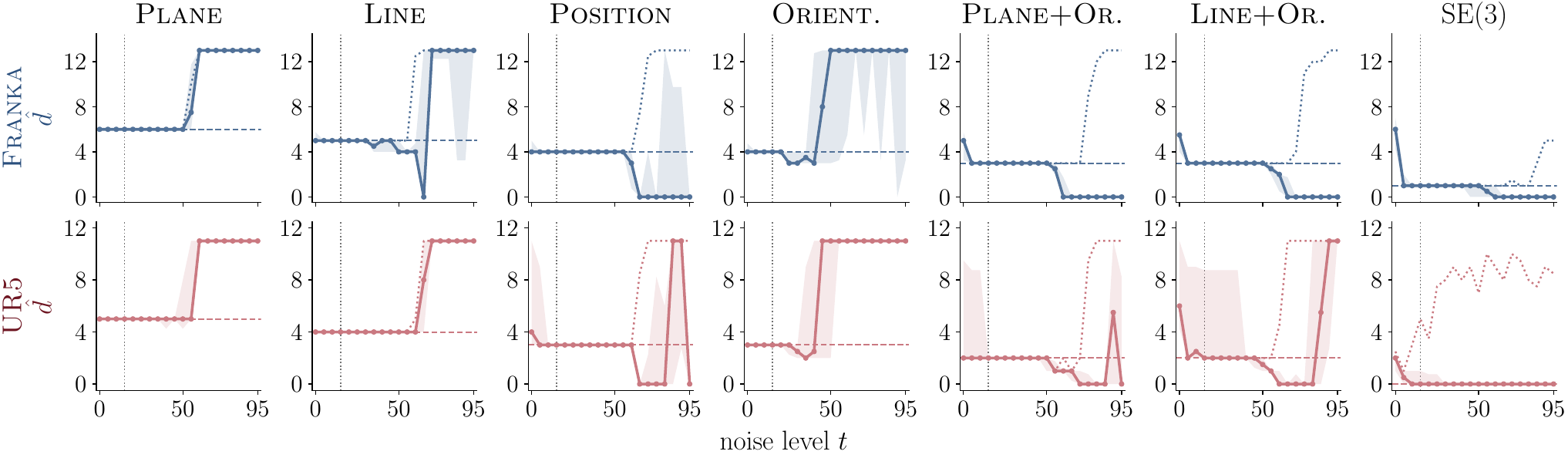}
\vspace{-4mm}
\caption{
\textbf{Score-based intrinsic-dimension estimate versus noise level $t$.}
Estimated intrinsic dimension $\smash{\hat{d}}$ of Algorithm~\ref{alg:score-svd-dagger} as a function of the noise level $t$, with the perturbation count fixed at $K = 50$; \textcolor{frankaBlue}{Franka} (top) and \textcolor{ur5Red}{UR5} (bottom), one constraint per column.
\@ \linedotted{scoreBlue} / \linedotted{scoreRed}~Score-SVD;
\@ \linethick{dagBlue} / \linethick{dagRed}~Score-SVD$^{\dagger}$ (median over targets, shaded interquartile band);
\@ \linedashed{scoreBlue} / \linedashed{scoreRed}~analytical intrinsic dimension of Table~\ref{tab:constraints}.
The dotted vertical line marks the canonical $t = 15$. The estimate stays at the analytical value across a wide low-to-moderate range that contains $t = 15$ on all fourteen conditions, and degrades only once the noised samples become nearly isotropic.
}
\label{fig:id-noise-sensitivity}
\end{figure}

\paragraph{Noise level.}
Figure~\ref{fig:id-noise-sensitivity} sweeps $t$ with $K$ fixed at $50$.
For both robots and all seven constraints the estimate stays at the analytical intrinsic dimension across a wide range of low-to-moderate noise, and the canonical $t = 15$ falls inside it on every one of the fourteen conditions.
This range is bounded on both sides.
As $t \to 0$, the perturbations are too small to spread the samples across the normal bundle, so the score matrix is near-degenerate and the gap rule becomes unstable. At the smallest noise level, it overestimates the highest-codimension conditions (both $\LieGroupSE{3}$ conditions, both \textsc{Line} $+$ \textsc{Orientation} conditions, and additionally Franka \textsc{Plane} $+$ \textsc{Orientation} and UR5 \textsc{Position}). Both $\LieGroupSE{3}$ conditions recover their analytical value by $t=10$, and the rest by $t = 5$.
As $t$ grows, the forward kernel of \citet{ho2020denoising} shrinks the clean-data component, and the noised samples become progressively isotropic.
The tangent--normal split then disappears, and the estimate either saturates near the ambient dimension $n$, when the surviving spectrum is dominated by a single direction, or collapses towards $0$, when the noised spectrum is flat and the appended zero singular value of Score-SVD$^{\dagger}$ gives the largest gap.
The estimator therefore deviates from the analytical dimension above a certain noise level, the same behaviour we exploit in Section~\ref{sec:interpolation}. It is reliable precisely in the low-noise region in which the intermediate representation has not yet expanded beyond the constraint manifold.

\begin{figure}[t]
\centering
\includegraphics[width=\linewidth]{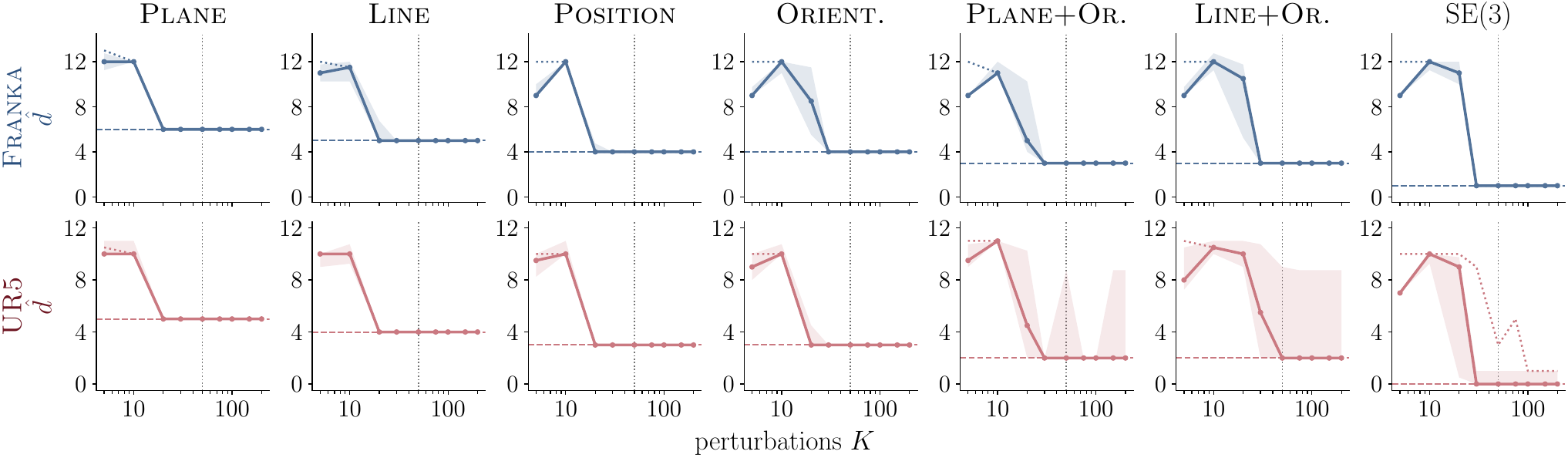}
\vspace{-4mm}
\caption{
\textbf{Score-based intrinsic-dimension estimate versus perturbation count $K$.}
Estimated intrinsic dimension $\smash{\hat{d}}$ of Algorithm~\ref{alg:score-svd-dagger} as a function of the perturbation count $K$ (log axis), with the noise level fixed at $t = 15$; \textcolor{frankaBlue}{Franka} (top) and \textcolor{ur5Red}{UR5} (bottom), one constraint per column. Score-SVD (dotted), Score-SVD$^{\dagger}$ (solid, shaded interquartile band), the dashed analytical intrinsic dimension, and the canonical $K = 50$ (dotted vertical line) follow Figure~\ref{fig:id-noise-sensitivity}. Below the ambient dimension $n$ the estimate is rank-limited and overshoots; it saturates at the analytical value around $K \geq 2n$ on most conditions and by $K = 50$ on all of them.
}
\label{fig:id-k-sensitivity}
\end{figure}

\paragraph{Perturbation count.}
Figure~\ref{fig:id-k-sensitivity} sweeps $K$ with $t$ fixed at $15$.
The number of perturbations sets the number of columns of the score matrix $S \in \Real{}^{n \times K}$, so for $K$ below the ambient dimension $n$, the matrix has rank at most $K$ and cannot expose the full tangent--normal split; the estimate is rank-limited and overshoots towards $n$.
As $K$ increases past $n$, the estimate drops onto the analytical intrinsic dimension and saturates, becoming flat for all larger $K$ on both robots.
The canonical $K = 50$ lies well inside this saturated range for the $n = 12$ (UR5) and $n = 14$ (Franka) score spaces, and is consistent with the $K = 4n$ rule of thumb of \citet{stanczuk2024diffusion}, whose values ($48$ and $56$) bracket our choice.

\subsection{Baseline methods}
\label{app:id-baselines}

We compare Score-SVD against the three classical neighbourhood-based estimators introduced in Section~\ref{sec:background}, which together form the standard reference family for intrinsic-dimension estimation~\citep{levina2004maximum,facco2017estimating,pope2021intrinsic}.
All three are agnostic to how the data was generated and infer the dimension from the local geometry of a point cloud, so we run them in the output space, on the valid joint configurations the model produces for each condition, rather than on the learned score.

\paragraph{Local PCA.}
The estimator of \citet{fukunaga1971algorithm} performs principal-component analysis within each sample's neighbourhood and counts how many leading components are needed to capture a fixed fraction of the local variance, taking that count as the dimension.
It is controlled by the neighbourhood size, which sets the scale at which the manifold is treated as locally flat, and by a cumulative-variance threshold, which determines how much of the local variance the leading components must capture.
The method is simple and directly geometric, but assumes the neighbourhood is well approximated by a linear subspace, so it is most reliable on weakly curved, well-sampled manifolds.

\paragraph{Maximum-likelihood estimation.}
\looseness=-1
The estimator of \citet{levina2004maximum} treats the neighbours of a point as a Poisson process on the manifold and derives a maximum-likelihood estimate of the dimension from the logarithms of the ratios of successive neighbour distances, aggregated across points in the bias-corrected form of \citet{pope2021intrinsic}.
Its one hyperparameter is the neighbourhood size, which also fixes how many distance ratios enter each local estimate.
It returns a smooth non-integer estimate and is the standard estimator for the intrinsic dimension of image datasets~\citep{pope2021intrinsic}, though it carries a downward bias that grows with the true dimension and cannot represent the discrete $\hat{d}=0$ case.

\paragraph{TwoNN.}
The estimator of \citet{facco2017estimating} uses only the two nearest neighbours of each point, exploiting that under a locally uniform density, the ratio of the second to the first neighbour distance follows a distribution whose shape is determined solely by the dimension, which yields a closed-form per-point estimate.
It is effectively parameter-free and the most local of the three, which makes it robust to curvature but reliant on the local-uniformity assumption and sensitive to the limited information in two distances.

The baselines measure the spread of the generated configurations in the output space, so they cannot separate the continuous solution manifold from the discrete inverse-kinematics branches that contribute to the same local spread, and none is defined for the $0$-dimensional UR5 $\LieGroupSE{3}$ case.
The score-based estimator instead reads the geometry of the learned score on the manifold itself rather than the density of samples around it.
Section~\ref{app:id-full-results} reports the hyperparameter settings and the full numerical comparison.

\subsection{Full numerical results and score spectra}
\label{app:id-full-results}

Table~\ref{tab:id-full} reports the recovered intrinsic dimension for every constraint--manipulator condition and every estimator we consider, extending the two-baseline summary of Figure~\ref{fig:id_estimation}.
Alongside Score-SVD and Score-SVD$^{\dagger}$, we evaluate the three neighbourhood-based baselines of Section~\ref{sec:background}, namely the maximum-likelihood estimator of \citet{levina2004maximum} at two neighbourhood sizes, local PCA \citep{fukunaga1971algorithm} at the $90\%$ and $99\%$ cumulative-variance thresholds, and the parameter-free TwoNN ratio of \citet{facco2017estimating}.
The neighbourhood baselines act in the output space, so we run them on the valid joint configurations drawn from the model for each condition, whereas the score-based estimators read the geometry of the learned score on the manifold itself, following the sample-selection procedure of Appendix~\ref{app:discrete}.
Every entry is the median over targets.

The two estimator families separate clearly.
The neighbourhood baselines recover the analytical dimension only on the lower-dimensional conditions and drift beyond rounding error as the intrinsic dimension grows.
Local MLE and local PCA at the $90\%$ threshold are biased low and match only once the dimension is small enough, except on the $\LieGroupSE{3}$ conditions, where they overshoort. 
TwoNN is noisier and generally biased high, falling within rounding error on a single condition per robot.
Raising the PCA threshold to $99\%$ trades one failure mode for another, recovering many more conditions, including several high-dimensional ones, but overshooting on the lowest-dimensional, highest-codimension conditions, most severely on the two $\LieGroupSE{3}$ conditions.
Score-SVD, in contrast, recovers the analytical dimension exactly on every one of the thirteen positive-dimensional conditions, and Score-SVD$^{\dagger}$ additionally returns the correct $\hat{d} = 0$ on the discrete UR5 $\LieGroupSE{3}$ case, where standard Score-SVD reports an incorrect $\hat{d} = 3$.

\begin{table}[t]
\centering
\small
\setlength{\tabcolsep}{6pt}
\renewcommand{\arraystretch}{1.15}
\caption{\textbf{Intrinsic-dimension estimates across all fourteen constraint--manipulator conditions.}
Recovered dimension $\smash{\hat{d}}$ for every estimator, with the analytical intrinsic dimension of the joint-space solution manifold in the ID column.
Neighbourhood baselines (local MLE \citep{levina2004maximum} at neighbourhood size $\mathrm{NN}$, local PCA \citep{fukunaga1971algorithm} at a cumulative-variance threshold, and TwoNN \citep{facco2017estimating}) run on valid model configurations in joint space; the score-based estimators run on the constraint manifold.
Entries are medians over targets; those within rounding error of the analytical dimension are shown in \textbf{bold}.}
\label{tab:id-full}
\begin{tabular}{@{}llcccccccc@{}}
\toprule
& & & \multicolumn{2}{c}{Local MLE} & \multicolumn{2}{c}{Local PCA} & & \multicolumn{2}{c}{Score-based} \\
\cmidrule(lr){4-5}\cmidrule(lr){6-7}\cmidrule(l){9-10}
& Constraint & ID & $\mathrm{NN}{=}2$ & $\mathrm{NN}{=}10$ & $90\%$ & $99\%$ & TwoNN & SVD & SVD$^{\dagger}$ \\
\midrule
\multirow{7}{*}{\rotatebox[origin=c]{90}{\textsc{Franka}}}
 & \textsc{Plane}                          & 6 & $4.2$          & $4.2$          & $4$          & $\mathbf{6}$ & $3.4$          & $\mathbf{6}$ & $\mathbf{6}$ \\
 & \textsc{Line}                           & 5 & $4.2$          & $3.7$          & $4$          & $\mathbf{5}$ & $\mathbf{5.3}$ & $\mathbf{5}$ & $\mathbf{5}$ \\
 & \textsc{Position}                       & 4 & $\mathbf{3.6}$ & $3.2$          & $3$          & $\mathbf{4}$ & $5.0$          & $\mathbf{4}$ & $\mathbf{4}$ \\
 & \textsc{Orientation}                    & 4 & $\mathbf{3.6}$ & $3.2$          & $3$          & $\mathbf{4}$ & $5.0$          & $\mathbf{4}$ & $\mathbf{4}$ \\
 & \textsc{Plane} $+$ \textsc{Orientation} & 3 & $\mathbf{2.9}$ & $\mathbf{2.5}$ & $\mathbf{3}$ & $4$          & $4.0$          & $\mathbf{3}$ & $\mathbf{3}$ \\
 & \textsc{Line} $+$ \textsc{Orientation}  & 3 & $\mathbf{2.9}$ & $\mathbf{2.6}$ & $\mathbf{3}$ & $4$          & $4.0$          & $\mathbf{3}$ & $\mathbf{3}$ \\
 & $\LieGroupSE{3}$                        & 1 & $3.7$          & $2.0$          & $2$          & $5$          & $5.3$          & $\mathbf{1}$ & $\mathbf{1}$ \\
\midrule
\multirow{7}{*}{\rotatebox[origin=c]{90}{\textsc{UR5}}}
 & \textsc{Plane}                          & 5 & $3.3$          & $3.4$          & $3$          & $\mathbf{5}$ & $2.4$          & $\mathbf{5}$ & $\mathbf{5}$ \\
 & \textsc{Line}                           & 4 & $3.2$          & $3.2$          & $3$          & $5$          & $\mathbf{3.5}$ & $\mathbf{4}$ & $\mathbf{4}$ \\
 & \textsc{Position}                       & 3 & $\mathbf{2.9}$ & $\mathbf{2.6}$ & $\mathbf{3}$ & $\mathbf{3}$ & $4.0$          & $\mathbf{3}$ & $\mathbf{3}$ \\
 & \textsc{Orientation}                    & 3 & $\mathbf{2.8}$ & $2.3$          & $\mathbf{3}$ & $\mathbf{3}$ & $3.9$          & $\mathbf{3}$ & $\mathbf{3}$ \\
 & \textsc{Plane} $+$ \textsc{Orientation} & 2 & $\mathbf{2.0}$ & $\mathbf{1.7}$ & $\mathbf{2}$ & $\mathbf{2}$ & $2.8$          & $\mathbf{2}$ & $\mathbf{2}$ \\
 & \textsc{Line} $+$ \textsc{Orientation}  & 2 & $\mathbf{2.1}$ & $\mathbf{1.7}$ & $\mathbf{2}$ & $3$          & $2.9$          & $\mathbf{2}$ & $\mathbf{2}$ \\
 & $\LieGroupSE{3}$                        & 0 & $4.8$          & $4.0$          & $4$          & $6$          & $7.6$          & $3$          & $\mathbf{0}$ \\
\bottomrule
\end{tabular}
\end{table}

\begin{figure}[t]
\centering
\includegraphics[width=\linewidth]{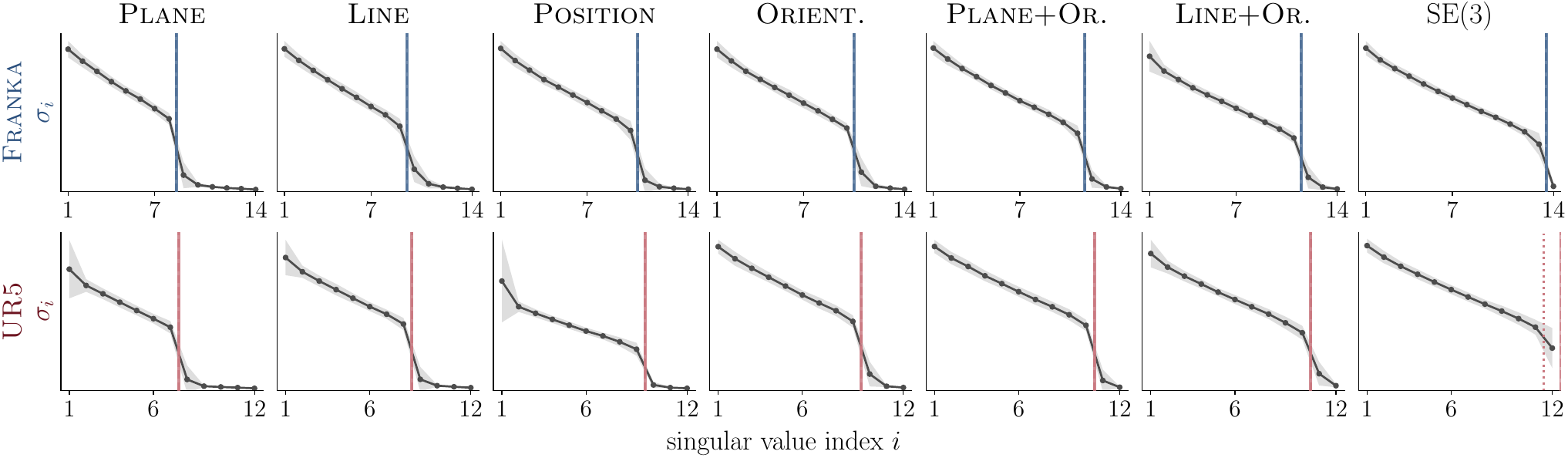}
\vspace{-4mm}
\caption{
\looseness=-1
\textbf{Score-matrix singular spectra.}
Mean descending singular spectrum of the score matrix $S$ at $t = 15$ and $K = 50$ (grey line, shaded $\pm$ one standard deviation over the pooled samples), for \textcolor{frankaBlue}{Franka} (top) and \textcolor{ur5Red}{UR5} (bottom), one constraint per column.
Vertical markers give the gap position $\smash{n - \hat{d}}$ implied by the
\@ \linedashed{scoreBlue} / \linedashed{scoreRed}~analytical intrinsic dimension,
\@ \linedotted{scoreBlue} / \linedotted{scoreRed}~Score-SVD,
\@ and \linethick{dagBlue} / \linethick{dagRed}~Score-SVD$^{\dagger}$.
The three markers fall at the spectral gap and coincide on every positive-dimensional condition. On the discrete UR5 $\LieGroupSE{3}$ case the spectrum has no clear separation, Score-SVD selects an incorrect interior gap, while Score-SVD$^{\dagger}$ places the gap after $\sigma_n$ to recover $\smash{\hat{d} = 0}$.
}
\label{fig:id-score-spectra}
\end{figure}

\looseness=-1
The behaviour of these estimators is visible in the score-matrix spectrum itself.
Figure~\ref{fig:id-score-spectra} shows the mean descending singular spectrum of $S$ for every condition, pooled over the samples and targets of the procedure of Appendix~\ref{app:discrete}.
On every positive-dimensional condition, the spectrum splits into two parts. A block of large singular values spans the normal directions of the manifold, while a near-zero tail, whose width equals the intrinsic dimension, spans the tangent directions that the score cannot resolve.
The gap between these two blocks is precisely what Score-SVD identifies. The markers in the figure are read from this averaged spectrum at index $n - \hat{d}$, and they coincide with the analytical value on all thirteen positive-dimensional conditions.
The discrete UR5 $\LieGroupSE{3}$ case is the only exception, and the figure makes its difficulty concrete.
Because every ambient direction is normal, no near-zero tail exists and the spectrum decays smoothly without a clear gap, so the standard rule is forced to select whichever interior drop happens to be largest.
On this averaged spectrum, the largest drop is the final one, $\sigma_{n-1} \to \sigma_n$, so the standard Score-SVD marker lands at $\hat{d} = 1$, whereas the per-target median reported in Table~\ref{tab:id-full} is $\hat{d} = 3$.
These two estimates disagree precisely because no spectral gap exists and the largest per-sample drop scatters across the interior, making the estimate sensitive to how it is aggregated.
Only Score-SVD$^{\dagger}$, which places the gap after the final singular value $\sigma_n$, recovers the correct $\hat{d} = 0$.

%% file: sections/appendix/latent-interpolation.tex
\section{Latent Interpolation}
\label{app:latent-interpolation}

This section expands the latent-interpolation study of Section~\ref{sec:interpolation}.
We first report the full quantitative results across all fourteen constraint--manipulator conditions and compare straight-line against spherical interpolation in the latent space (Appendix~\ref{app:interp-quantitative}), and then discuss the conditions that push our method to its limits, including the disconnected $\LieGroupSE{3}$ solution set of the UR5 (Appendix~\ref{app:interp-disconnected}).
We then give the formal definitions and derivations behind the local-flattening diagnostics of Section~\ref{sec:interpolation} (Appendix~\ref{app:interp-definitions}) and report additional analyses: the dependence on the intermediate noise level and the growth of error with endpoint separation (Appendix~\ref{app:interp-extended}).

Unless stated otherwise, all results use the same setup as Section~\ref{sec:interpolation}: for each condition we draw $30$ task-space targets, and at each target select five endpoint pairs of IK solutions indexed by a separation level $\Delta \in \{10, 25, 50, 75, 100\}$.
For twelve of the fourteen conditions, we sweep one free task-space coordinate at the target (a free position coordinate, a rotation about the vertical axis for the Franka under \textsc{Position}, or the redundant joint for the Franka under $\LieGroupSE{3}$), sort the resulting IK solutions along the sweep, and take the pair at sweep percentiles $50 \mp \Delta/2$.
For the UR5 under \textsc{Position}, we take the pair whose joint-space distance is closest to $\Delta\%$ of the solution-set diameter, and for the UR5 under $\LieGroupSE{3}$ the pair at the $\Delta$-th percentile of the pairwise distances between IK branches.
The joint-space separation grows monotonically with $\Delta$ in every condition.
Both endpoints are inverted to an intermediate noise level $\bar t$ of the partial DDIM process~\eqref{eq:ddim_inv}, their latent codes are interpolated, and the interpolants are decoded back to joint configurations, whose forward-kinematics error to the analytical constraint set $\mathcal{S}_\mathbf{c}$ is then measured.

\subsection{Quantitative results}
\label{app:interp-quantitative}

Table~\ref{tab:interp-full} reports, for every condition, the median forward-kinematics error of the decoded paths with its $95\%$ bootstrap confidence interval, separately for position- and orientation-constrained coordinates.
We compare both straight-line interpolation as described in Section~\ref{sec:interpolation} and spherical linear interpolation (slerp), which follows the great-circle arc between the two latent codes on a sphere of fixed radius;  each is evaluated at a per-condition noise level $\bar t \in \{0, 10, \dots, 80, 85, 90, 95\}$, chosen to maximize the percentage of decoded configurations within a validity tolerance, minus penalties on the mean position error, the mean orientation error for the orientation-constrained families, and the 95th-percentile end-effector step between consecutive decoded configurations (Table~\ref{tab:tbar}).

\begin{table}[t]
\centering
\small
\setlength{\tabcolsep}{4.5pt}
\renewcommand{\arraystretch}{1.3}
\caption{\textbf{Latent interpolation accuracy across all fourteen constraint--manipulator conditions.}
Median forward-kinematics error of the decoded paths for straight-line (lin) and spherical (sph) interpolation, with the $95\%$ bootstrap CI of the median given as the bracketed subscript $[\text{low},\text{high}]$; for each metric the lower of the two medians is set in \textbf{bold}.
Position error is the Euclidean distance over the constrained axes; orientation error is the full $\LieGroupSO{3}$ geodesic angle~\eqref{eq:app-so3-geodesic} for fully orientation-constrained tasks, and the angle between the constrained body-axis directions for \textsc{Line}\,$+$\,\textsc{Orientation} (Appendix~\ref{app:constraints}). Dashes mark coordinates left unconstrained.}
\label{tab:interp-full}
\begin{tabular}{@{}llcccc@{}}
\toprule
& & \multicolumn{2}{c}{Pos.\ err.\ [mm]} & \multicolumn{2}{c}{Orn.\ err.\ [$^\circ$]} \\
\cmidrule(lr){3-4}\cmidrule(l){5-6}
& Constraint & lin & sph & lin & sph \\
\midrule
\multirow{7}{*}{\rotatebox[origin=c]{90}{\textsc{Franka}}}
 & \textsc{Plane}                          & $\mathbf{0.40}_{[0.35,0.47]}$ & $0.46_{[0.36,0.54]}$ & --                   & --                   \\
 & \textsc{Line}                           & $\mathbf{1.10}_{[0.87,1.42]}$ & $1.21_{[0.91,1.47]}$ & --                   & --                   \\
 & \textsc{Position}                       & $\mathbf{0.79}_{[0.62,0.90]}$ & $0.87_{[0.70,1.00]}$          & --                            & --                            \\
 & \textsc{Orientation}                    & --                            & --                            & $\mathbf{0.38}_{[0.31,0.44]}$ & $0.43_{[0.34,0.60]}$          \\
 & \textsc{Plane} $+$ \textsc{Orientation} & $\mathbf{0.43}_{[0.34,0.51]}$ & $0.47_{[0.38,0.57]}$ & $\mathbf{0.25}_{[0.20,0.30]}$ & $0.28_{[0.23,0.34]}$ \\
 & \textsc{Line} $+$ \textsc{Orientation}  & $\mathbf{1.29}_{[0.94,1.89]}$ & $1.32_{[1.07,1.87]}$ & $\mathbf{0.23}_{[0.18,0.29]}$ & $0.24_{[0.20,0.39]}$ \\
 & $\LieGroupSE{3}$                        & $0.70_{[0.58,0.81]}$          & $\mathbf{0.68}_{[0.61,0.81]}$ & $\mathbf{0.24}_{[0.22,0.29]}$ & $0.25_{[0.21,0.31]}$          \\
\midrule
\multirow{7}{*}{\rotatebox[origin=c]{90}{\textsc{UR5}}}
 & \textsc{Plane}                          & $0.89_{[0.65,1.23]}$ & $\mathbf{0.75}_{[0.62,1.08]}$ & --                   & --                   \\
 & \textsc{Line}                           & $\mathbf{1.60}_{[1.27,2.02]}$ & $1.92_{[1.53,2.44]}$ & --                   & --                   \\
 & \textsc{Position}                       & $\mathbf{1.08}_{[0.84,1.28]}$ & $1.20_{[0.88,1.51]}$ & --                   & --                   \\
 & \textsc{Orientation}                    & --                   & --                   & $\mathbf{0.34}_{[0.28,0.44]}$ & $0.35_{[0.29,0.50]}$ \\
 & \textsc{Plane} $+$ \textsc{Orientation} & $\mathbf{0.77}_{[0.59,1.12]}$ & $1.08_{[0.79,1.27]}$ & $\mathbf{0.45}_{[0.35,0.52]}$ & $0.52_{[0.37,0.61]}$ \\
 & \textsc{Line} $+$ \textsc{Orientation}  & $\mathbf{2.41}_{[1.98,3.14]}$ & $3.24_{[2.41,4.22]}$ & $\mathbf{1.00}_{[0.79,1.26]}$ & $1.22_{[0.95,1.61]}$ \\
 & $\LieGroupSE{3}$                        & $\mathbf{1.61}_{[1.42,1.70]}$ & $2.30_{[2.08,2.44]}$ & $\mathbf{0.35}_{[0.31,0.38]}$ & $0.47_{[0.44,0.52]}$ \\
\bottomrule
\end{tabular}
\end{table}

\begin{table}[t]
\centering
\small
\setlength{\tabcolsep}{5pt}
\caption{\textbf{Intermediate noise level $\bar t$ selected for each condition}, given as straight-line\,/\,spherical.
The selection score is the percentage of valid decoded configurations minus $0.5$ times the mean position error in mm, $1.0$ times the mean orientation error in degrees (orientation-constrained families only), and $0.1$ times the 95th-percentile end-effector step in mm.
A configuration is valid if each constrained position axis is within $50$~mm (\textsc{Plane}, \textsc{Plane}\,$+$\,\textsc{Orientation}, $\LieGroupSE{3}$), $80$~mm (\textsc{Line}, \textsc{Line}\,$+$\,\textsc{Orientation}) or $100$~mm (\textsc{Position}), and the orientation error is within $0.3$~rad.}
\label{tab:tbar}
\begin{tabular}{@{}lccccccc@{}}
\toprule
 & \textsc{Plane} & \textsc{Line} & \textsc{Position} & \textsc{Orient.} & \textsc{Plane\,+\,Or.} & \textsc{Line\,+\,Or.} & $\LieGroupSE{3}$ \\
\midrule
\textsc{Franka} & $85\,/\,90$ & $60\,/\,60$ & $95\,/\,95$ & $50\,/\,50$ & $80\,/\,90$ & $60\,/\,85$ & $95\,/\,95$ \\
\textsc{UR5}    & $50\,/\,60$ & $90\,/\,80$ & $95\,/\,95$ & $60\,/\,60$ & $60\,/\,60$ & $90\,/\,70$ & $95\,/\,95$ \\
\bottomrule
\end{tabular}
\end{table}

Across all conditions the decoded paths stay close to the constraint manifold: the median position error of linear interpolation ranges from $0.40$ to $2.41$~mm and the median orientation error from $0.23$ to $1.00^\circ$. Most confidence intervals are narrow relative to their medians, although the hardest position-constrained cases---notably the UR5 under \textsc{Line} $+$ \textsc{Orientation}---have appreciably wider intervals.
Both interpolation methods are statistically comparable on most conditions, with overlapping confidence intervals; straight-line interpolation has lower median values on the hardest conditions, especially on the UR5 under $\LieGroupSE{3}$.

\newcommand{\failph}[1]{%
  \fbox{\begin{minipage}[c][0.205\linewidth][c]{0.205\linewidth}%
    \centering\scriptsize\textcolor{gray}{#1}%
  \end{minipage}}}

\begin{figure}[t]
\centering
\includegraphics[width=0.9\linewidth]{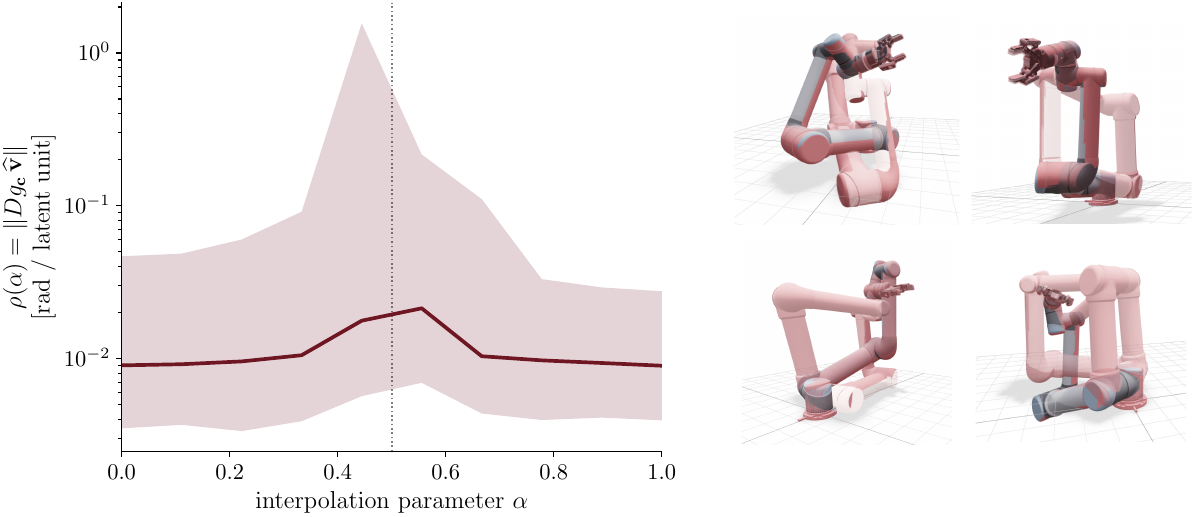}
\caption{\textbf{Disconnected-branch interpolation on the UR5 under $\LieGroupSE{3}$.}
\emph{Left:} directional sensitivity $\rho(\alpha)$ along the interpolation path; the solid line is the median over endpoint pairs, the shaded region the $2.5$--$97.5$ percentile band, and the dotted vertical line marks the path midpoint. The median stays small and flat across the whole path, while the upper percentile rises by more than an order of magnitude near the midpoint: each pair switches branches over a narrow interval of $\alpha$, so at any fixed $\alpha$ only the pairs mid-switch exhibit the large joint-space velocity, yet the decoded poses stay close to the constraint set throughout.
\emph{Right:} four interpolations for the widest (maximum-separation, $\Delta = 100$) endpoint pairs.
Each panel shows the generated configurations along the straight-line interpolation; the end-effector approximately holds the fixed $\LieGroupSE{3}$ target while the arm switches between distinct IK branches, crossing from two up to four branches in a single decoded path.
}
\label{fig:interp-se3-branch}
\end{figure}

\subsection{Disconnected solution sets and branch switching}
\label{app:interp-disconnected}

\begin{figure}[t]
\centering
\includegraphics[width=\linewidth]{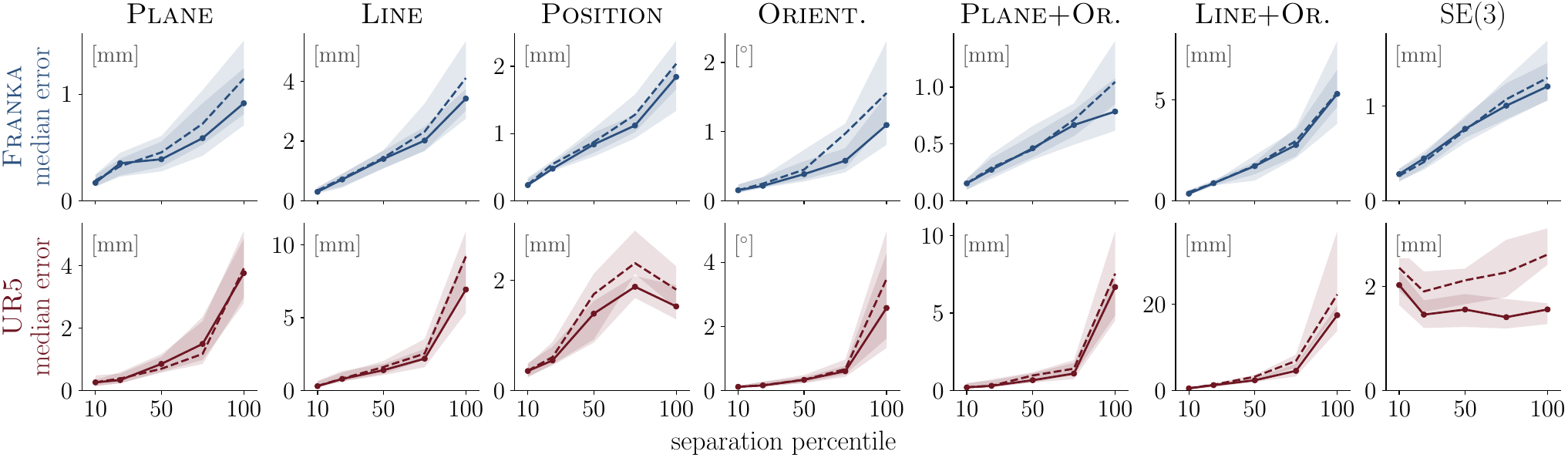}
\vspace{-4mm}
\caption{\textbf{Median constraint error versus endpoint separation.}
Median error of the decoded path, with a shaded $95\%$ bootstrap CI of the median, as a function of the separation level $\Delta$ of the endpoint pair, for straight-line (solid) and spherical (dashed) interpolation, each at its per-condition noise level $\bar t$; \textcolor{frankaBlue}{Franka} (top) and \textcolor{ur5Red}{UR5} (bottom). Panels report position error in millimetres, except the orientation-only \textsc{Orientation} column, which reports the geodesic angle in degrees ($^\circ$); for the combined position\,$+$\,orientation constraints the orientation residuals are summarized in Table~\ref{tab:interp-full} rather than plotted. For the continuous conditions the error generally increases with endpoint separation; the UR5 under $\LieGroupSE{3}$ is the exception, staying nearly flat because the decoded path switches branches while holding close to the fixed pose.
}
\label{fig:interp-gap-error}
\end{figure}

\paragraph{Disconnected solution sets (UR5 under $\LieGroupSE{3}$).}
For the non-redundant UR5 under a full $\LieGroupSE{3}$ pose constraint, the feasible configurations form a finite set of isolated IK branches, and no continuous pose-preserving path connects two branches in joint space.
This is the clearest discrete case in our study: the solution set is zero-dimensional, a finite collection of isolated IK branches.
It therefore reveals what a straight latent path decodes to when no continuous on-constraint curve exists to follow.
We find that the decoded path does not form a continuous feasible curve between branches (which a finite solution set could not support), but instead stays close to the $\LieGroupSE{3}$ constraint while concentrating the joint-space motion into one or more narrow intervals of $\alpha$.
Its constraint error stays low (Table~\ref{tab:interp-full}, $1.61$~mm / $0.35^\circ$) and, unlike the continuous conditions, does not grow with endpoint separation but stays flat at the widest pairs (Figure~\ref{fig:interp-gap-error}), consistent with rapid switching between branches rather than a long infeasible bridge through joint space.
Figure~\ref{fig:interp-se3-branch} makes the switching explicit through the directional sensitivity $\rho$ (Appendix~\ref{app:interp-definitions}): its median is small and flat along the whole path, while its upper percentile rises by more than an order of magnitude near the midpoint, exactly where the decoded configuration crosses from one branch to the next.

\paragraph{Hardest conditions.}
Our approach is weakest on the UR5 \textsc{Line} $+$ \textsc{Orientation} condition, the tightest continuous constraint we study on the non-redundant arm, where it has the largest median errors ($2.41$~mm / $1.00^\circ$) in Table~\ref{tab:interp-full} and the lowest tail tangent overlap, $\tau^{\mathbf q}_{\rm p5}=0.754$ (Table~\ref{tab:sensitivity}), although this value remains well above the matched random-direction baseline.
Error also concentrates on the small number of targets whose solution set is not fully covered by the training distribution: the widest endpoint pairs at such targets are where the decoded path drifts farthest from the constraint set, which is consistent with the additional capacity required at high noise levels being unavailable when the target lies at the edge of the trained region.
Even there, the drift stays within a few centimetres.

\subsection{Definitions of \texorpdfstring{$\rho$, $\nu$, $\tau^{\mathbf q}$, and the random-direction baseline}{rho, nu, tau, and the random-direction baseline}}
\label{app:interp-definitions}

This section gives the full definitions and derivations of the local-flattening diagnostics of Section~\ref{sec:interpolation} (Table~\ref{tab:sensitivity}).
For a fixed constraint $\mathbf c$, we invert an endpoint pair to the intermediate noise level $\bar t$, giving latent codes $\mathbf x_a,\mathbf x_b$.
We interpolate these along the straight path $\mathbf x(\alpha)=(1-\alpha)\,\mathbf x_a+\alpha\,\mathbf x_b$, with unit direction $\widehat{\mathbf v}=(\mathbf x_b-\mathbf x_a)/\Norm{\mathbf x_b-\mathbf x_a}$.
The DDIM map $g_{\mathbf c}$~\eqref{eq:ddim_step} from $\bar t$ to $0$ maps each interpolant to a joint configuration $\mathbf q(\alpha)=g_{\mathbf c}(\mathbf x(\alpha))$. Its Jacobian $Dg_{\mathbf c}(\mathbf x)\in\Real{}^{n\times 2n}$, where the latent dimension $2n$ ($12$ for the UR5, $14$ for the Franka) is the joint count doubled by the sine--cosine encoding, is taken by automatic differentiation at fixed $\mathbf c$ and gives the decoded joint-space direction $\mathbf u(\alpha)=Dg_{\mathbf c}(\mathbf x(\alpha))\,\widehat{\mathbf v}$. Both $\mathbf q(\alpha)$ and $Dg_{\mathbf c}$ use the same conditioning vector and the same $100$-step DDIM map that produce the decoded paths of Table~\ref{tab:interp-full}.

\paragraph{Directional sensitivity and joint-space speed.}
The directional sensitivity is the norm of the decoded direction, and the joint-space speed rescales it by the endpoint separation,
\begin{equation}
\rho(\alpha)=\Norm{\mathbf u(\alpha)}_2,
\qquad
\nu(\alpha)=\Norm{\mathbf x_b-\mathbf x_a}\,\rho(\alpha),
\end{equation}
where $\rho$ has units of radians per latent unit and $\nu$ units of radians per unit interpolation parameter; a small $\rho$ means a unit latent step produces only a small, smoothly varying change in joint space.

\begin{figure}[t]
\centering
\includegraphics[width=\linewidth]{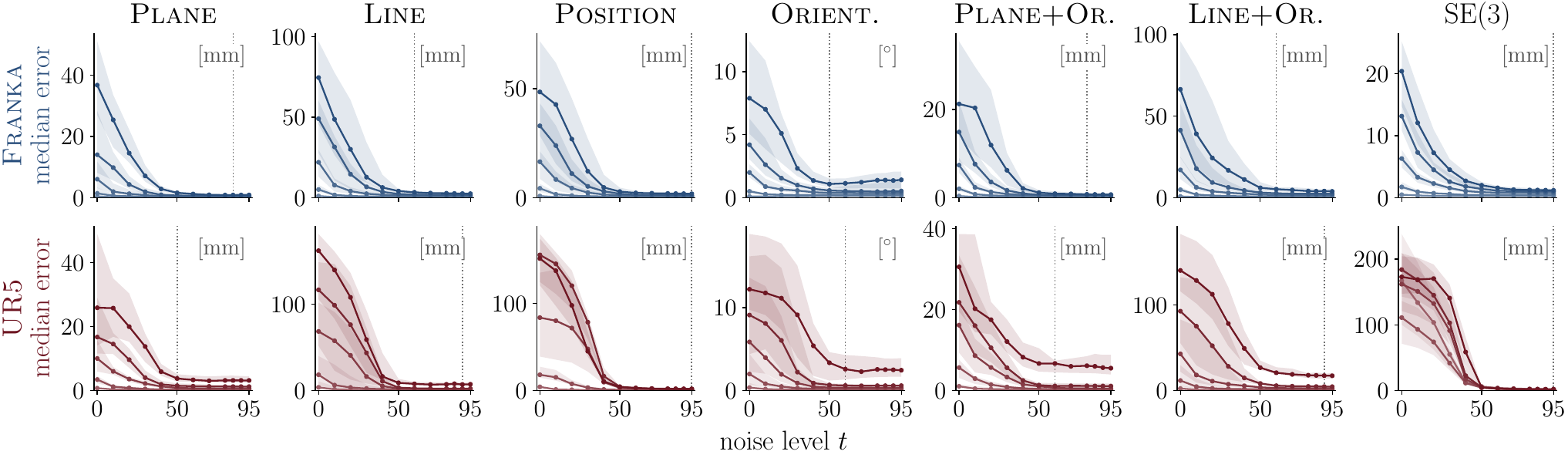}
\vspace{-4mm}
\caption{\textbf{Constraint error versus intermediate noise level.}
Constraint error of the decoded straight-line paths as a function of the noise level $\bar t$ to which the endpoints are inverted,
as in the bottom rows of Figure~\ref{fig:error-svd-tmid-lineplots-main} with a separate scale for each panel.
Within each panel the five curves are endpoint pairs at increasing joint-space separation (lighter to darker: separation level $\Delta = 10$ to $100$), each the median over targets with a shaded $95\%$ bootstrap CI. Panels report position error in millimetres, except the orientation-only \textsc{Orientation} column, which reports the geodesic angle in degrees ($^\circ$). The dotted vertical line marks the $\bar t$ used for Table~\ref{tab:interp-full}. The error is largest for the widest pairs and at low noise, and all separations converge to a low floor as the noise level increases. For the combined position\,$+$\,orientation constraints these panels report position error only; the corresponding orientation residuals are summarized in Table~\ref{tab:interp-full} rather than plotted.
}
\label{fig:interp-noise-ablation}
\end{figure}

\begin{figure}[t]
\centering
\includegraphics[width=\linewidth]{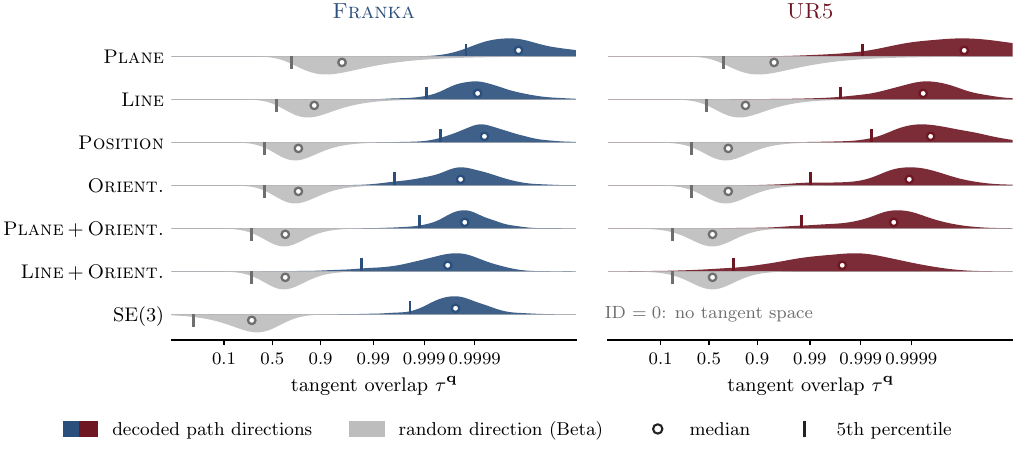}
\vspace{-4mm}
\caption{\textbf{Distribution of the tangent overlap against the random-direction baseline.}
For each condition, the upper half of the row (in the manipulator's colour; \textcolor{frankaBlue}{Franka} left, \textcolor{ur5Red}{UR5} right) is a kernel density estimate of $\tau^{\mathbf q}$ over all $(\textrm{pair},\alpha)$ samples of the decoded straight-line paths, and the lower half (grey) is the density of $\Norm{\Pi\,\mathbf r}_2$ for a uniformly random direction $\mathbf r$; both are drawn on a logit axis and scaled to the same height.
Circles mark the medians and bars the fifth percentiles, which are the values reported in Table~\ref{tab:sensitivity}.
The UR5 under $\LieGroupSE{3}$ has no tangent space and is left empty.
}
\label{fig:interp-tau-dist}
\end{figure}

\paragraph{Tangent overlap and the random-direction baseline.}
Let $h_{\mathbf c}$ be the constraint-error map, with $\mathcal M_{\mathbf c}=\{\mathbf q:h_{\mathbf c}(\mathbf q)=\Zero\}$ and Jacobian $Dh_{\mathbf c}(\mathbf q)\in\Real{}^{k\times n}$ (Appendix~\ref{app:smooth-manifolds}). The tangent overlap is the fraction of the decoded direction lying in the null space of the constraint Jacobian,
\begin{equation}
\tau^{\mathbf q}(\alpha)=\frac{\Norm{\Pi_{\ker Dh_{\mathbf c}(\mathbf q(\alpha))}\,\mathbf u(\alpha)}_2}{\Norm{\mathbf u(\alpha)}_2},
\end{equation}
where $\Pi_{\ker Dh_{\mathbf c}(\mathbf q)}$ is the orthogonal projector~\eqref{eq:app-projector} onto $\ker Dh_{\mathbf c}(\mathbf q)$, which is the tangent space $\mathcal T_{\mathbf q(\alpha)}\mathcal M_{\mathbf c}$~\eqref{eq:app-tangent} when $\mathbf q(\alpha)$ lies on $\mathcal M_{\mathbf c}$; a value near one then means the motion follows along the manifold and leaves the constraint error unchanged to first order. 
For comparison, let $\mathbf r$ be a random unit vector distributed uniformly on the unit sphere $S^{n-1}$ of joint space.
For any fixed rank-$(n-k)$ orthogonal projector $\Pi$, the squared projected length $\Norm{\Pi\,\mathbf r}_2^2$ follows a $\mathrm{Beta}\big(\tfrac{n-k}{2},\tfrac{k}{2}\big)$ distribution with mean $\Trace{\Pi}/n=(n-k)/n$.
We compare the median and fifth percentile of $\tau^{\mathbf q}$ with the median $\tau^{\mathbf q}_{\rm rand,med}$ and fifth percentile $\tau^{\mathbf q}_{\rm rand,p5}$ of $\Norm{\Pi\,\mathbf r}_2$, the quantiles of the same order under the random-direction model.
For each condition we report the median over all $(\textrm{pair},\alpha)$ samples, $\tau^{\mathbf q}_{\rm med}$, the fifth percentile $\tau^{\mathbf q}_{\rm p5}$, and the relative improvement of the median over the baseline,
\begin{equation}
\Delta^{\mathbf q}_{\rm rel}=(\tau^{\mathbf q}_{\rm med}-\tau^{\mathbf q}_{\rm rand,med})\,/\,\tau^{\mathbf q}_{\rm rand,med};
\end{equation}
the same medians and percentiles define $\rho_{\rm med}$ and $\nu_{\rm med}$ in Table~\ref{tab:sensitivity}. For the UR5 under $\LieGroupSE{3}$ the constraint is full ($k=n$), so $\ker Dh_{\mathbf c}$ is zero-dimensional and both $\tau^{\mathbf q}$ and its random-direction baseline are zero by construction; this condition is therefore excluded from the tangent-overlap comparison and treated through the disconnected-topology analysis above (Figure~\ref{fig:interp-se3-branch}).
Figure~\ref{fig:interp-tau-dist} shows the full distribution of $\tau^{\mathbf q}$ for every condition against that of the random-direction baseline.

\subsection{Extended analysis}
\label{app:interp-extended}

\paragraph{Dependence on the intermediate noise level.}
Figure~\ref{fig:interp-noise-ablation} repeats the error-versus-noise panels of Figure~\ref{fig:error-svd-tmid-lineplots-main} with a separate scale for each panel and marks the noise level used for Table~\ref{tab:interp-full}.
The error is large at low noise and largest for the widest pairs, but every separation falls onto a common low floor as the noise level increases.
This is consistent with our observation in Section~\ref{sec:interpolation}, Figure~\ref{fig:error-svd-tmid-lineplots-main} that the interpolation error collapses only once the intermediate representation has expanded beyond the analytical intrinsic dimension.
The one clear exception is the orientation-only \textsc{Orientation} condition on the Franka, whose error is already small at low noise and rises slightly as more noise is injected into an already-near-feasible range; this is why we use a comparatively low noise level for this specific constraint.

\paragraph{Growth of error with endpoint separation.}
Figure~\ref{fig:interp-gap-error} reports the median constraint error as a function of the separation level $\Delta$ of the endpoint pair. 

\paragraph{Gaussian-score control.}
To test whether the low interpolation error requires the learned score, we consider replacing it with the exact score of a Gaussian fitted to the IK solutions of each target.
The outcome does not depend on the Gaussian: its predicted noise is affine in $\mathbf x_t$, so every DDIM step, and hence the whole inversion and decoding pipeline, is an affine map, which sends a straight latent path to a straight line between the two endpoints in the sine--cosine encoding.
This straight line, computed without evaluating the network, has a larger median position error than the learned model on all twelve position-constrained conditions, by a median factor of $12$, and a larger median orientation error on seven of the eight orientation-constrained conditions (median factor $7$); the exception is the UR5 under \textsc{Line}~$+$~\textsc{Orientation} ($0.90^\circ$ against $1.00^\circ$).

%% file: sections/appendix/applications.tex
\section{Applications of the Recovered Geometry}
\label{app:applications}

This section expands the two applications of Section~\ref{sec:applications}.
We first state the score-projection algorithm, its evaluation protocol and its per-condition results, together with the diagnostics that separate the projected from the raw score (Appendix~\ref{app:projection}), and then report the seen--unseen intrinsic-dimension gap on all fourteen conditions and its dependence on training length (Appendix~\ref{app:convergence}).

\subsection{Score projection onto the estimated normal bundle}
\label{app:projection}

\paragraph{Algorithm.}
Algorithm~\ref{alg:score-projection} states the iteration of~\eqref{eq:score-svd-update} as run in our experiments.
The iterate $\mathbf{x}$ is a noise-free point in the standardized sine--cosine space of Appendix~\ref{app:training}; the score is read at the forward-kernel mean $\sqrt{\bar\alpha_t}\,\mathbf{x}$ with no noise added, so the noise level $t$ selects the smoothed score field that is consulted, and nothing in the loop is denoised.
The normal basis $U_{\hat{k}}$ is the Score-SVD$^{\dagger}$ basis of Algorithm~\ref{alg:score-svd-dagger} at the current iterate: the rank $\hat{k}$ is fixed at its value at the start point and only the basis is re-estimated, before iterations $0, R, 2R, \dots$, so a refresh never changes the rank.
With $M = 20$ and $R = 5$ the projected update costs $4K + M = 220$ network evaluations per configuration, against $20$ for the raw score.

\begin{algorithm}[t]
\caption{Score projection of a configuration $\mathbf{q}_0$ onto the learned constraint manifold. $U_{\hat{k}}$ and $\hat{k}$ are read from the Score-SVD$^{\dagger}$ spectrum of Algorithm~\ref{alg:score-svd-dagger}; $\hat{k}$ is fixed at the start point.}
\label{alg:score-projection}
\begin{algorithmic}[1]
\Require conditional score model $s_\theta(\cdot,\cdot,\mathbf{c})$; constraint $\mathbf{c}$; start configuration $\mathbf{q}_0$; noise level $t$; perturbation count $K$; step size $\gamma$; iterations $M$; refresh period $R$
\State $\mathbf{x} \gets \phi(\mathbf{q}_0)$ \Comment{sine--cosine encoding, standardized}
\For{$m = 0, \dots, M-1$}
  \If{$m \bmod R = 0$}
    \State $U, (\sigma_i) \gets$ left singular vectors and values of $S$ at $\mathbf{x}$ \Comment{Algorithm~\ref{alg:score-svd-dagger}, $K$ evaluations}
    \If{$m = 0$} $\hat{k} \gets \ArgMax{i=1,\dots,n}\,(\sigma_i - \sigma_{i+1})$ \Comment{rank fixed at the start, $\sigma_{n+1} = 0$} \EndIf
    \State $U_{\hat{k}} \gets$ first $\hat{k}$ columns of $U$
  \EndIf
  \State $\mathbf{s} \gets -\,\hat{\boldsymbol{\epsilon}}_\theta(\sqrt{\bar\alpha_{t}}\,\mathbf{x},\, t,\, \mathbf{c}) \,/\, \sqrt{1-\bar\alpha_{t}}$ \Comment{score at the forward-kernel mean; no noise}
  \State $\mathbf{x} \gets \mathbf{x} + \gamma\,\frac{1-\bar\alpha_{t}}{\sqrt{\bar\alpha_{t}}}\; U_{\hat{k}} U_{\hat{k}}^{\T}\, \mathbf{s}$ \Comment{update~\eqref{eq:score-svd-update}}
\EndFor
\State \Return $\mathbf{q} \gets \phi^{-1}(\mathbf{x})$ \Comment{$\atantwo$ decode of each $(\sin,\cos)$ pair}
\end{algorithmic}
\end{algorithm}

\paragraph{Protocol.}
We use the two models of Appendix~\ref{app:training} and, for each of the fourteen conditions, $30$ held-out targets drawn as in Appendix~\ref{app:discrete}.
All runs use $t = 15$, $K = 50$, $M = 20$, $R = 5$ and $\gamma = 1$; the step size was chosen once per robot on a separate validation split of $10$ targets of \textsc{Position} and \textsc{Plane} $+$ \textsc{Orientation} over $\gamma \in \{0.25, 0.5, 1, 2\}$, where every $\gamma \le 1$ reached the same final error and $\gamma = 2$ overshot.
Two populations are projected.
P1 consists of $4$ configurations per target generated by the model with the $100$-step DDIM sampler ($120$ per condition), carrying their natural constraint error.
P2 consists of $2$ exact analytical IK solutions per target displaced off $\mathcal{M}_{\mathbf{c}}$ by the minimum-norm joint motion $Dh_{\mathbf{c}}^{+}\,\delta\,\hat{\mathbf{u}}$ that produces a task-space error of prescribed size $\delta \in \{1, 5, 10, 20\} \times 10^{-3}$ along a random unit direction $\hat{\mathbf{u}}$ ($240$ per condition); the initial error equals $\delta$ by construction.
We report $\Norm{h_{\mathbf{c}}(\mathbf{q})}$, whose components are in metres and radians, in units of $10^{-3}$, i.e.\ millimetres for position and milliradians for orientation, and ``Red.'' is the median over configurations of the per-configuration ratio of initial to final error.
Four updates run from identical starts: the projected score of Algorithm~\ref{alg:score-projection}, the raw score without $U_{\hat{k}}U_{\hat{k}}^{\T}$, the analytical Newton--Raphson step $-Dh_{\mathbf{c}}^{+}h_{\mathbf{c}}$ as a reference that reaches zero error, and a random direction of the same norm as the score step as a control, which diverges on every condition.
$h_{\mathbf{c}}$ and $Dh_{\mathbf{c}}$ enter the evaluation and the reference only.

\begin{figure}[t]
\centering
\includegraphics[width=\linewidth]{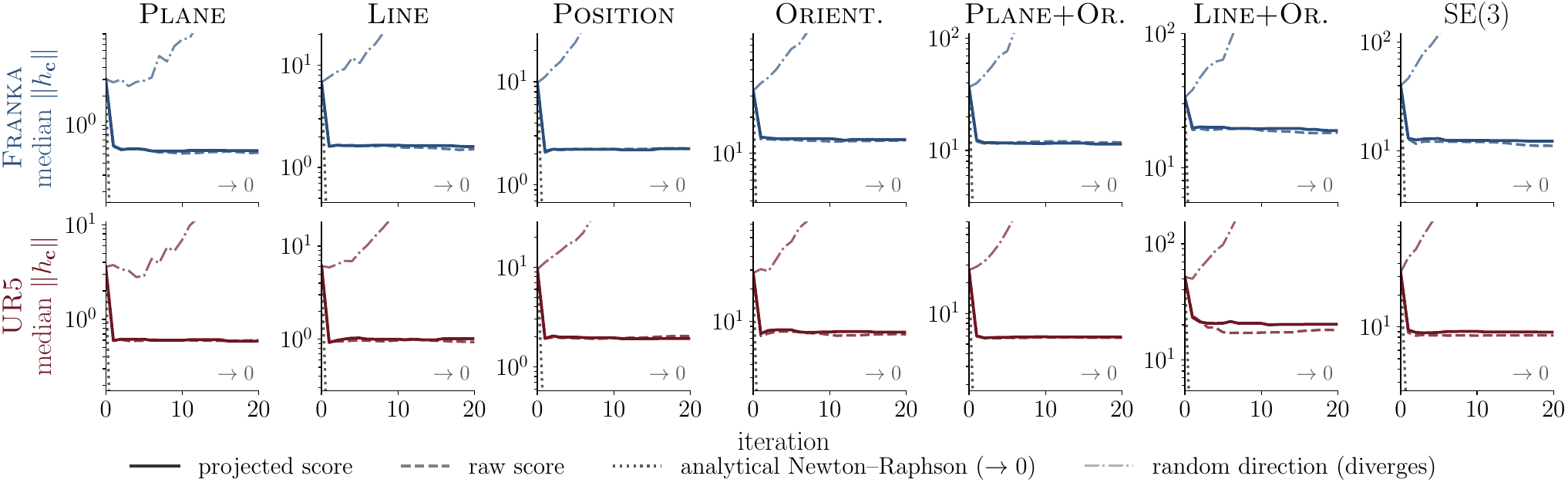}
\vspace{-4mm}
\caption{\textbf{Constraint error versus projection iteration on all fourteen conditions.}
Median $\Norm{h_{\mathbf{c}}}$ ($\times 10^{-3}$, log scale) over the $120$ generated configurations for \textcolor{frankaBlue}{Franka} (top) and \textcolor{ur5Red}{UR5} (bottom), one constraint per column: projected score (solid), raw score (dashed), analytical Newton--Raphson (grey dotted; it reaches zero and leaves the axis), and a random direction of matched norm (dash-dotted; it diverges).
Both score updates settle within about two iterations at the same value.}
\label{fig:proj-residual}
\end{figure}

\paragraph{Results.}
Figure~\ref{fig:proj-residual} shows the four updates on every condition, and Table~\ref{tab:proj-p1} gives the initial and final errors of the projected score on the generated configurations.
The median error falls on every condition, by factors of $2.6$--$6.8$ on the UR5 and $1.5$--$5.9$ on the Franka, and both score updates reach their final value within about two iterations.
Table~\ref{tab:proj-p2} gives the final error from the displaced exact solutions: every row is nearly flat, and the $\delta = 1$ starts are pulled \emph{up} to the same value on every condition except \textsc{Plane}, so the final error does not depend on the initial one.
The iteration therefore converges to the \emph{learned} manifold, and the remaining error is its offset from $\mathcal{M}_{\mathbf{c}}$, the model's conditional bias; it is largest on the orientation-bearing families, where the generated configurations also start furthest from the manifold.
Table~\ref{tab:proj-posorn} reports the same P1 configurations in the convention of Table~\ref{tab:size-sweep-compare}.
Projection lowers the mean error on every position and orientation entry; the mean position error falls to $0.7$--$2.6$~mm on nine of the twelve position-constrained conditions, and the remaining means, UR5 \textsc{Position} ($6.1$~mm), Franka \textsc{Line} $+$ \textsc{Orientation} ($4.6$~mm) and Franka $\LieGroupSE{3}$ ($3.1$~mm), are the bias floors of those conditions.
The fraction of configurations inside the validity tolerances of Table~\ref{tab:tbar} is $100\%$ before and after projection on nine conditions, rises to $100\%$ on UR5 \textsc{Position}, \textsc{Line} $+$ \textsc{Orientation} and $\LieGroupSE{3}$ (from $99$, $92$ and $99\%$), and stays at $98\%$ on Franka \textsc{Orientation} and \textsc{Plane} $+$ \textsc{Orientation}.

\begin{table}[t]
\centering
\small
\setlength{\tabcolsep}{5pt}
\renewcommand{\arraystretch}{1.15}
\caption{\textbf{Constraint error of generated configurations before and after score projection.}
Median $\Norm{h_{\mathbf{c}}(\mathbf{q})}$ ($\times 10^{-3}$: mm for position, mrad for orientation) over the $120$ P1 configurations per condition, before (Init.) and after (Final) $M = 20$ iterations of Algorithm~\ref{alg:score-projection}, with the median per-configuration reduction factor (Red.).
On the same starts, analytical Newton--Raphson reaches zero error and the random-direction control diverges.}
\label{tab:proj-p1}
\begin{tabular}{@{}llccc@{}}
\toprule
& Constraint & Init. & Final & Red. \\
\midrule
\multirow{7}{*}{\rotatebox[origin=c]{90}{\textsc{Franka}}}
 & \textsc{Plane} & $3.0$ & $0.54$ & $\times5.9$ \\
 & \textsc{Line} & $6.8$ & $1.6$ & $\times3.9$ \\
 & \textsc{Position} & $9.8$ & $2.2$ & $\times4.4$ \\
 & \textsc{Orientation} & $33.3$ & $13.0$ & $\times2.7$ \\
 & \textsc{Plane} $+$ \textsc{Orientation} & $36.8$ & $11.4$ & $\times3.1$ \\
 & \textsc{Line} $+$ \textsc{Orientation} & $33.5$ & $18.9$ & $\times1.5$ \\
 & $\LieGroupSE{3}$ & $40.1$ & $12.3$ & $\times3.3$ \\
\midrule
\multirow{7}{*}{\rotatebox[origin=c]{90}{\textsc{UR5}}}
 & \textsc{Plane} & $3.6$ & $0.59$ & $\times6.2$ \\
 & \textsc{Line} & $6.1$ & $1.0$ & $\times6.8$ \\
 & \textsc{Position} & $9.5$ & $1.9$ & $\times3.1$ \\
 & \textsc{Orientation} & $28.2$ & $8.0$ & $\times4.1$ \\
 & \textsc{Plane} $+$ \textsc{Orientation} & $26.9$ & $5.9$ & $\times4.6$ \\
 & \textsc{Line} $+$ \textsc{Orientation} & $51.8$ & $20.3$ & $\times2.6$ \\
 & $\LieGroupSE{3}$ & $33.0$ & $8.9$ & $\times4.3$ \\
\bottomrule
\end{tabular}
\end{table}

\begin{table}[t]
\centering
\small
\setlength{\tabcolsep}{5pt}
\renewcommand{\arraystretch}{1.15}
\caption{\textbf{Re-projection of exact IK solutions displaced off the manifold.}
Final median $\Norm{h_{\mathbf{c}}(\mathbf{q})}$ ($\times 10^{-3}$) after $M = 20$ iterations of Algorithm~\ref{alg:score-projection}, starting from exact analytical IK solutions displaced by a task-space error $\delta \in \{1, 5, 10, 20\} \times 10^{-3}$, which is also the initial error.
Each row is nearly flat and the $\delta = 1$ starts are pulled up to it except on \textsc{Plane}, so the final error is the offset between the learned and the analytical manifold.}
\label{tab:proj-p2}
\begin{tabular}{@{}llcccc@{}}
\toprule
& Constraint & $\delta = 1$ & $5$ & $10$ & $20$ \\
\midrule
\multirow{7}{*}{\rotatebox[origin=c]{90}{\textsc{Franka}}}
 & \textsc{Plane} & $0.71$ & $0.72$ & $0.69$ & $0.80$ \\
 & \textsc{Line} & $1.6$ & $1.5$ & $1.6$ & $2.3$ \\
 & \textsc{Position} & $1.7$ & $1.9$ & $2.0$ & $2.8$ \\
 & \textsc{Orientation} & $12.3$ & $12.4$ & $12.7$ & $12.7$ \\
 & \textsc{Plane} $+$ \textsc{Orientation} & $10.1$ & $10.5$ & $10.8$ & $11.4$ \\
 & \textsc{Line} $+$ \textsc{Orientation} & $19.3$ & $19.6$ & $19.4$ & $20.4$ \\
 & $\LieGroupSE{3}$ & $9.5$ & $9.7$ & $10.1$ & $12.6$ \\
\midrule
\multirow{7}{*}{\rotatebox[origin=c]{90}{\textsc{UR5}}}
 & \textsc{Plane} & $0.70$ & $0.74$ & $0.80$ & $0.81$ \\
 & \textsc{Line} & $1.2$ & $1.2$ & $1.2$ & $1.4$ \\
 & \textsc{Position} & $2.5$ & $2.5$ & $2.5$ & $3.8$ \\
 & \textsc{Orientation} & $8.8$ & $8.4$ & $7.8$ & $9.3$ \\
 & \textsc{Plane} $+$ \textsc{Orientation} & $7.7$ & $7.5$ & $7.3$ & $7.2$ \\
 & \textsc{Line} $+$ \textsc{Orientation} & $21.4$ & $20.4$ & $21.4$ & $19.9$ \\
 & $\LieGroupSE{3}$ & $7.6$ & $8.1$ & $9.2$ & $10.5$ \\
\bottomrule
\end{tabular}
\end{table}

\begin{table}[t]
\centering
\small
\setlength{\tabcolsep}{6pt}
\renewcommand{\arraystretch}{1.15}
\caption{\textbf{Position and orientation error before and after projection, in the convention of Table~\ref{tab:size-sweep-compare}.}
Mean position error (mm) and mean orientation error ($^\circ$) over all $120$ generated configurations per condition, before (Generated) and after (After projection) $M = 20$ iterations of Algorithm~\ref{alg:score-projection}; unlike Table~\ref{tab:size-sweep-compare}, the means include configurations outside the validity tolerances.
The ID column is the analytical intrinsic dimension; a dash marks a metric the constraint does not fix.}
\label{tab:proj-posorn}
\begin{tabular}{@{}llccccc@{}}
\toprule
& & & \multicolumn{2}{c}{Generated} & \multicolumn{2}{c}{After projection} \\
\cmidrule(lr){4-5}\cmidrule(l){6-7}
& Constraint & ID & Pos.\ [mm] & Orn.\ [$^\circ$] & Pos.\ [mm] & Orn.\ [$^\circ$] \\
\midrule
\multirow{7}{*}{\rotatebox[origin=c]{90}{\textsc{Franka}}}
 & \textsc{Plane} & 6 & $3.9$ & -- & $0.7$ & -- \\
 & \textsc{Line} & 5 & $8.5$ & -- & $2.1$ & -- \\
 & \textsc{Position} & 4 & $11.3$ & -- & $2.5$ & -- \\
 & \textsc{Orientation} & 4 & -- & $4.9$ & -- & $3.2$ \\
 & \textsc{Plane} $+$ \textsc{Orientation} & 3 & $4.4$ & $3.0$ & $1.1$ & $1.8$ \\
 & \textsc{Line} $+$ \textsc{Orientation} & 3 & $10.6$ & $2.2$ & $4.6$ & $1.4$ \\
 & $\LieGroupSE{3}$ & 1 & $12.3$ & $2.3$ & $3.1$ & $0.8$ \\
\midrule
\multirow{7}{*}{\rotatebox[origin=c]{90}{\textsc{UR5}}}
 & \textsc{Plane} & 5 & $4.5$ & -- & $0.7$ & -- \\
 & \textsc{Line} & 4 & $7.0$ & -- & $1.1$ & -- \\
 & \textsc{Position} & 3 & $13.4$ & -- & $6.1$ & -- \\
 & \textsc{Orientation} & 3 & -- & $2.0$ & -- & $0.5$ \\
 & \textsc{Plane} $+$ \textsc{Orientation} & 2 & $4.5$ & $1.6$ & $0.9$ & $0.3$ \\
 & \textsc{Line} $+$ \textsc{Orientation} & 2 & $7.7$ & $5.7$ & $2.6$ & $1.7$ \\
 & $\LieGroupSE{3}$ & 0 & $9.2$ & $2.1$ & $2.2$ & $0.6$ \\
\bottomrule
\end{tabular}
\end{table}

\paragraph{What the projector changes.}
The raw and the projected score reach the same final error (Figure~\ref{fig:proj-residual}): the raw score's final median is within $10\%$ of the projected one on every condition, and slightly lower on $10$ of $14$, so the projector does not lower the error, but changes the denoising path.
The fraction of each step lying in the analytical normal space $\mathrm{row}\,Dh_{\mathbf{c}}$, over iterations $1$--$5$, is $0.83$--$1.00$ for the projected score against $0.01$--$0.21$ for the raw score on eleven of the twelve continuous conditions ($0.46$ on UR5 \textsc{Line} $+$ \textsc{Orientation}); on UR5 $\LieGroupSE{3}$ every direction is normal and the two coincide.
The estimated normal space agrees with the analytical one: at $t = 15$ the median largest principal angle between $\mathrm{span}\,U_{\hat{k}}$ and $\mathrm{row}\,Dh_{\mathbf{c}}$ is $3.6$--$8.7^\circ$ on the Franka and $0$--$11.9^\circ$ on the UR5, and the median $\hat{k}$ equals the codimension of $\mathcal{M}_{\mathbf{c}}$ in the sine--cosine space on all fourteen conditions ($59$--$91\%$ of the individual estimates).
On the $\LieGroupSE{3}$ conditions the projected update leaves $97.5\%$ of the Franka configurations at their initial nearest solution against $96.4\%$ for the raw score, and $100\%$ of the UR5 configurations for both.

\begin{figure}[t]
\centering
\includegraphics[width=\linewidth]{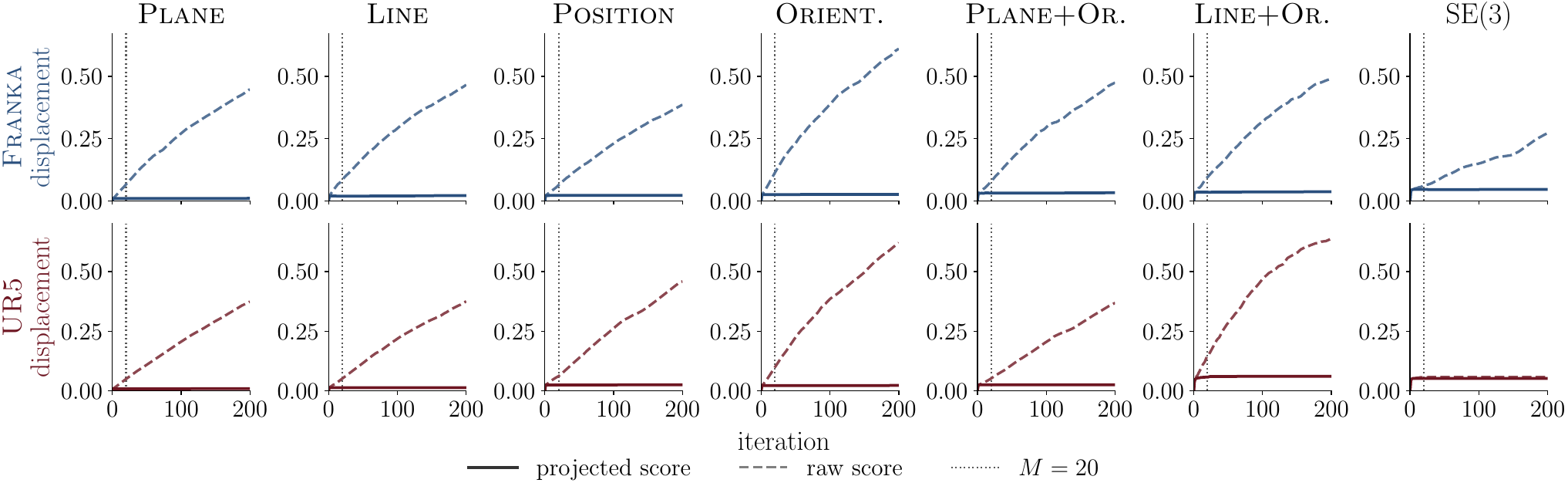}
\vspace{-4mm}
\caption{\textbf{Displacement from the start configuration over $200$ iterations.}
Median distance in the standardized sine--cosine space between the current and the initial configuration over the $120$ generated configurations, for the raw score (dashed) and the projected score (solid); \textcolor{frankaBlue}{Franka} (top) and \textcolor{ur5Red}{UR5} (bottom). The dotted vertical line marks the $M = 20$ of the main experiment.
The raw iterate keeps sliding along $\mathcal{M}_{\mathbf{c}}$; the projected one is stationary once the normal error is removed.}
\label{fig:proj-drift}
\end{figure}

\paragraph{Long-run behaviour.}
Running both updates for $M = 200$ iterations on the generated configurations separates them (Figure~\ref{fig:proj-drift}).
The displacement of the raw-score iterate from its start grows by a factor of $4.3$--$7.6$ between iterations $20$ and $200$ on the twelve continuous conditions, whereas that of the projected iterate changes by a factor of $1.0$--$1.1$; at iteration $200$ the raw iterate has moved $10.5$--$44.7$ times further.
The constraint error at iteration $200$ is within $25\%$ of its value at iteration $20$ for both updates on every condition, so the motion is along $\mathcal{M}_{\mathbf{c}}$: the tangential component of the raw score carries the configuration towards regions of higher density, whereas the projected update has a fixed point once the normal error is removed.
On Franka $\LieGroupSE{3}$ the raw score changes the nearest solution of $60$ of the $120$ configurations against $12$ for the projected score, and on UR5 $\LieGroupSE{3}$, which has no tangent space, the two are indistinguishable.

\subsection{Intrinsic dimension as a convergence diagnostic}
\label{app:convergence}

\paragraph{Protocol.}
We use the per-family models of Appendix~\ref{app:samples} ($100{,}000$ steps) at $N \in \{500, 1\mathrm{k}, 2\mathrm{k}, 5\mathrm{k}, 10\mathrm{k}, 20\mathrm{k}, 40\mathrm{k}\}$, together with one further model per condition at $N = 50{,}000$ trained with the identical configuration (none for Franka \textsc{Orientation} and \textsc{Line} $+$ \textsc{Orientation}, whose largest model is $N = 40\mathrm{k}$).
For each model we compute $S$ with $K = 50$ at $t = 15$ at two kinds of configuration: \emph{unseen}, $5$ held-out analytical IK solutions of each of $15$ targets, drawn uniformly from the solution set as in Appendix~\ref{app:id-hyperparameters}, and \emph{seen}, $75$ random rows of the model's own training set; every estimate is the median over the $75$ configurations.
From the singular values $\sigma_i$ of $S$ we report the effective-rank estimate $\hat{d}_{\mathrm{eff}}$ of Section~\ref{sec:convergence}, which varies continuously, and the integer estimate $\hat{d}$ of Algorithm~\ref{alg:score-svd-dagger}.
Constraint errors are computed as in Appendix~\ref{app:samples} for the UR5 and at $20$ targets $\times$ $256$ samples for the Franka.

\begin{figure}[t]
\centering
\includegraphics[width=\linewidth]{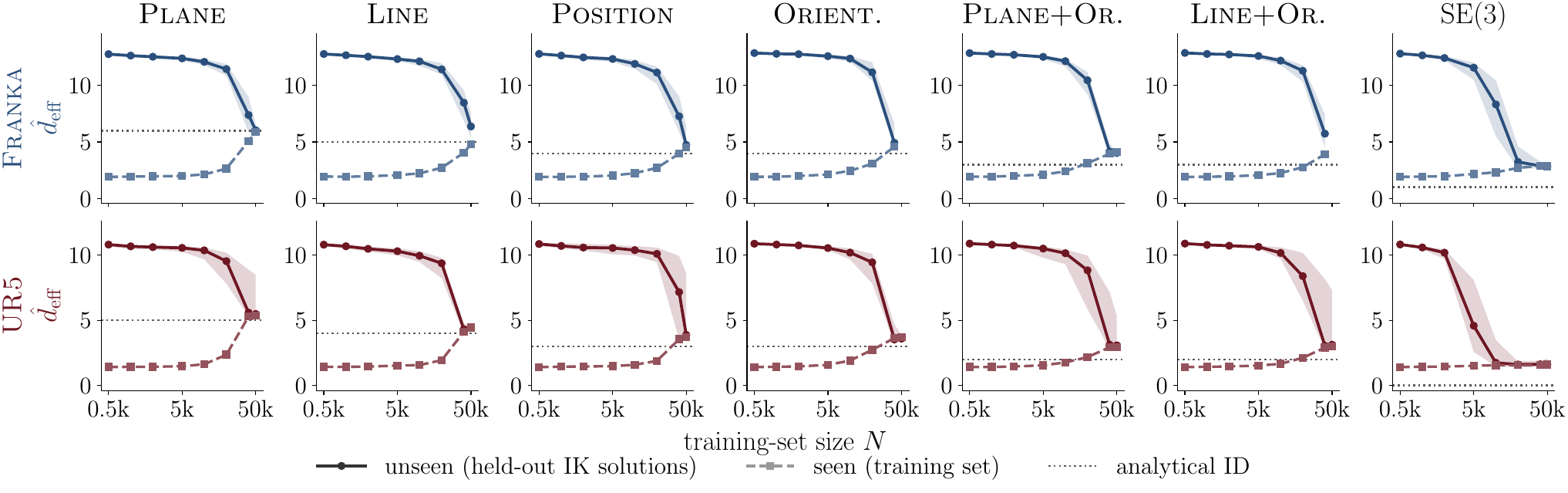}
\vspace{-4mm}
\caption{\textbf{Effective-rank intrinsic dimension at unseen and seen configurations versus training-set size, all fourteen conditions.}
$\hat{d}_{\mathrm{eff}}$ at unseen configurations (held-out analytical IK solutions; solid, median with interquartile band) and at seen configurations (training set; dashed) against $N$, for \textcolor{frankaBlue}{Franka} (top) and \textcolor{ur5Red}{UR5} (bottom); the dotted line is the analytical intrinsic dimension.
The two estimates approach the analytical value from opposite sides; the right panel of Figure~\ref{fig:idgap} is the difference of the two UR5 curves.}
\label{fig:idgap-grid}
\end{figure}

\paragraph{Results.}
Figure~\ref{fig:idgap-grid} shows both estimates on every condition and Table~\ref{tab:idgap-full} their values at the smallest and the largest $N$.
On the UR5 the gap at $N = 50\mathrm{k}$ is below $0.3$ on all seven constraints, and Score-SVD$^{\dagger}$ returns the analytical dimension on all seven.
On the Franka the gap has closed to within $0.3$ on \textsc{Plane}, \textsc{Position}, \textsc{Plane} $+$ \textsc{Orientation} and $\LieGroupSE{3}$, where the integer estimate is also exact except on \textsc{Plane} ($5$ against $6$), and is still decreasing on \textsc{Line} (from $4.4$ at $N = 40\mathrm{k}$ to $1.6$ at $50\mathrm{k}$).
For \textsc{Orientation} and \textsc{Line} $+$ \textsc{Orientation}, whose largest models are at $N = 40\mathrm{k}$, the gaps are $0.34$ and $1.82$, down from $8.1$ and $8.5$ at $N = 20\mathrm{k}$.

\begin{table}[t]
\centering
\small
\setlength{\tabcolsep}{5pt}
\renewcommand{\arraystretch}{1.15}
\caption{\textbf{Effective-rank intrinsic dimension at unseen and seen configurations.}
$\hat{d}_{\mathrm{eff}}$ at unseen (held-out analytical IK solutions) and seen (training set) configurations at the smallest and largest training-set size, and their difference (gap) at the largest size.
$^\ast$\,$N = 40\mathrm{k}$, the largest model of this family.}
\label{tab:idgap-full}
\begin{tabular}{@{}llcccccc@{}}
\toprule
& & & \multicolumn{2}{c}{$N = 500$} & \multicolumn{3}{c}{$N = 50\mathrm{k}$} \\
\cmidrule(lr){4-5}\cmidrule(lr){6-8}
& Constraint & ID & unseen & seen & unseen & seen & gap \\
\midrule
\multirow{7}{*}{\rotatebox[origin=c]{90}{\textsc{Franka}}}
 & \textsc{Plane} & 6 & $12.8$ & $1.9$ & $6.0$ & $5.9$ & $0.17$ \\
 & \textsc{Line} & 5 & $12.8$ & $2.0$ & $6.4$ & $4.8$ & $1.57$ \\
 & \textsc{Position} & 4 & $12.8$ & $1.9$ & $4.8$ & $4.5$ & $0.24$ \\
 & \textsc{Orientation}$^\ast$ & 4 & $12.9$ & $1.9$ & $4.9$ & $4.6$ & $0.34$ \\
 & \textsc{Plane} $+$ \textsc{Orientation} & 3 & $12.9$ & $1.9$ & $4.0$ & $4.1$ & $-0.07$ \\
 & \textsc{Line} $+$ \textsc{Orientation}$^\ast$ & 3 & $12.9$ & $1.9$ & $5.7$ & $3.9$ & $1.82$ \\
 & $\LieGroupSE{3}$ & 1 & $12.8$ & $1.9$ & $2.8$ & $2.9$ & $-0.01$ \\
\midrule
\multirow{7}{*}{\rotatebox[origin=c]{90}{\textsc{UR5}}}
 & \textsc{Plane} & 5 & $10.8$ & $1.4$ & $5.5$ & $5.4$ & $0.12$ \\
 & \textsc{Line} & 4 & $10.8$ & $1.4$ & $4.5$ & $4.4$ & $0.03$ \\
 & \textsc{Position} & 3 & $10.9$ & $1.4$ & $3.9$ & $3.7$ & $0.17$ \\
 & \textsc{Orientation} & 3 & $10.9$ & $1.4$ & $3.6$ & $3.7$ & $-0.10$ \\
 & \textsc{Plane} $+$ \textsc{Orientation} & 2 & $10.9$ & $1.4$ & $3.1$ & $2.9$ & $0.14$ \\
 & \textsc{Line} $+$ \textsc{Orientation} & 2 & $10.9$ & $1.4$ & $3.1$ & $3.0$ & $0.13$ \\
 & $\LieGroupSE{3}$ & 0 & $10.8$ & $1.4$ & $1.6$ & $1.6$ & $0.00$ \\
\bottomrule
\end{tabular}
\end{table}

\begin{figure}[t]
\centering
\includegraphics[width=0.6\linewidth]{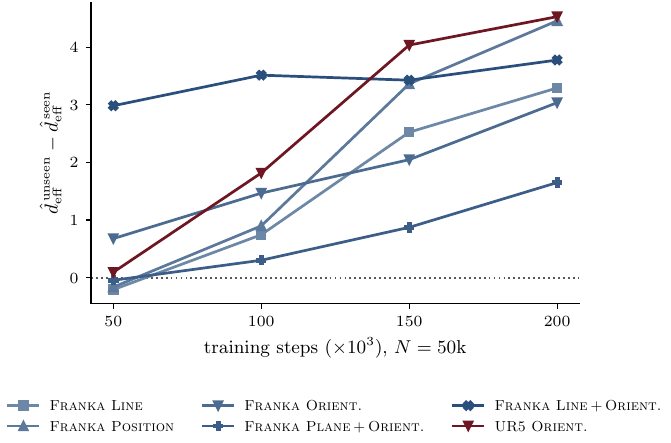}
\vspace{-2mm}
\caption{\textbf{Seen--unseen gap versus training steps at $N = 50\mathrm{k}$.}
The six cells of Table~\ref{tab:idgap-steps}; the gap is smallest at $50{,}000$ steps and grows with further training while the constraint error stays flat.}
\label{fig:idgap-steps}
\end{figure}

\paragraph{Dependence on training length.}
Six $N = 50\mathrm{k}$ models were trained for $200{,}000$ steps with a checkpoint every $50{,}000$. These runs come from an earlier version of the sweep that used the cosine noise schedule and, for Franka \textsc{Orientation} and \textsc{Line} $+$ \textsc{Orientation}, the data-generation procedure of Table~\ref{tab:target-sampling}, so their $100{,}000$-step values differ from those of Table~\ref{tab:idgap-full}.
Table~\ref{tab:idgap-steps} and Figure~\ref{fig:idgap-steps} show that the gap is smallest at $50{,}000$ steps on every cell and larger at $200{,}000$ steps on every cell: the unseen estimate rises above the analytical value while the seen one falls below it, the signature of a model that fits its training configurations more tightly than the manifold, whereas the constraint error changes by at most $1.4$~mm or $0.6^\circ$ between $100{,}000$ and $200{,}000$ steps.
The models of Table~\ref{tab:idgap-full} show the same effect between their two checkpoints on Franka \textsc{Line}, whose gap is $-0.05$ at $50{,}000$ steps and $1.57$ at $100{,}000$.
The gap therefore also detects training past what the data supports, which the constraint error does not show.

\begin{table}[t]
\centering
\small
\setlength{\tabcolsep}{5pt}
\renewcommand{\arraystretch}{1.15}
\caption{\textbf{The seen--unseen gap reopens with training length.}
Gap $\hat{d}_{\mathrm{eff}}$ (unseen) $-$ $\hat{d}_{\mathrm{eff}}$ (seen) of six $N = 50\mathrm{k}$ models (cosine noise schedule) at four training lengths, and their constraint error (mean over valid samples) at $100{,}000$ and $200{,}000$ steps.
$^\ast$\,$N = 47{,}500$.}
\label{tab:idgap-steps}
\begin{tabular}{@{}llcccccc@{}}
\toprule
& & & \multicolumn{4}{c}{gap at training steps ($\times 10^3$)} & \\
\cmidrule(lr){4-7}
& Constraint & ID & $50$ & $100$ & $150$ & $200$ & error, $100 \to 200$ \\
\midrule
\textsc{Franka} & \textsc{Line} & 5 & $-0.21$ & $0.74$ & $2.52$ & $3.29$ & $17.9 \to 18.6$~mm \\
\textsc{Franka} & \textsc{Position} & 4 & $-0.16$ & $0.90$ & $3.36$ & $4.46$ & $22.2 \to 23.1$~mm \\
\textsc{Franka} & \textsc{Orientation} & 4 & $0.68$ & $1.47$ & $2.05$ & $3.04$ & $7.3 \to 7.3^\circ$ \\
\textsc{Franka} & \textsc{Plane} $+$ \textsc{Orientation} & 3 & $-0.05$ & $0.30$ & $0.87$ & $1.65$ & $8.9 \to 9.1$~mm, $5.3 \to 5.9^\circ$ \\
\textsc{Franka} & \textsc{Line} $+$ \textsc{Orientation}$^\ast$ & 3 & $2.98$ & $3.51$ & $3.43$ & $3.78$ & $36.5 \to 37.9$~mm, $9.2 \to 9.8^\circ$ \\
\textsc{UR5} & \textsc{Orientation} & 3 & $0.09$ & $1.82$ & $4.04$ & $4.53$ & $4.1 \to 4.0^\circ$ \\
\bottomrule
\end{tabular}
\end{table}

